\documentclass{article}

\PassOptionsToPackage{numbers, compress}{natbib}
\usepackage[final]{neurips_2021}

\usepackage[utf8]{inputenc} % allow utf-8 input
\usepackage[T1]{fontenc}    % use 8-bit T1 fonts
\usepackage{hyperref}       % hyperlinks
\usepackage{url}            % simple URL typesetting
\usepackage{booktabs}       % professional-quality tables
\usepackage{amsfonts}       % blackboard math symbols
\usepackage{nicefrac}       % compact symbols for 1/2, etc.
\usepackage{microtype}      % microtypography
\usepackage{xcolor}         % colors
\usepackage{amsmath}
\usepackage{bbm}
\usepackage{graphicx}
\usepackage{subcaption}
\usepackage{soul}
\usepackage{multirow}

\usepackage{algorithm}
\usepackage{algorithmic}

\usepackage{xcolor}

\usepackage{amsfonts}
\usepackage{amsmath}
\usepackage{amssymb}
\usepackage{dsfont}
\usepackage{xcolor}
\usepackage{tcolorbox}

\usepackage{xspace}  % comments
\usepackage[normalem]{ulem}  %sout
\usepackage{wrapfig}

\usepackage{xr}
\makeatletter
\newcommand*{\addFileDependency}[1]{
  \typeout{(#1)}
  \@addtofilelist{#1}
  \IfFileExists{#1}{}{\typeout{No file #1.}}
}
\makeatother

\newcommand*{\myexternaldocument}[1]{
    \externaldocument{#1}
    \addFileDependency{#1.tex}
    \addFileDependency{#1.aux}
}

\renewcommand{\vec}[1]{\mathbf{#1}}

\newcommand{\argmax}{\mathop{\mathrm{argmax}}}

\myexternaldocument{supp}

\usepackage{xcolor}
\definecolor{dark-red}{rgb}{0.4,0.15,0.15}
\definecolor{dark-blue}{rgb}{0.15,0.15,0.4}
\definecolor{medium-blue}{rgb}{0,0,0.5}
\hypersetup{
   colorlinks, linkcolor={dark-blue},
   citecolor={dark-blue}, urlcolor={medium-blue}
}

\title{Does Knowledge Distillation Really Work?}

\author{%
  Samuel Stanton \\
  NYU \\
  \And
  Pavel Izmailov \\
  NYU \\
  \And Polina Kirichenko \\
  NYU \\
  \AND Alexander A. Alemi \\
  Google Research \\
  \And
  Andrew Gordon Wilson \\
  NYU \\
}

\begin{document}

\definecolor{lightblue}{HTML}{afe9af}
\definecolor{lighterblue}{HTML}{afe9af}
\newtcolorbox{mybox}{colback=lighterblue,colframe=lightblue}

\maketitle

\begin{abstract}
Knowledge distillation is a popular technique for training a small student network to emulate a larger teacher model, such as an ensemble of networks. We show that while knowledge distillation can improve student generalization, it does not typically work as it is commonly understood: there often remains a surprisingly large discrepancy between the predictive distributions of the teacher and the student, even in cases when the student has the capacity to perfectly match the teacher. 
We identify difficulties in optimization as a key reason for why the student is unable to match the teacher. We also show how the details of the dataset used for distillation play a role in how closely the student matches the teacher --- and that more closely matching the teacher paradoxically does not always lead to better student generalization.
\end{abstract}

\section{Introduction}
\label{main/sec:intro}

Large, deep networks can learn representations that generalize well.
While smaller, more efficient networks lack the \emph{inductive biases} to find these representations from training data alone, they may have the \emph{capacity} to represent these solutions \citep[e.g.,][]{ba2014deep, han2015deep, lecun1989optimal, romero2014fitnets}. 
Influential work on \emph{knowledge distillation} \citep{hinton2015distilling} argues that \citet{bucila2006model} ``demonstrate convincingly that the knowledge acquired by a large ensemble of models [the teacher] can be transferred to a single small model [the student]''. 
Indeed this quote encapsulates the conventional narrative of knowledge distillation: a  student model learns a high-fidelity representation of a larger teacher, enabled by the teacher's soft labels.

\begin{figure}[t]
    \centering

  \begin{tabular}{cc}
      \includegraphics[width=0.45\textwidth]{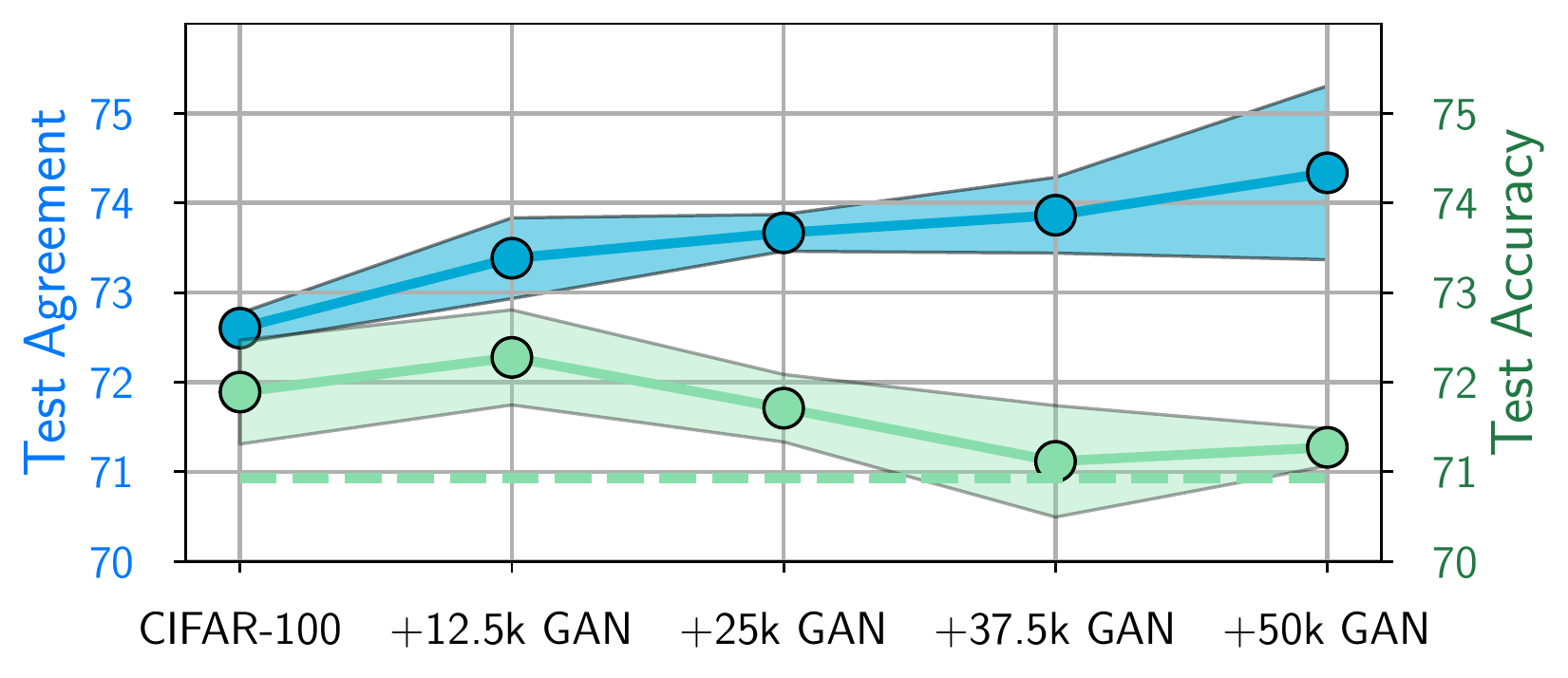}
      &
      \includegraphics[width=0.45\textwidth]{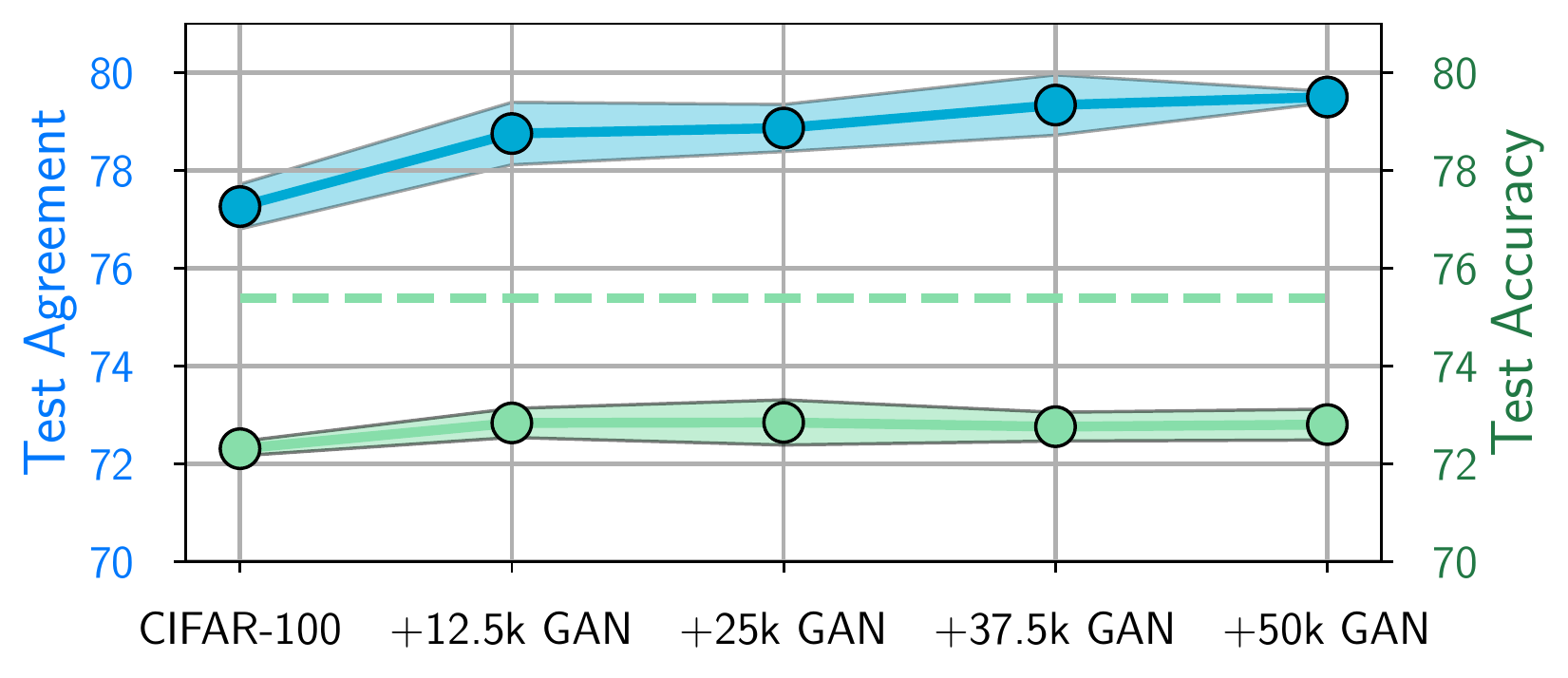}
    \\
      \multicolumn{2}{c}{\includegraphics[width=0.55\textwidth]{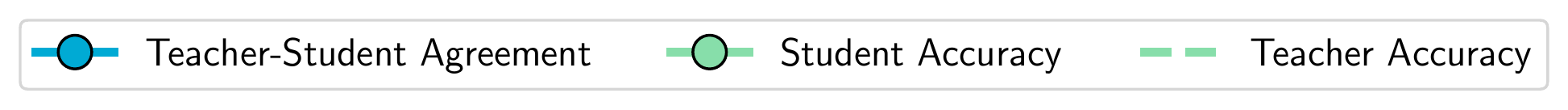}}
    \\
      {\small (a) Self-distillation} 
      & 
      {\small (b) Ensemble distillation}
  \end{tabular}
    \caption{
    \textbf{Evaluating the fidelity of knowledge distillation.}
    The effect of enlarging the CIFAR-100 distillation dataset with GAN-generated samples.
    \textbf{(a)}: The student and teacher are both single ResNet-56 networks. Student fidelity increases as the dataset grows, but test accuracy decreases.
    \textbf{(b)}: The student is a single ResNet-56 network and the teacher is a 3-component ensemble. Student fidelity again increases as the dataset grows, but test accuracy now slightly increases.
    The shaded region corresponds to $\mu \pm \sigma$, estimated over 3 trials.
    }
    \label{main/fig:cifar100_motivation}
\end{figure}

Conversely, in Figure \ref{main/fig:cifar100_motivation} we show that with modern architectures knowledge distillation can lead to students with very different predictions from their teachers, even when the student has the capacity to perfectly match the teacher. Indeed, it is becoming well-known that in self-distillation the student fails to match the teacher and, paradoxically, student generalization improves as a result \citep{furlanello2018born, mobahi2020self}. However, when the teacher is a large model (e.g. a deep ensemble) improvements in fidelity translate into improvements in generalization, as we show in Figure \ref{main/fig:cifar100_motivation}(b). For these large models there is still a significant accuracy gap between student and teacher, so fidelity is aligned with generalization.

We will distinguish between \emph{fidelity}, the ability of a student to match a teacher's predictions, and \emph{generalization}, the performance of a student in predicting unseen, in-distribution data. We show that in many cases it is surprisingly difficult to obtain good student fidelity. In Section \ref{main/sec:identifiability} we investigate the hypothesis that low  fidelity is an \emph{identifiability} problem that can be solved by augmenting the distillation dataset.
In Section \ref{main/sec:optimization} we investigate the hypothesis that  low fidelity is an \emph{optimization} problem resulting in a failure of the student to  
match the teacher even on the original training dataset.
We present a summary of our conclusions in Section \ref{main/sec:discussion}.

\emph{Does knowledge distillation really work?} 
In short: \emph{Yes}, in the sense that it often improves student generalization, though there is room for further improvement.
\emph{No}, in that knowledge distillation often fails to live up to its name, transferring very limited knowledge from teacher to student.

\section{Related Work}
\label{sec: related}

Knowledge distillation can improve model efficiency \citep{mishra2018apprentice, romero2014fitnets}, unsupervised domain adaptation \citep{meng2018adversarial}, improved object detection \citep{chen2017learning}, model transparency \citep{tan2018distill}, and adversarial robustness \citep{goldblum2020adversarially, papernot2016distillation}. 

Seminal work by \citet{bucila2006model} showed that teacher-ensembles with thousands of simple components could be compressed into a single shallow network that matched or outperformed its teacher. Other early work proposed distilling ensembles of shallow networks into a single network \citep{zeng2000using}, an idea which resonates with more recent work on the distillation of deep ensembles \citep{ba2014deep, chebotar2016distilling, shen2019meal, urban2016deep, you2017learning}. 
Recently \citet{fakoor2020fast} developed a data-augmentation scheme for the distillation of large ensembles of simple models for tabular data, achieving impressive results on a wide range of tabular benchmarks. \citet{malinin2019ensemble} proposed a method to model the implicit distribution over predictive distributions from which the ensemble component predictive distributions are drawn, rather than just the ensemble model average.

Our work focuses explicitly on student fidelity, decoupling our understanding of good fidelity from good generalization. We show that achieving good fidelity is extremely difficult, even with a variety of interventions, and seek to \emph{understand}, by systematically considering several hypotheses, why knowledge distillation does not produce high fidelity students for modern architectures and datasets.
In contrast, the distillation literature focuses largely on improving student generalization, without particularly distinguishing between fidelity and generalization.

For example, concurrent work by \citet{beyer2021knowledge} does not carefully distinguish generalization and fidelity metrics, but they assert that high student fidelity is \textit{conceptually} desirable and apparently difficult to achieve when measured as the gap between teacher and student accuracy.
As a result their work focuses most heavily on practical modifications to the distillation procedure for the best student top-1 accuracy.
In this paper we investigate many of the same prescriptions, including careful treatment of data augmentation (such as showing the teacher and student the exact same input images), the addition of MixUp, and extended training duration. 
We also find that such interventions do improve student accuracy, but there still remains a large discrepancy between the predictive distributions of the teacher and the student. We also investigate multiple optimizers. While we do not pursue Shampoo \citep{gupta2018shampoo, anil2021scalable} specifically, \citet{beyer2021knowledge} find similar qualitative results for Shampoo and Adam, besides faster convergence for Shampoo.

\section{Preliminaries}
\label{main/sec:preliminaries}

We will focus on the supervised classification setting, with input space $\mathcal{X}$ and label space $\mathcal{Y}$, where $|\mathcal{Y}| = c$.
Let $f: \mathcal{X} \times \Theta \rightarrow \mathbb{R}^c$ be a classifier parameterized by $\theta \in \Theta$ whose outputs define a categorical predictive distribution over $\mathcal{Y}$, $\hat p(y=i | \vec x) = \sigma_i(f(\vec x, \theta))$, where $\sigma_i(\vec z) := \exp(z_i) / \sum_j \exp(z_j)$ is the softmax link function. 
We will often refer to the outputs of a classifier $\vec z := f(\vec x, \theta)$ as \textit{logits}. For convenience, we will use $t$ and $s$ as shorthand for  $f_{\mathrm{teacher}}$ and $f_{\mathrm{student}}$, respectively. When the teacher is an $m$-component ensemble, the component logits $(\vec z_1, \dots, \vec z_m)$, where $\vec z_i = f_i(\vec x, \theta_i)$, are combined to form the teacher logits: $\vec z_t = \log \left(\sum_{i=1}^m \sigma(\vec z_i) / m \right)$.
These combined logits correspond to the predictive distribution of the ensemble model average. 
The experiments in the main text consider $m \in \{1, 3, 5\}$, and we include results up to $m=12$ in Appendix \ref{supp/subsec:ensemble_size_ablation}.\footnote{Code for all experiments can be found at \url{https://github.com/samuelstanton/gnosis}.}

\subsection{Knowledge Distillation}
\label{main/subsec:kd_background}

\citet{hinton2015distilling} proposed
a simple approach to knowledge distillation.
The student minimizes a weighted combination of two objectives, 
$\mathcal{L}_s := \alpha \mathcal{L}_{\mathrm{NLL}} + (1 - \alpha) \mathcal{L}_{\mathrm{KD}}$, where $\alpha \in [0, 1)$. Specifically, 
\begin{align}
    \mathcal{L}_{\mathrm{NLL}}(\vec z_s, \vec y) := -\sum_{j=1}^c y_j \log\sigma_j(\vec z_s), \hspace{4mm}
    \mathcal{L}_{\mathrm{KD}}(\vec z_s, \vec z_t) :=
    - \tau^2 \sum_{j=1}^c \sigma_j\left(\frac{\vec z_t}{\tau}\right) \log \sigma_j\left(\frac{\vec z_s}{\tau}\right). \label{main/eq:kd_loss}
\end{align}
 $\mathcal{L}_{\mathrm{NLL}}$ is the usual supervised cross-entropy between the student logits $\vec z_s$ and the one-hot labels $\vec y$. Recalling that $\mathrm{KL}(p || q) = \sum_j p_j (\log q_j - \log p_j)$, we see that $\mathcal{L}_{\mathrm{NLL}}$ is equivalent (up to a constant) to the KL from the empirical data distribution to the student predictive distribution ($\hat p_s$).
 $\mathcal{L}_{\mathrm{KD}}$ is the added knowledge distillation term that encourages the student to match the teacher. It is the cross-entropy between the teacher and student predictive distributions $\hat p_t = \sigma(\vec z_t)$ and $\hat p_s = \sigma(\vec z_s)$, \emph{both} scaled by a temperature hyperparameter $\tau > 0$. If $\tau = 1$ then $\mathcal{L}_{\mathrm{KD}}$ is similarly equivalent to the KL from the teacher to the student, $\mathrm{KL}(\hat p_t || \hat p_s)$. Since we focus on distillation fidelity, we choose $\alpha = 0$ for all experiments in the main text to avoid any confounding from true labels, but we also include a limited ablation of $\alpha$ in Figure \ref{supp/fig:alpha_ablation} in Appendix \ref{supp/subsec:alpha_ablation} for the curious reader. 
 
As $\tau \rightarrow +\infty$, $\nabla_{\vec z_s} \mathcal{L}_{\mathrm{KD}}(\vec z_s, \vec z_t) \approx \vec z_t - \vec z_s$, and thus in the limit $\nabla_{\vec z_s} \mathcal{L}_{\mathrm{KD}}$ is approximately equivalent to $\nabla_{\vec z_s} || \vec z_t - \vec z_s ||^2_2 / 2$, assigning equal significance to every class logit, regardless of its contribution to the predictive distribution.
In other words $\tau$ determines the ``softness'' of the teacher labels, which in turn determines the allocation of student capacity. 
If the student is much smaller than the teacher, the student capacity can be focused on matching the teacher's top-$k$ predictions, rather than matching the full teacher distribution by choosing a moderate  value (e.g. $\tau=4$). 
In Appendix \ref{supp/subsec:understanding_ensemble_variation} we include further discussion on the interplay of teacher ensemble size, teacher network capacity, and distillation temperature on the student labels.

The teacher and student often share at least some training data. 
It is also common to enlarge the student training data in some way (e.g. incorporating unlabeled examples as in \citet{ba2014deep}). 
When there is a possibility of confusion, we will refer to the student's training data as the \emph{distillation data} to distinguish it from the teacher's training data.

\subsection{Metrics and Evaluation}
\label{main/subsec:define_metrics}

To measure generalization, we report top-1 accuracy, negative log-likelihood (NLL) and expected calibration error (ECE) \citep{guo2017calibration}. To measure fidelity, we report the following:
\begin{align}
    & \text{Average Top-1 Agreement} := \frac{1}{n} \sum_{i=1}^n \mathds{1}\{ \argmax_j \sigma_j(\vec z_{t,i}) = \argmax_j \sigma_j(\vec z_{s, i}) \}, \label{main/eq:agreement_defn} \\
    & \text{Average Predictive KL} := \frac{1}{n} \sum_{i=1}^n \mathrm{KL}\left(\hat p_t(\vec y | \vec x_i) \; || \; \hat p_s(\vec y | \vec x_i)\right), \label{main/eq:ts_kl_defn}
\end{align}
Eqn.~\eqref{main/eq:agreement_defn} 
is the average \emph{agreement} between the student and teacher's top-1 label.
Eqn.~\eqref{main/eq:ts_kl_defn} is the average KL divergence from the predictive distribution of the teacher to that of the student, a measure of fidelity sensitive to all of the labels.

While improvements in generalization metrics are relatively easy to understand, 
interpreting fidelity metrics requires some care.
For example, suppose we have three independent models: $f_1$, $f_2$, and $f_3$ that respectively achieve 
55\%, 75\%, and 95\% test accuracy. $f_1$ and $f_3$ can agree on at most 60\% of points, whereas $f_2$ and $f_3$ agree on at least 70\%, but it would obviously be incorrect to make any claim about $f_2$ being a better distillation of $f_3$ since each model was trained completely independently. To account for such confounding when evaluating the distillation of a student $s$ from a teacher $t$, we also evaluate another student $s'$ distilled through an identical procedure from an independent teacher.  
By comparing the fidelity of $(t, s)$ and $(t, s')$ we can distinguish between a generic improvement in generalization and an improvement specifically to fidelity. If $s$ and $s'$ have comparable fidelity, then the students agree with the teacher at many points because they generalize well, and not the reverse.

\section{Knowledge Distillation Transfers Knowledge Poorly}
\label{main/sec:kd_fails}

In this section, we present evidence that we are not able to distill large networks such as a ResNet-56 with high fidelity, and discuss why high fidelity is an important objective.

\subsection{When is knowledge transfer successful?}
\label{main/subsec:motivation}

We first consider the easy task of distilling a LeNet-5 teacher into an identical student network as a motivating example.
We train the teacher on a random subset of 200 examples from the MNIST training set for 100 epochs, 
resulting in a $84\%$ to  $86\%$ teacher test accuracy across different subsets.\footnote{We took only a subset of the MNIST train set since otherwise every teacher network as well as the ensemble would achieve over $99\%$ test accuracy.}
We then distill the teacher using the full MNIST train dataset with 60,000 examples, as well as 25\%, 50\%, and 100\% of the EMNIST train dataset \citep{cohen2017emnist}. The EMNIST train set contains 697,932 images.

\begin{wrapfigure}{r}{0.4\textwidth}
\centering
\includegraphics[width=0.38\textwidth]{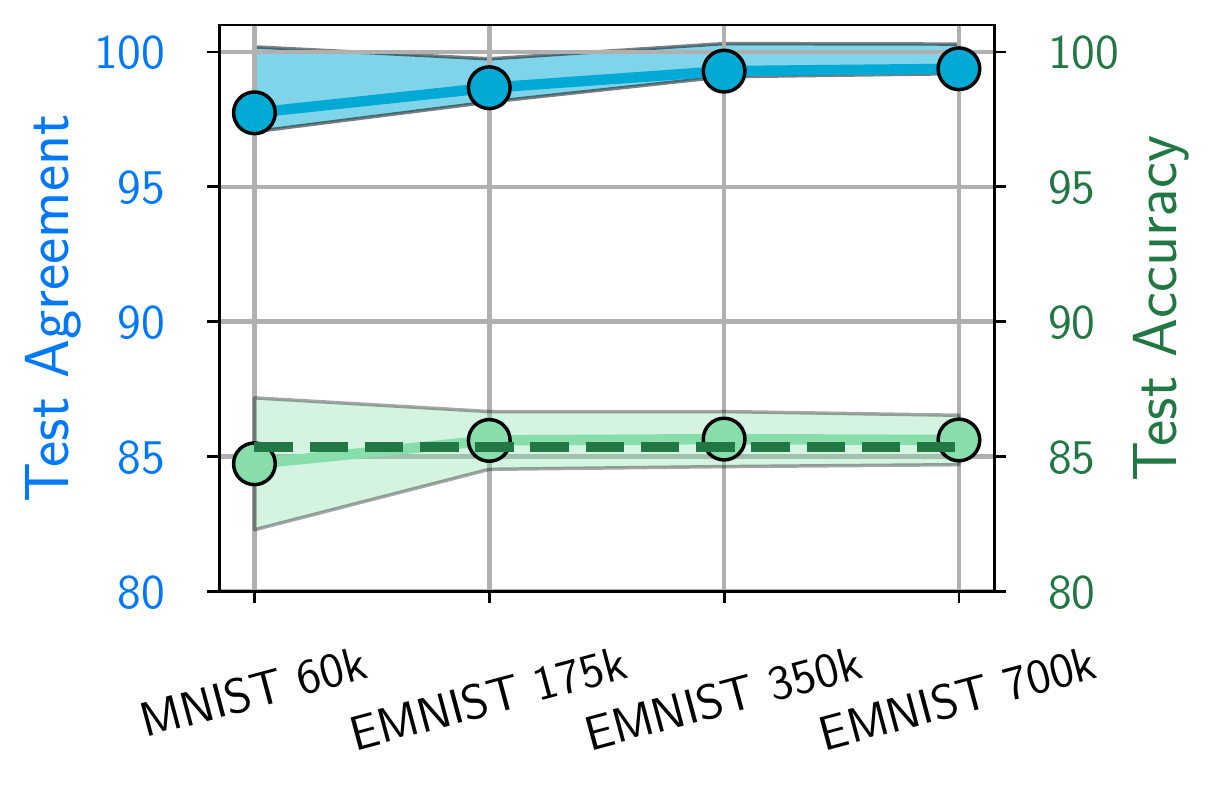}
\includegraphics[width=0.30\textwidth]{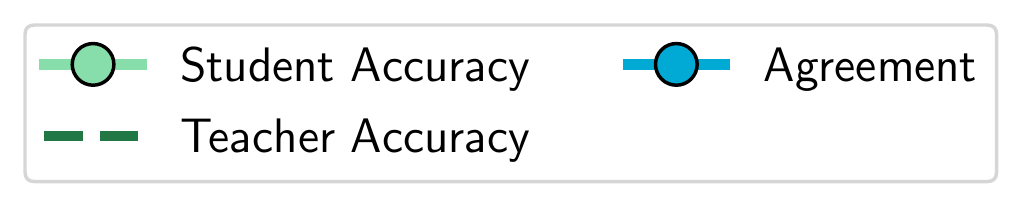}
\caption{
LeNet-5 self-distillation on MNIST with additional distillation data.
The shaded region corresponds to $\mu \pm \sigma$, estimated over 3 trials.
}
\label{main/fig:mnist_motivation}
\end{wrapfigure}

In Figure \ref{main/fig:mnist_motivation} we see that knowledge distillation works as expected.
With enough examples the student learns to make the same predictions as the teacher (over 99\% top-1 test agreement).
Notably, in this case, self-distillation does not \emph{improve} generalization, since the slight difference between the teacher and student accuracy is explained by variance between trials.

Now we consider a more challenging task: distilling a ResNet-56 teacher trained on CIFAR-100 into an identical student network (Figure \ref{main/fig:cifar100_motivation}, left).
Since no dataset drawn from the same distribution as CIFAR-100 is publicly available, to augment the distillation data, we instead combined samples from an SN-GAN \citep{miyato2018spectral} pre-trained on CIFAR-100 with the original CIFAR-100 train dataset. 
Appendix~\ref{supp/subsec:training} details 
the hyperparameters and training procedure for the GAN, teacher, and student.

Like the MNIST experiment, as we enlarge the distillation dataset the student fidelity improves. 
However, in this case the improvement is modest, with the fidelity reaching nowhere near 99\% test agreement. 
Since a ResNet-56 has many more parameters than a LeNet-5, it is possible that the student simply has not seen enough examples to perfectly emulate the teacher, a hypothesis we discuss in more detail in Section \ref{main/subsec:insufficient_data_hypothesis}. Also, like the MNIST experiment, as the distillation dataset grows the student accuracy approaches the teacher's. \emph{Unlike} the MNIST experiment, the student test accuracy is higher than the teacher's when the distillation dataset is small, so increasing fidelity \textit{decreases} student generalization.

\subsection{What can self-distillation tell us about knowledge distillation in general?}

We have seen in Figure \ref{main/fig:cifar100_motivation}(a) that with self-distillation the student can exceed the teacher performance, in accordance with
\citet{furlanello2018born}. This result is only possible by virtue of failing at the distillation procedure: if the student matched the teacher perfectly
then the student could not outperform the teacher. On the other hand, if the teacher generalizes significantly better than an independently trained student, 
we would expect the benefits of fidelity to dominate other regularization effects associated with not matching the teacher. This setting reflects the original 
motivation for knowledge distillation, where we wish to faithfully transfer the representation discovered by a large model or ensemble of models into a more
efficient student. 
 
In Figure \ref{main/fig:cifar100_motivation}(b) we see that if we move from self-distillation to the distillation of a 3 ResNet-56 teacher ensemble, 
fidelity becomes positively correlated with generalization. But there is still a significant gap in fidelity, even after the distillation set is enlarged with
$50k$ GAN samples. In practice, the gap remains large enough that higher fidelity students do not always have better
generalization, and the regularization effects we see in self-distillation do play a role for more broadly understanding student generalization.
We will indeed show in Section \ref{main/sec:identifiability} that higher fidelity students do not always generalize better, even if the teacher generalizes much better than the student.

\subsection{If distillation already improves generalization, why care about fidelity?}

While knowledge distillation does often improve generalization, understanding the relationship between fidelity and generalization, and how to maximize fidelity, is important for several reasons --- including better generalization! 

\textbf{Better generalization in distilling large teacher models and ensembles.}\quad
Knowledge distillation was initially motivated 
as a means to deploy powerful models to small devices or low-latency controllers \citep[e.g.,][]{cho2019efficacy, heo2019knowledge, kim2016sequence, yim2017gift, zagoruyko2016paying}. 
While in self-distillation generalization and fidelity are in tension, there is often a significant disparity in generalization between
large teacher models, including ensembles, and smaller students. We have seen this disparity in Figure \ref{main/fig:cifar100_motivation}(b).
We additionally show in Figure \ref{supp/fig:num_teacher_components} in Appendix \ref{supp/subsec:understanding_ensemble_variation} that as we increase the number of ensemble components, the generalization disparity between teacher and distilled student increases. Improving student fidelity is the most obvious way to close the generalization
disparity between student and teacher in these settings. Even if one exclusively cares about student accuracy, fidelity is a key consideration outside self-distillation.

\textbf{Interpretability and reliability.}\quad
Knowledge distillation has been identified as a means to \emph{transfer representations} discovered by large black-box models
into simpler more interpretable models, for example to provide insights into medical diagnostics, or discovering rules for understanding
sentiment in text \citep[e.g.,][]{hu2016harnessing, hu2016deep, che2015distilling, liu2018improving, chen2020online}. The ability to perform this transfer could have extraordinary scientific 
consequences: large models can often discover structure in data that we would not have anticipated a priori. Moreover, 
we often want to transfer properties such as well-calibrated uncertainties or robustness, which have been well-established for 
larger models, so that we can safely deploy more efficient models in their place. In both cases, achieving good distillation fidelity
is crucial.

\textbf{Understanding.}\quad
The name \emph{knowledge distillation} implies we are transferring knowledge from the teacher to the student. For this reason, improved 
student generalization as a consequence of a distillation procedure is sometimes conflated with fidelity. Decoupling fidelity and generalization,
and explicitly studying fidelity, is foundational to understanding how knowledge distillation works and how we can make it more useful across
a variety of applications.

\begin{figure*}[t]
\centering
\includegraphics[width=0.95\textwidth]{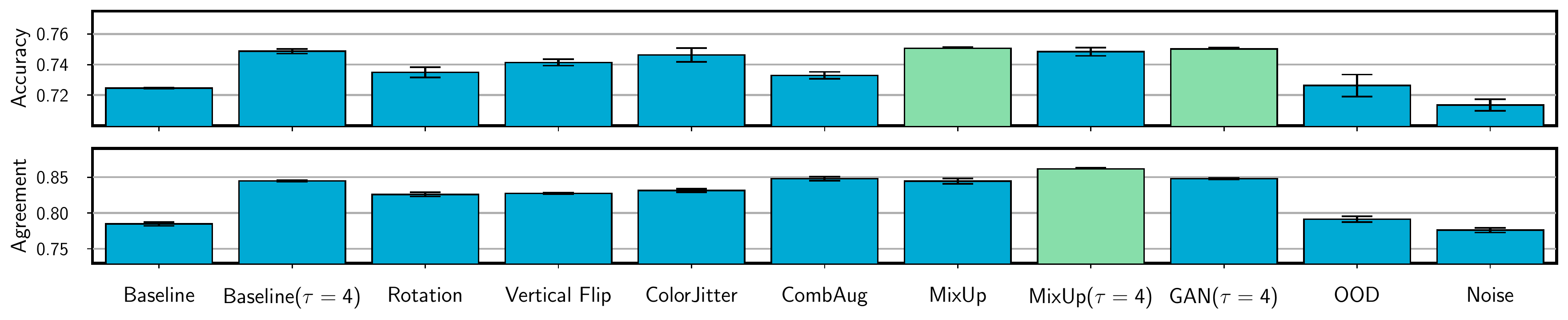}
\caption{
\textbf{Data augmentation and distillation}:
Test accuracy and teacher-student agreement when distilling a 5-component ResNet-56 teacher ensemble into a ResNet-56 student on CIFAR-100 with varying augmentation policies.
The best performing policy is shown in green, results averaged over 3 runs.
Additional metrics are reported in Figure~\ref{supp/fig:detailed_results} in Appendix \ref{supp/sec:additional_ablations}.
Mixup and GAN augmentation provide the best generalization, and Mixup($\tau=4$) provides the best fidelity. The baseline policy (crops and flips) with $\tau=4$ is a surprisingly strong baseline. The error bars indicate $\pm \sigma$.
}
\label{main/fig:augmentation_comparison}
\end{figure*}

\subsection{Possible causes of low distillation fidelity}
\label{main/subsec:possible_causes}

If we are able to match the student model to the teacher on a comprehensive distillation dataset, we expect it to match on the 
test data as well,
achieving high distillation fidelity\footnote{
See, for example, Lemma 1 in \citet{fakoor2020fast}.}.
Possible causes of the poor distillation fidelity in our CIFAR-100 experiments include:

\textbf{Student capacity} -- We observe low fidelity even in the self-distillation setting, so we can rule out student capacity as a primary cause, but we also confirm in Figure \ref{supp/fig:capacity_ablation} in Appendix \ref{supp/subsec:student_capacity_ablation} that increasing the student capacity has very little effect on fidelity in the ensemble-distillation setting. 

\textbf{Network architecture} -- Low fidelity could be specific to ResNet-like architectures, an explanation we rule out by showing similar results with VGG networks \citep{simonyan2014very} in Figure \ref{supp/fig:vgg_distillation_results} in Appendix \ref{supp/subsec:distilling_vgg}.

\textbf{Dataset scale and complexity} -- we provide similar results in Section \ref{supp/subsec:imagenet} for ImageNet, showing that our findings apply to datasets of larger scale and complexity.

\textbf{Data domain} -- Similarly in Section \ref{supp/subsec:imdb} we observe low distillation fidelity in the context of text classification (sentiment analysis on the IMDB dataset),
showing our results are relevant beyond image classification.

\textbf{Identifiability} (Section \ref{main/sec:identifiability}) -- the distillation data is insufficient to distinguish high-fidelity and low-fidelity students.
In other words, matching the teacher predictions on the distillation dataset does not lead to matching predictions on the test data.

\textbf{Optimization} (Section \ref{main/sec:optimization}) -- we are unable to solve the distillation optimization problem sufficiently well. The student does not agree with the teacher on test because it does not even agree on train.

\begin{figure*}
\centering
\begin{subfigure}{0.165\textwidth}
    \includegraphics[width=\textwidth]{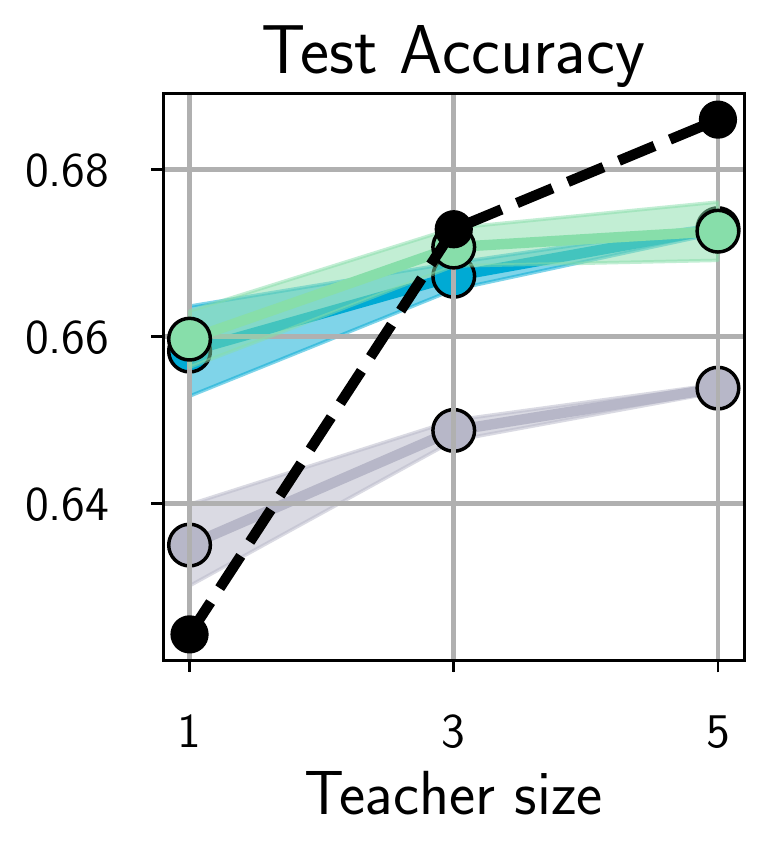}
\end{subfigure}
\hfill
\begin{subfigure}{0.165\textwidth}
    \includegraphics[width=\textwidth]{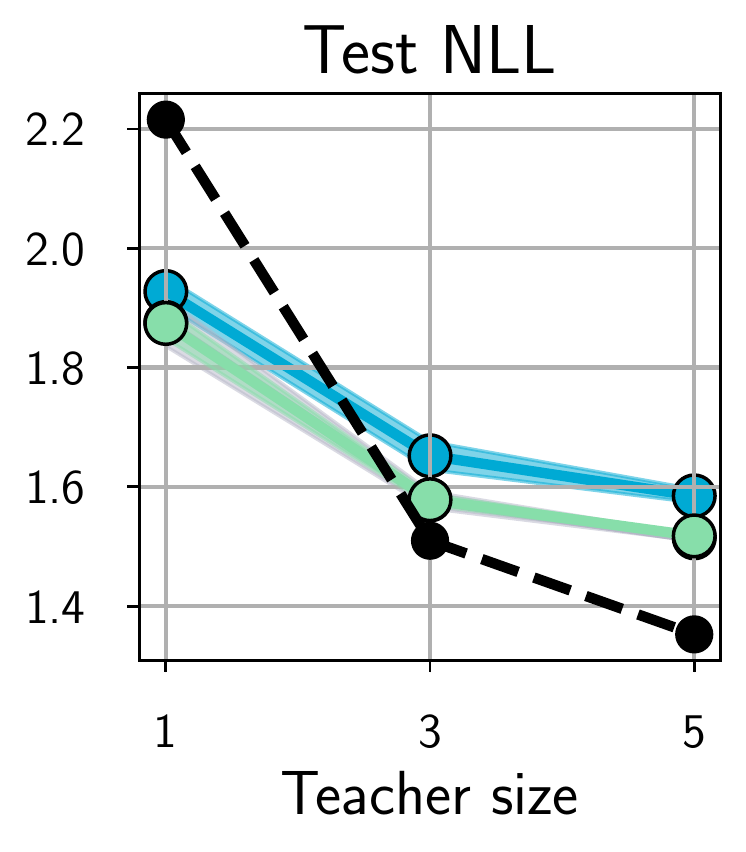}
\end{subfigure}
\hfill
\begin{subfigure}{0.165\textwidth}
    \includegraphics[width=\textwidth]{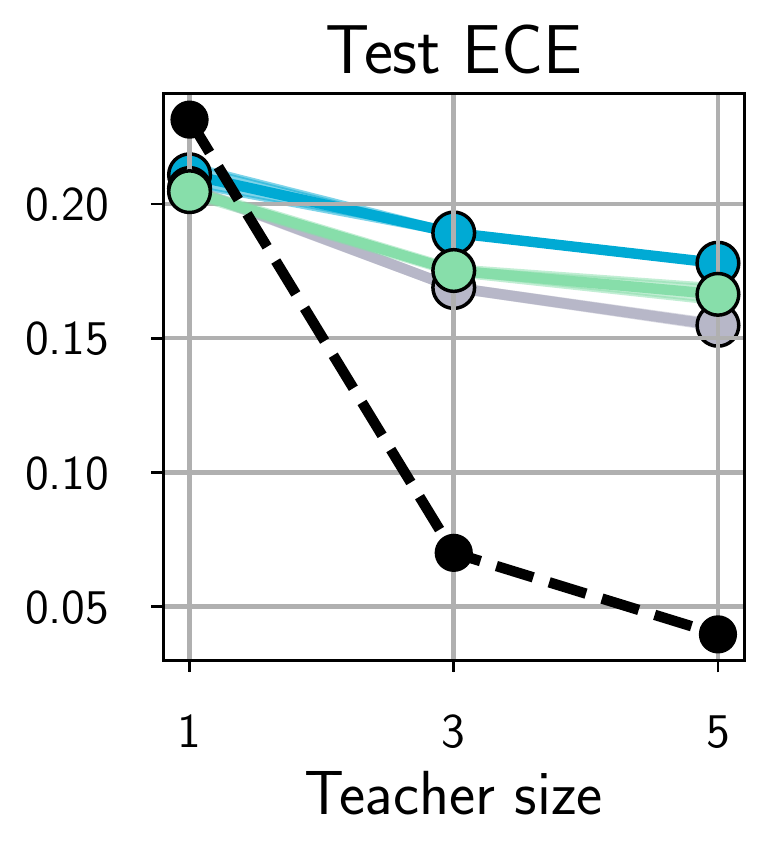}
\end{subfigure}
\hfill
\begin{subfigure}{0.165\textwidth}
    \includegraphics[width=\textwidth]{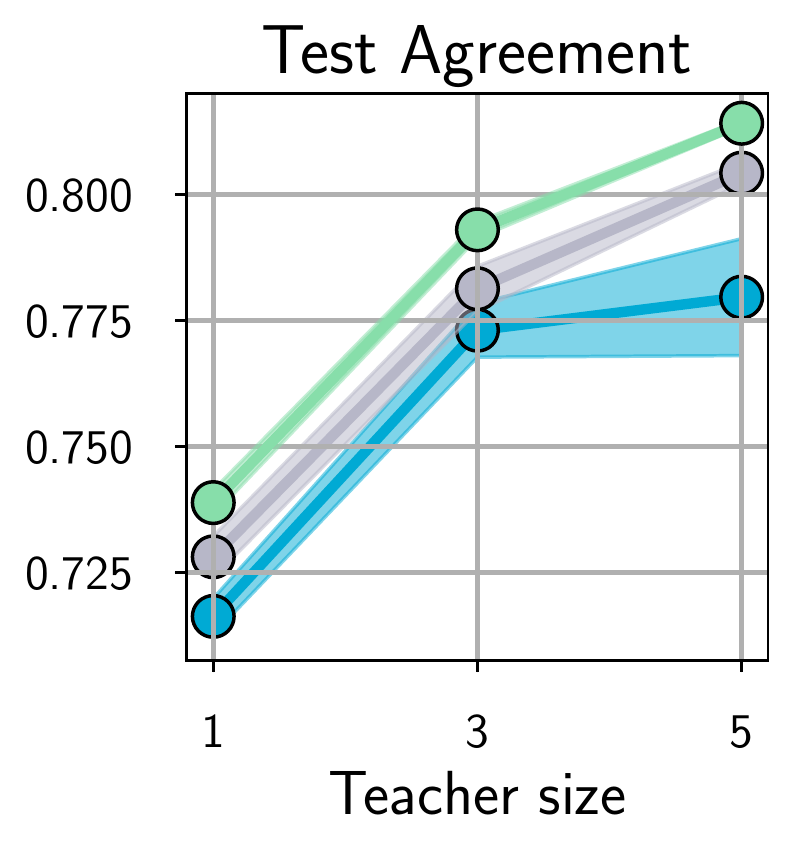}
\end{subfigure}
\hfill
\begin{subfigure}{0.165\textwidth}
    \includegraphics[width=\textwidth]{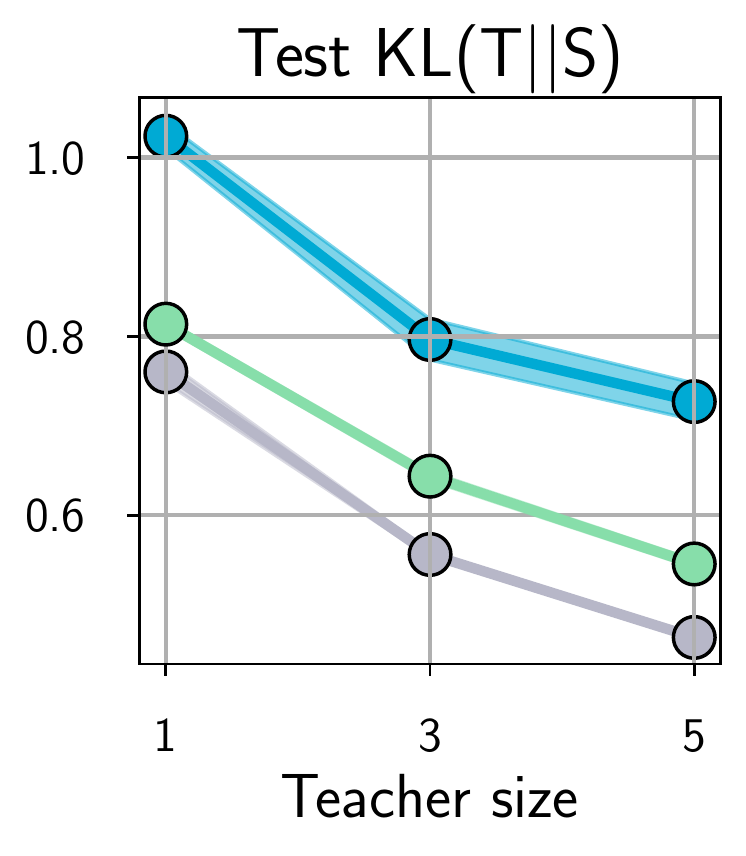}
\end{subfigure}
\begin{subfigure}{0.165\textwidth}
    \includegraphics[width=\textwidth]{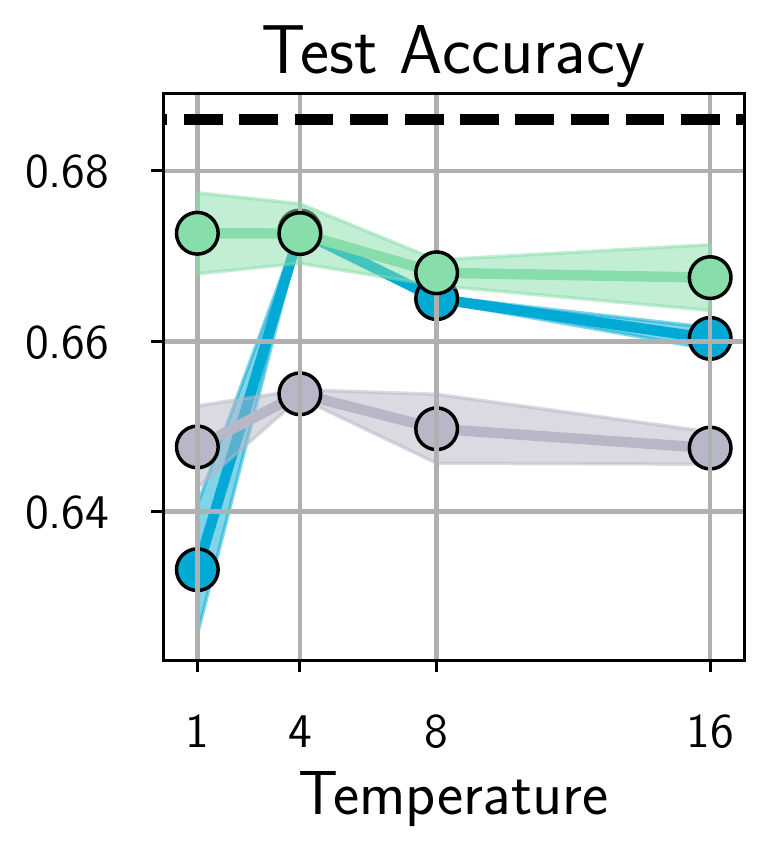}
\end{subfigure}
\hfill
\begin{subfigure}{0.165\textwidth}
    \includegraphics[width=\textwidth]{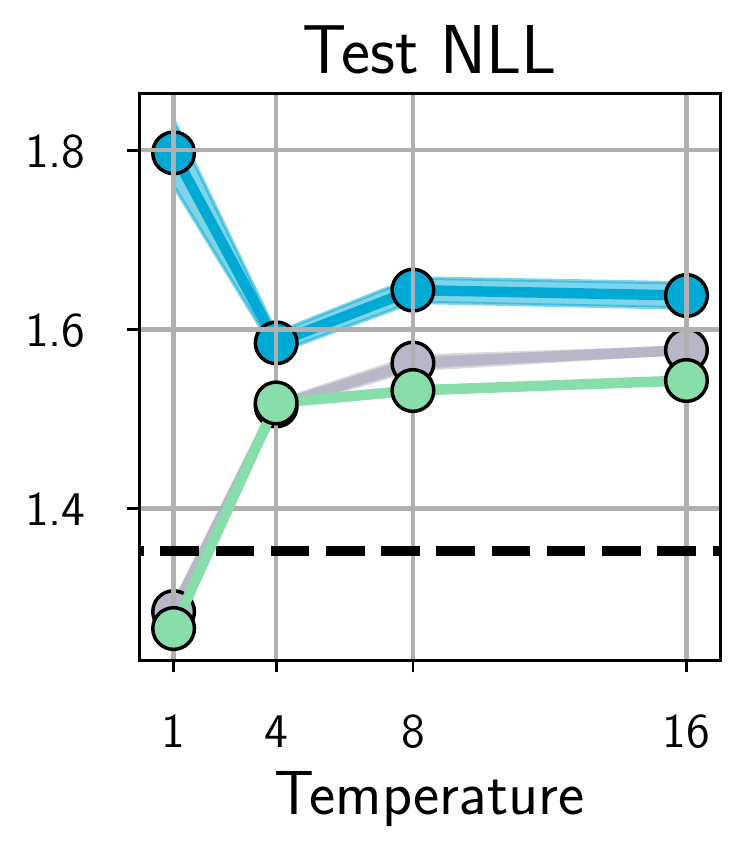}
\end{subfigure}
\hfill
\begin{subfigure}{0.165\textwidth}
    \includegraphics[width=\textwidth]{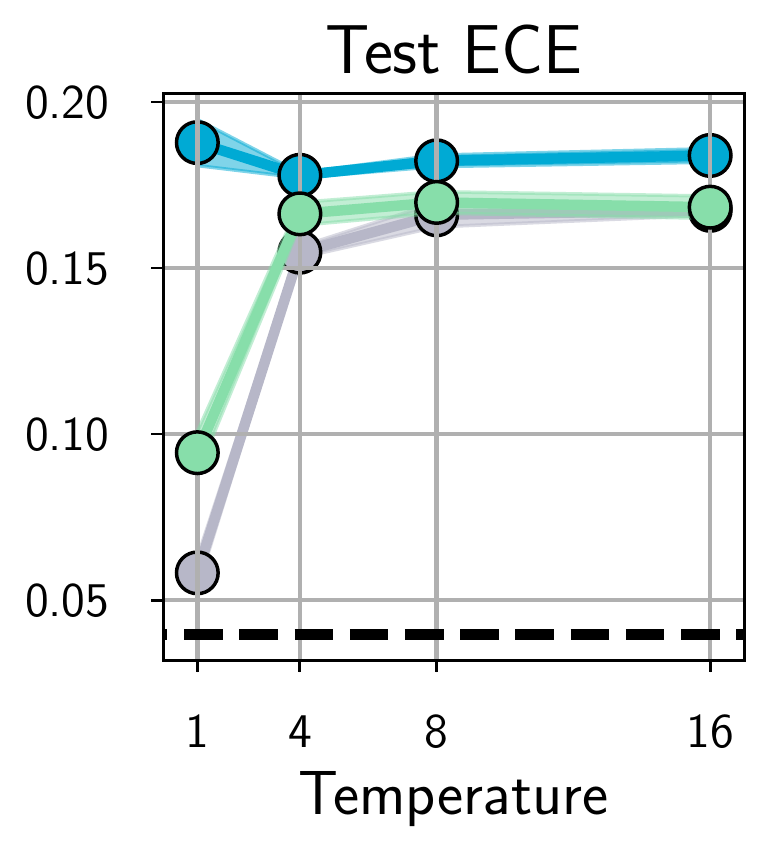}
\end{subfigure}
\hfill
\begin{subfigure}{0.165\textwidth}
    \includegraphics[width=\textwidth]{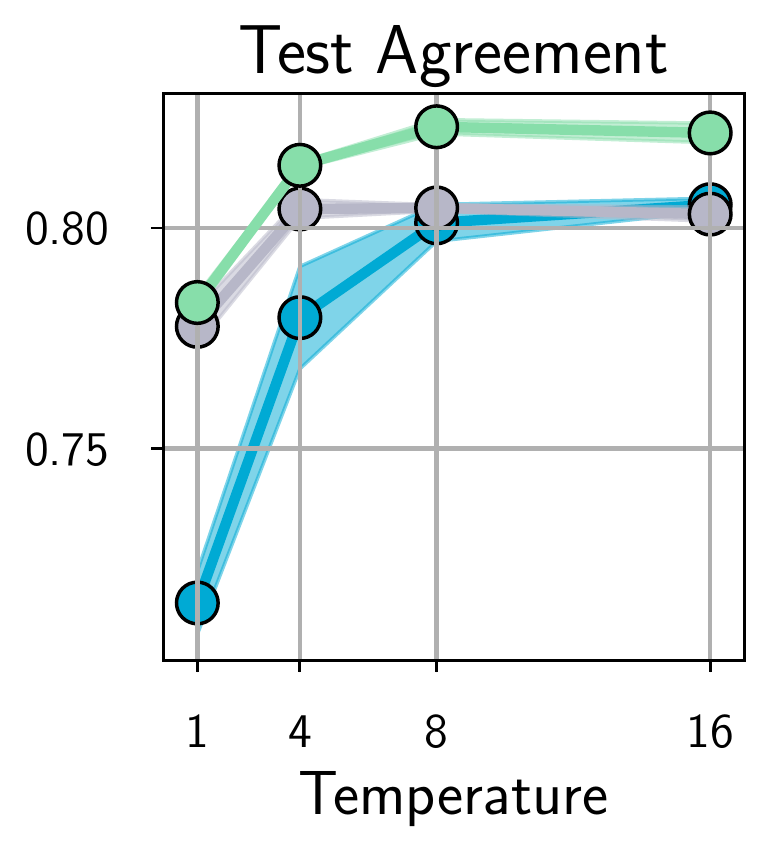}
\end{subfigure}
\hfill
\begin{subfigure}{0.165\textwidth}
    \includegraphics[width=\textwidth]{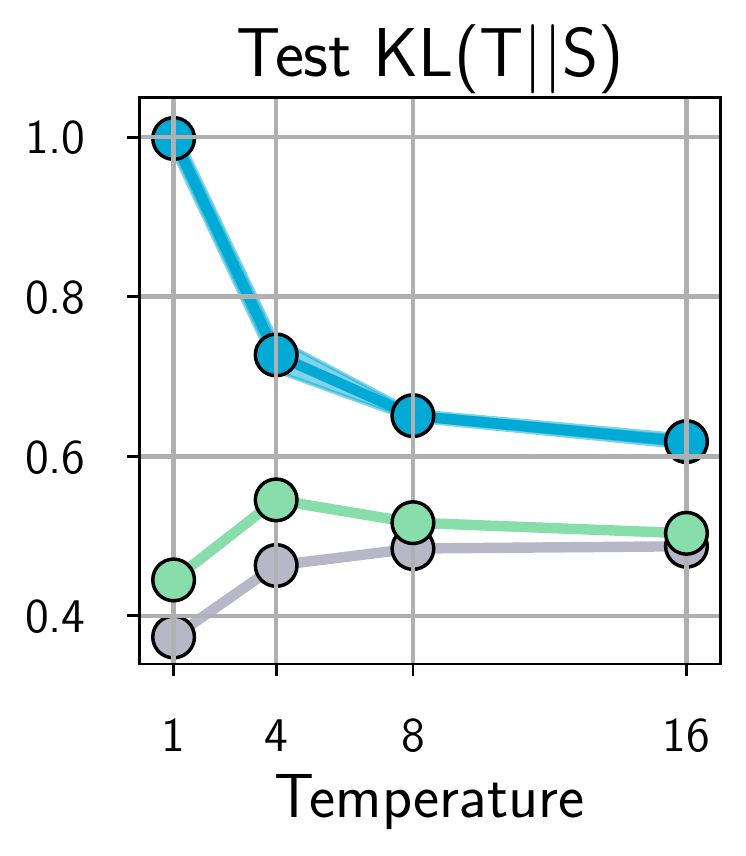}
\end{subfigure}
\begin{subfigure}{0.4\textwidth}
    \includegraphics[width=\textwidth]{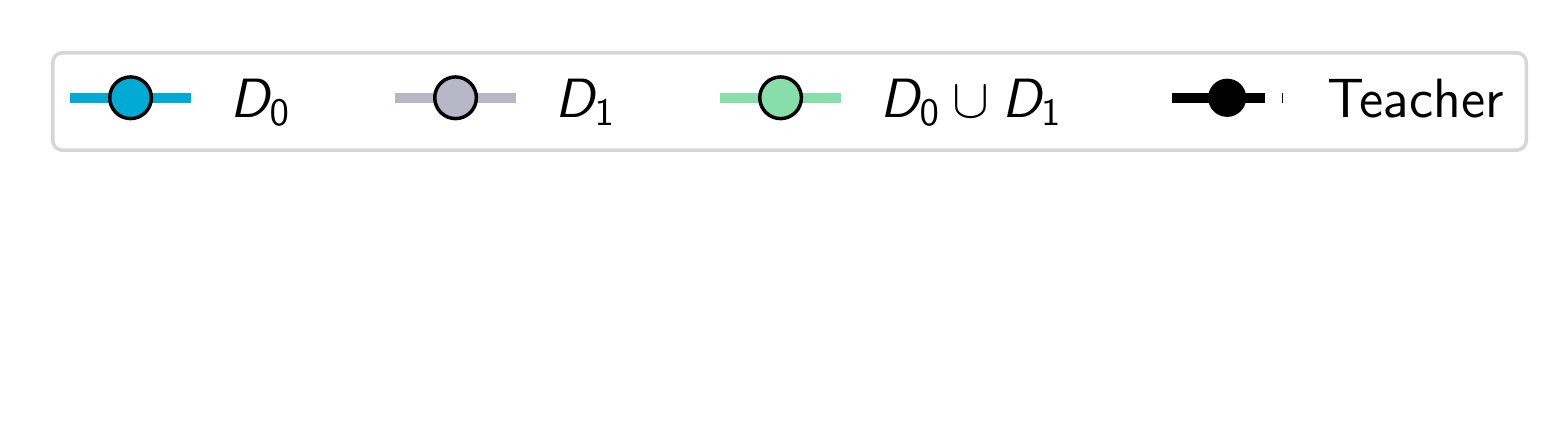}
\end{subfigure}
\vspace{-8mm}
\caption{\textbf{Data recycling and distillation}: results on subsampled CIFAR-100. \textbf{Top:}
We fix the temperature ($\tau=4$) and vary the number of ensemble components ($m$), comparing students distilled on the same dataset as the teacher ($\mathcal{D}_0/\mathcal{D}_0$), a reserved dataset ($\mathcal{D}_0/\mathcal{D}_1$), or both ($\mathcal{D}_0/\mathcal{D}_0 \cup \mathcal{D}_1$). Distilling on both produces the best result, while distilling on $\mathcal{D}_0$ increases accuracy and decreases fidelity, relative to $\mathcal{D}_1$. \textbf{Bottom:} We repeat the experiment, but fix $m=3$ and vary $\tau$. 
The shaded region corresponds to $\mu \pm \sigma$, estimated over 3 trials.
}
\label{main/fig:subsample_split}
\end{figure*}

\section{Identifiability: Are We Using the Right Distillation Dataset?}
\label{main/sec:identifiability}

We investigate whether it is possible to attain the level of fidelity observed with LeNet-5s on MNIST with ResNets on CIFAR-100 by addressing the \textit{identifiability} problem --- have we shown the student enough of the right input-teacher label pairs to define the solution we want?

\subsection{Should we do more data augmentation?}
\label{main/subsec:insufficient_data_hypothesis}

Data augmentation is a simple and practical method to increase the support of the distillation data distribution.
If identifiability is a primary cause of poor distillation fidelity, using a more extensive data augmentation strategy during distillation should improve fidelity.

To test this hypothesis, we evaluated the effect of several augmentation strategies on student fidelity and generalization.
In Figure~\ref{main/fig:augmentation_comparison}, the teacher is a 5-component ensemble of ResNet-$56$ networks trained on CIFAR-100 with the \textit{Baseline} augmentation strategy: horizontal flips and random crops.
We report the student accuracy and teacher-student agreement for each augmentation strategy, and also include results for \emph{Baseline} with $\tau=1$ and $\tau=4$ to demonstrate the effect of logit tempering.

We first observe that the best augmentation policies for generalization, \textit{MixUp}, and \textit{GAN}\footnote{Unlike Figure \ref{main/fig:cifar100_motivation}, for Figure \ref{main/fig:augmentation_comparison} we generated new GAN samples every epoch, to mimic data augmentation.}, are not the best policies for fidelity.
Furthermore, although many augmentation strategies enable slightly higher distillation fidelity compared to \textit{Baseline ($\tau=1$)}, even the best augmentation policy, \textit{Mixup} ($\tau=4$), only achieves a modest 86\% test agreement.
In fact the \textit{Baseline ($\tau=4$)} policy is quite competitive, achieving $84.5\%$ test agreement. 
Many of the augmentation strategies also slightly improve teacher-student KL relative to \textit{Baseline ($\tau=4$)} (see Figure \ref{supp/fig:detailed_results}).

In Figure \ref{supp/fig:detailed_results} in Appendix \ref{supp/subsec:detailed_aug_results} we report all generalization and fidelity metrics for a range of ensemble sizes, as well as the results for the independent student baseline discussed in Section \ref{main/subsec:define_metrics}.
Often these independent students, taught how to mimic a completely different model,
have nearly as good test agreement with the teacher as the student explicitly trained to emulate it.
See Appendix~\ref{supp/subsec:aug_procedures} for a detailed description of the augmentation procedures.

\textbf{Should data augmentation be close to the data distribution?}\quad
In theory, \textit{any} data augmentation should help with identifiability: if a student matches a teacher on more data, it is more likely to 
match the teacher elsewhere.
However, the \textit{Noise} and \textit{OOD} augmentation strategies based on noise and out-of-distribution data fail on all metrics, decreasing performance compared to the baseline.
In practice, data augmentation has an effect beyond improving identifiability --- it has a regularizing effect, making optimization more challenging.
We explore this facet of data augmentation in Section~\ref{main/sec:optimization}.

The slight improvements to fidelity with extensive augmentations suggest that increasing the support of the distillation dataset can indeed improve distillation fidelity.
However, since the benefit is so small compared to heuristics like logit tempering (which does not modify the support at all), it is very unlikely that an insufficient quantity of teacher labels is the primary obstacle to high fidelity.

\subsection{The data recycling hypothesis}
\label{main/subsec:recycling_hypothesis}

If simply showing the student \textit{more} labels does not always significantly improve fidelity, perhaps we are not showing the student the \textit{right} labels. 
Additional data augmentation during distillation does give the student more teacher labels to match, but also introduces a distribution shift between the images the teacher was trained on and the images the student is distilling on. 
Even when the teacher and student have the same augmentation policy, reusing the teacher's training data for distillation violates the assumptions of empirical risk minimization (ERM) because the distillation data is \emph{not} an independent draw from the true joint distribution over images and teacher labels.
What if there was no augmentation distribution shift, and the student was distilled on a fresh draw from the joint test distribution over images and teacher labels?

To investigate the effect of recycling teacher data during distillation we randomly split the CIFAR-100 training dataset $\mathcal{D}$ into two equal parts, $\mathcal{D}_0$ and $\mathcal{D}_1$. 
We train teacher ResNet-56 ensembles on $\mathcal{D}_0$, and then compare $s_0$, a student  distilled on the original $\mathcal{D}_0$, $s_1$, a student distilled on the unseen $\mathcal{D}_1$, and $s_{0\cup 1}$, a student distilled on both: $\mathcal{D}_0 \cup \mathcal{D}_1$. Note that the students cannot access the true labels, only those provided by the teacher.
We present the results in Figure \ref{main/fig:subsample_split}, varying the ensemble size in the top row and the logit temperature in the bottom row.

Surprisingly, $s_0$ attains higher test accuracy than $s_1$, while showing worse ECE and lower fidelity (measured by test teacher-student agreement and test teacher-student KL). 
Therefore, the hypothesis that $s_1$ should be a higher fidelity distillation of the teacher than $s_0$ does hold, but the gain in fidelity \textit{does not} result in $s_1$ best replicating the teacher's accuracy. 
The best attributes of $s_0$ and $s_1$ are combined by $s_{0\cup 1}$, which coincides with how unlabeled data is typically used in practice \citep{ba2014deep}. The reason for this puzzling observation is simply that for the larger teachers fidelity has not improved \textit{enough} to also improve generalization.
In fact, the best teacher-student agreement is only around $85\%$, no improvement when compared to the results from extensive data augmentation in the last section.
We again find that modifying the distillation data can slightly improve fidelity, but the evidence does not support blaming poor distillation fidelity on the wrong choice of distillation data.

\section{Optimization: Does the Student Match the Teacher on Distillation Data?}
\label{main/sec:optimization}

If poor fidelity is not primarily an identifiability problem from the wrong choice of distillation data, perhaps there is a simpler explanation.
Up to this point, we have focused on student fidelity on a held-out test set. 
Now we turn our attention to student behavior on the distillation data itself.  Does the student match the teacher on the data it is trained to match it on?

\subsection{More distillation data lowers train agreement}

\begin{figure}
\centering
\begin{tabular}{ccccc}
    \hspace{-0.3cm}\includegraphics[height=0.14\textwidth]{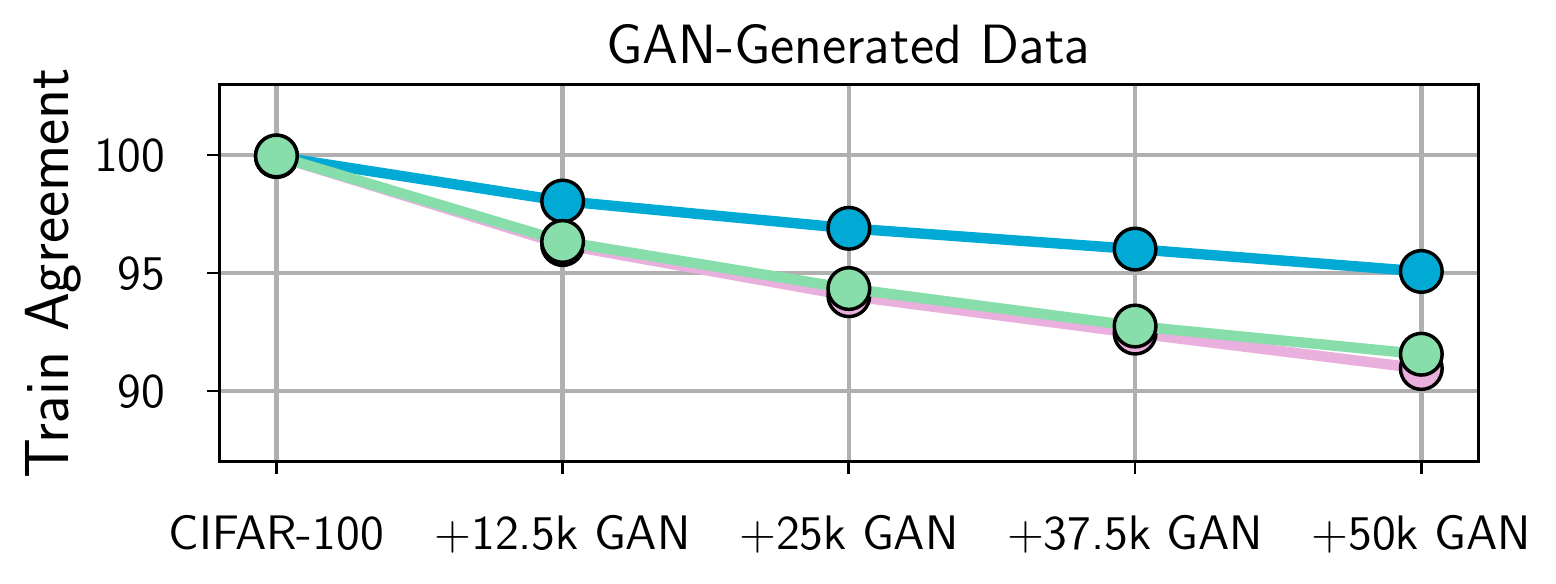}
    &
        \hspace{-0.4cm}\includegraphics[height=0.14\textwidth]{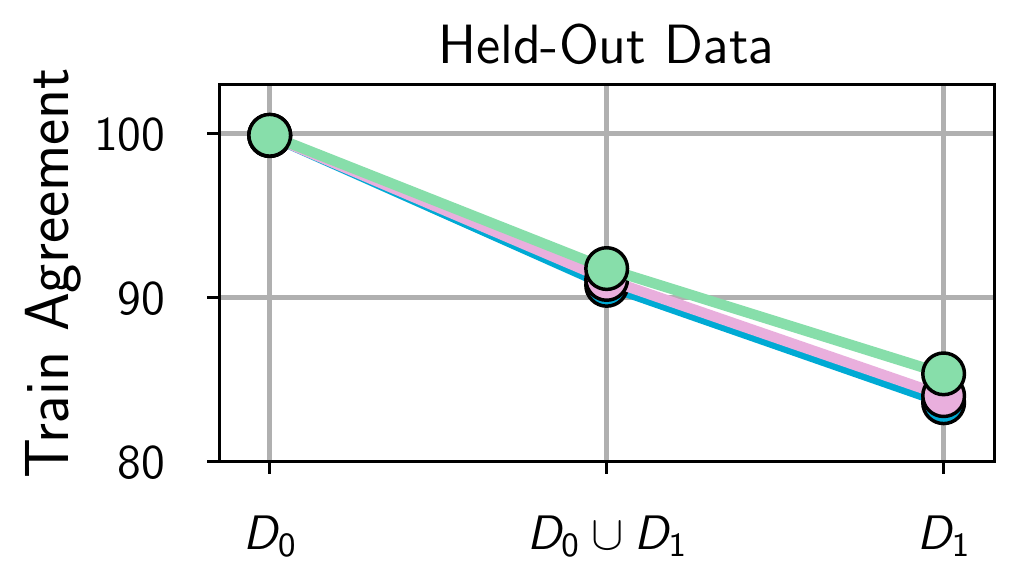}
    &
    \hspace{-0.3cm}\includegraphics[height=0.14\textwidth]{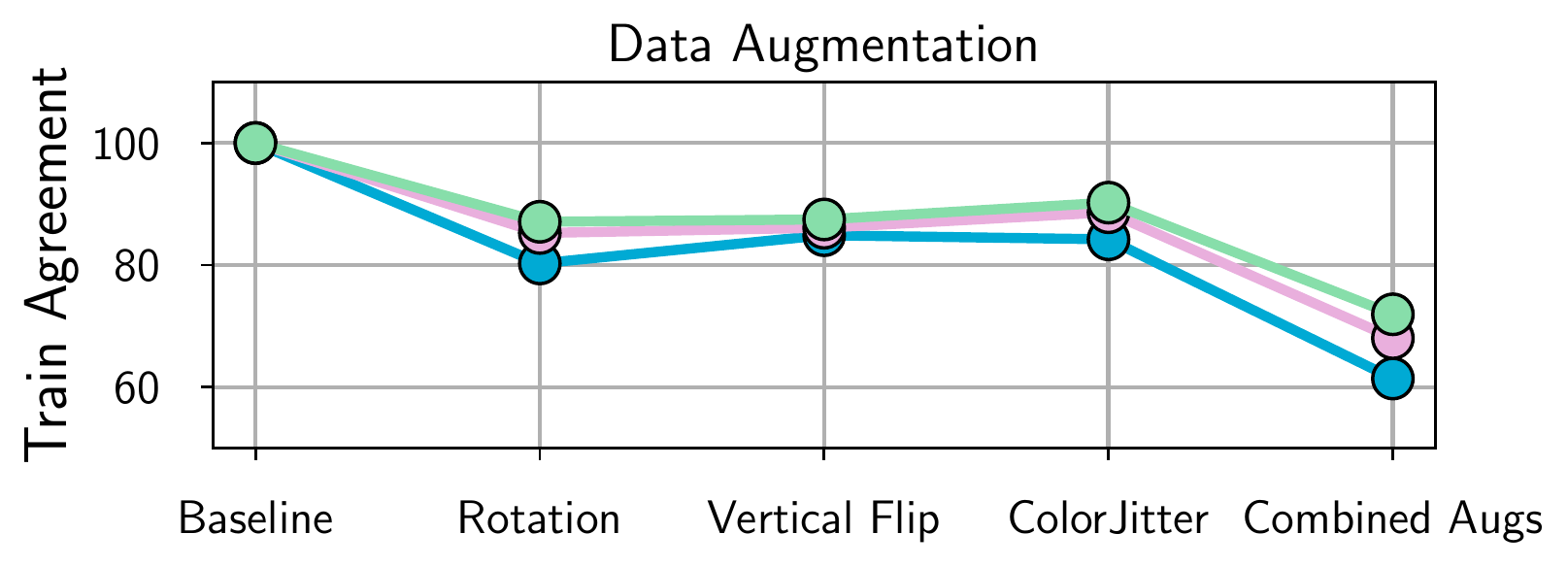}
    \\[-0.2cm]
    \multicolumn{3}{c}{
      \includegraphics[width=0.4\textwidth, trim=0cm 2.2cm 0.cm 0cm, clip]{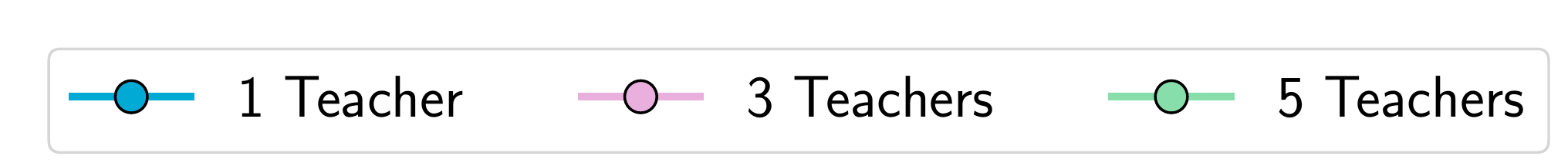}
    }
\end{tabular}
\caption{
The train agreement for teacher ensembles ($m \in \{1, 3, 5\}$) and student on the distillation data for a ResNet-56 on CIFAR-100 under different augmentation policies. In all panels, increasing the softness of the teacher labels by adding examples not in the teacher train data makes distillation more difficult.
\textbf{Left}: agreement for the synthetic GAN-augmentation policy from Figure \ref{main/fig:cifar100_motivation}.
\textbf{Middle}: agreement from subsampled CIFAR-100 experiment in Figure \ref{main/fig:subsample_split}.
\textbf{Right}: agreement for some of the augmentation policies in Figure \ref{main/fig:augmentation_comparison}.
The shaded region is not visible because the variance is very low.
}
\label{main/fig:optimization_problem}
\end{figure}

In Figure \ref{main/fig:cifar100_motivation} we presented an experiment distilling ResNet-56 networks on CIFAR-100 augmented with synthetic GAN-generated images.
We saw that enlarging the distillation dataset leads to improved teacher-student agreement on test, but the agreement remains relatively low (below $80\%$) even for the largest distillation dataset that we considered.
In Figure~\ref{main/fig:optimization_problem} (left panel), we report the teacher-student agreement for the same experiment,
but now on the distillation dataset.
We now observe the opposite trend: as the distillation dataset becomes larger, it becomes more challenging for the student to match the teacher.
Even when the student has identical capacity to the teacher, the student only achieves $95\%$ agreement with the teacher when we use $50k$ synthetic images for distillation.

The drop in train agreement is even more pronounced when we use extensive data augmentation.
In Figure~\ref{main/fig:optimization_problem}, right panel, we report the teacher-student agreement on the train set
with data augmentation for a subset of augmentation strategies presented in Section~\ref{main/subsec:insufficient_data_hypothesis}.
We use the CIFAR-100 dataset and the ResNet-56 model for the teachers and the students (for details, see Section~\ref{main/subsec:insufficient_data_hypothesis}).
In each case, we measure agreement on the augmented training set that was used during distillation.
While for the baseline augmentation strategy, we can achieve almost perfect teacher-student agreement, for heavier augmentations
the agreement drops dramatically.
For the \textit{Rotation}, \textit{Vertical Flip} and \textit{Color Jitter} augmentations, the agreement is between
$80\%$ and $90\%$ for all the considered teacher sizes.
For \textit{Combined Augs}, the combination of these three augmentation strategies, the agreement drops even further, to just $60\%$ in self-distillation!

Our intuition about how knowledge distillation should work largely hinges on the assumption that after distillation
the student matches the teacher on the distillation set.
However, the results presented in this section suggest that in practice the optimization method is unable to achieve
high fidelity \emph{even on the distillation dataset} when extensive data augmentation or synthetic data is used.
The inability to solve the optimization problem undermines distillation: in order to find a student that would match the teacher
on all inputs, we need to at least be able to find a student that would match the teacher on all of the distillation data.

\textbf{Optimization and the train-test fidelity gap.}\quad
Notably, despite having the lowest train agreement, the \textit{Combined Augs} policy results in better test agreement than other polices with better train agreement (Figure \ref{main/fig:augmentation_comparison}).
This result highlights a fundamental trade-off in knowledge distillation:
the student needs many teacher labels match the teacher on test,
but introducing examples not in the teacher train data makes matching the teacher on the distillation data very difficult.

\subsection{Why is train agreement so low?}

\textbf{A simplified distillation experiment.}
To simplify our exploration, we focus on self-distillation of a ResNet-20 on CIFAR-100.
We use the \textit{Baseline} data augmentation strategy, as we found that a ResNet-20 student is unable to match the teacher on train even with
basic augmentation.
We also replace the BatchNorm layers \citep{ioffe2015batch} in ResNet-20 with LayerNorm \citep{ba2016layer}, because
we found that with BatchNorm layers even when the teacher and the student have identical weights, they can make different predictions due to differences in the activation statistics accumulated by the BatchNorm layers.
Layer normalization does not collect any activation statistics, so the student will match the teacher as long as the weights coincide.

\begin{figure}
\centering
\begin{tabular}{ccccc}
    \hspace{-0.4cm}\includegraphics[height=0.17\textwidth]{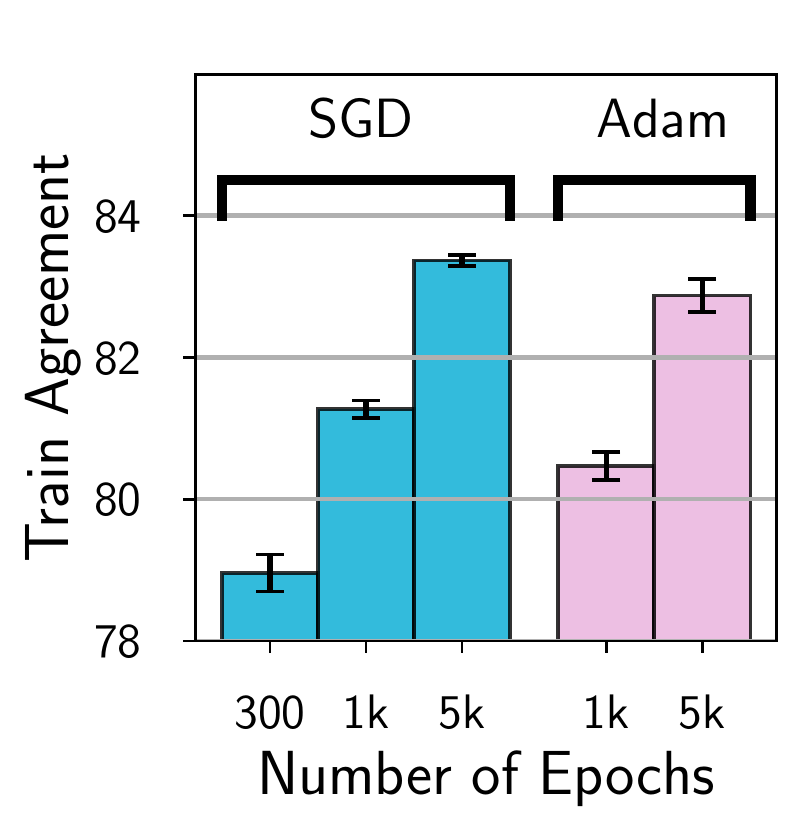}
    &
    \hspace{-0.cm}\includegraphics[height=0.17\textwidth]{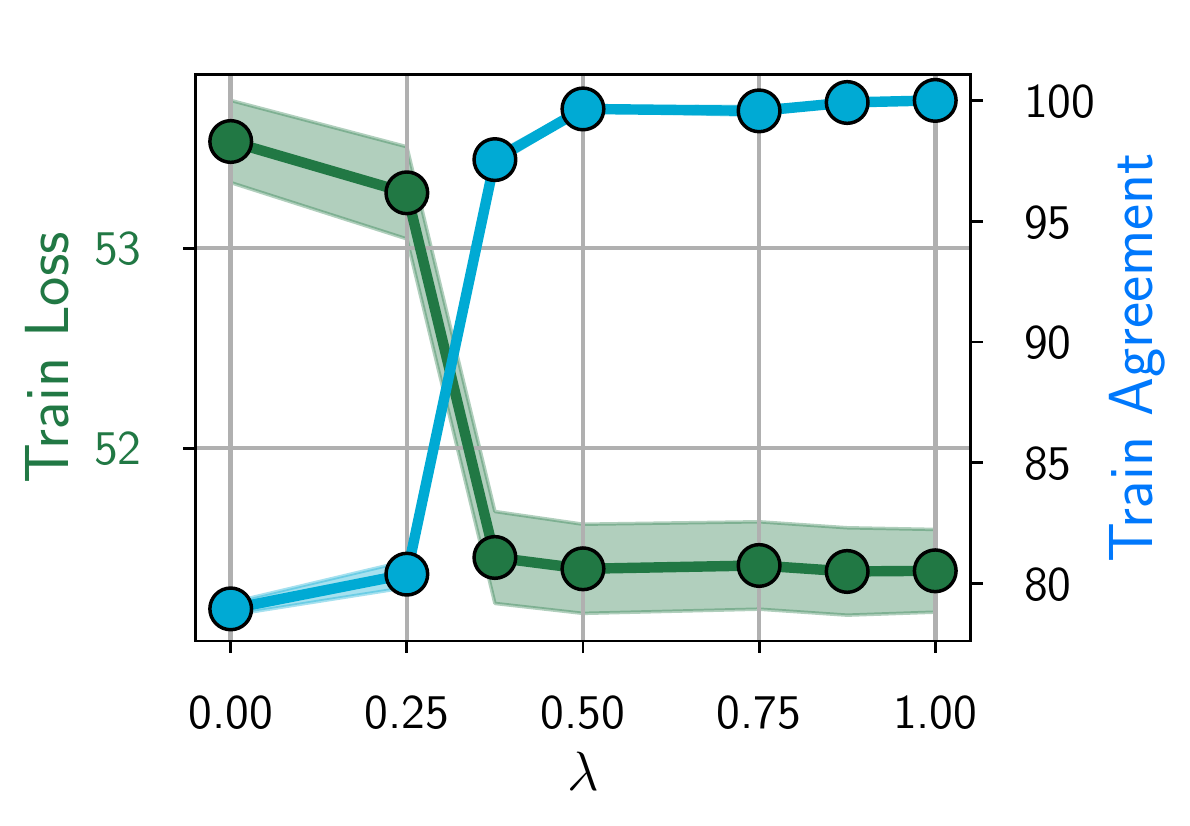}
    &
    \hspace{-0.2cm}\includegraphics[height=0.17\textwidth]{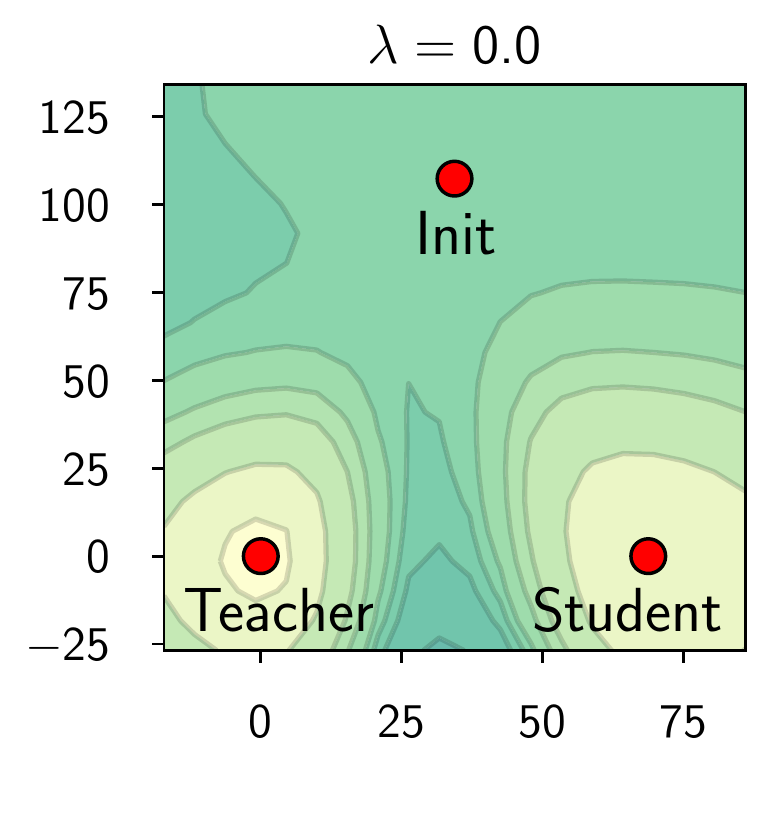}
    &
    \hspace{-0.4cm}\includegraphics[height=0.17\textwidth]{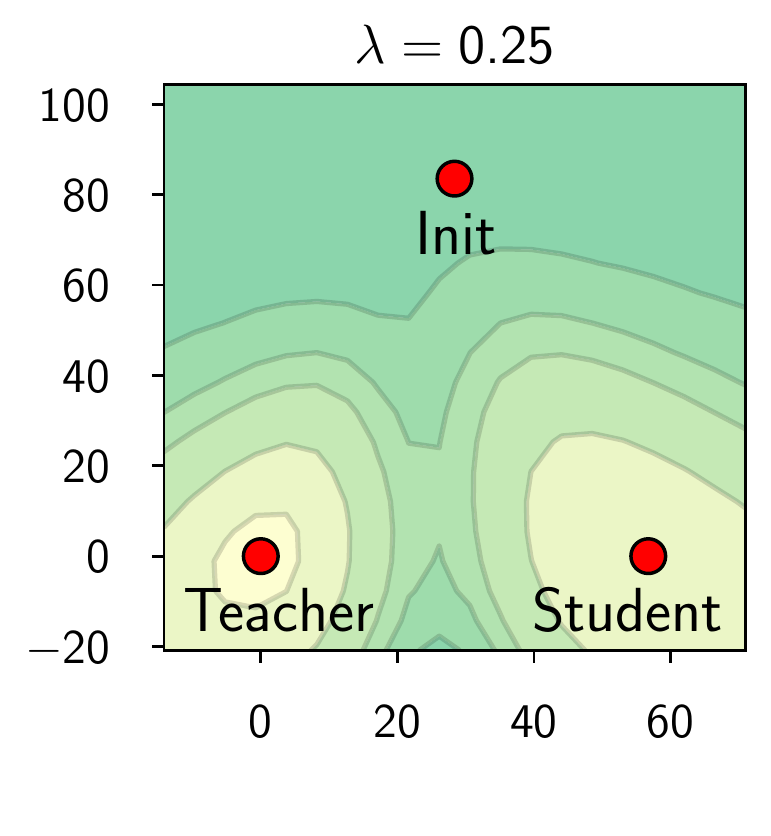}
    &
    \hspace{-0.4cm}\includegraphics[height=0.17\textwidth]{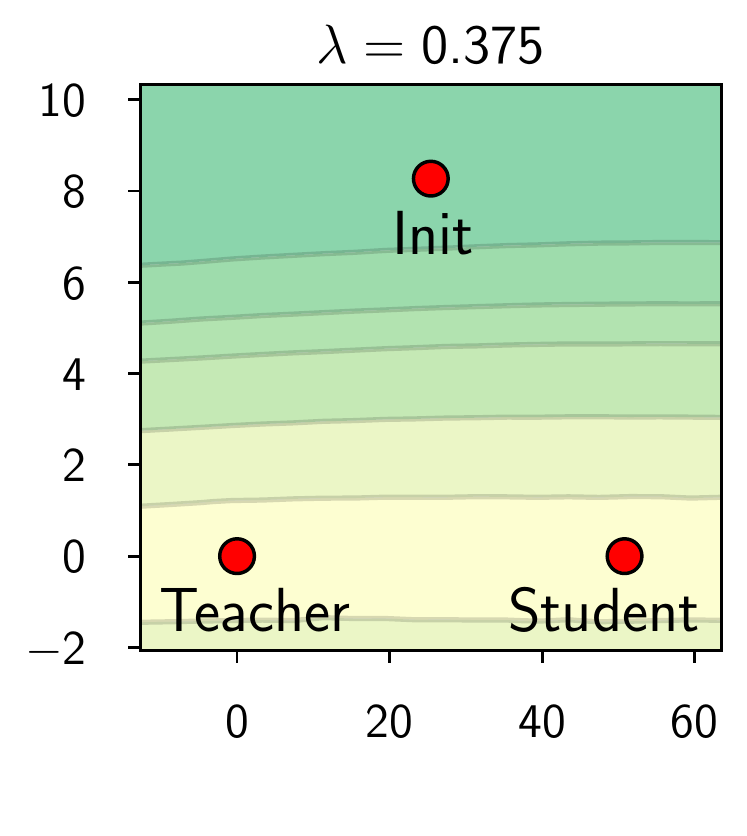}
    \includegraphics[height=0.16\textwidth, trim=0cm -1cm 0.cm 0cm, clip]{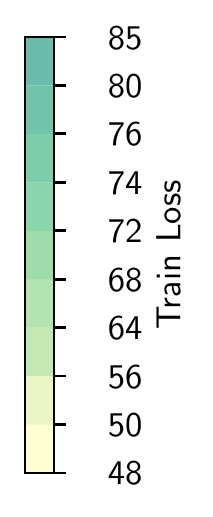}
    \\[-0.1cm]
    \hspace{-0.4cm}\small(a) Optimizer effect &
    \small(b) Initialization effect & 
    \multicolumn{3}{c}{\small(c) Loss Visualization}
\end{tabular}
\caption{
  \textbf{Optimization and distillation}: self-distillation with ResNet-20s with LayerNorm on CIFAR-100. 
  \textbf{(a)}: Final train agreement for SGD and Adam optimizers. Training longer improves agreement, but it remains below $85\%$ even after $5k$ epochs.
  \textbf{(b)}: Final train loss and agreement when the initialization is a convex combination of teacher and random weights, $\theta_s = \lambda \theta_t + (1 - \lambda) \theta_r$.
  \textbf{(c)}: Projections of the distillation loss surface on the plane intersecting $\theta_t$, the initial student weights, and the final student weights for different $\lambda$.
  When $\lambda$ is small, the student converges to a suboptimal solution with low agreement.
  The uncertainty regions correspond to $\mu \pm \sigma$, estimated over 3 trials.
}
\label{main/fig:init_ablation}
\end{figure}

\textbf{Can we solve the optimization problem better?}
We verify that the distillation fidelity cannot be significantly improved by training longer or with a different optimizer.
By default, in our experiments we use stochastic gradient descent (SGD) with momentum, train the student for $300$ epochs, and use a weight decay value of $10^{-4}$.
In Figure~\ref{main/fig:init_ablation} we report the results for the SGD and Adam \citep{kingma2014adam} optimizers run for $1k$ and $5k$ epochs without weight decay. Switching from SGD to Adam only reduced fidelity.

For both optimizers, training for more epochs does slightly improve train agreement.
In particular, with SGD we achieve $83.3\%$ agreement when training for $5k$ epochs compared to $78.95\%$ when training for $300$ epochs.
It is possible, though unlikely, that if we train for even more epochs the train agreement could reach $100\%$.
However, training for $5k$ epochs is significantly longer than what is typically done in practice ($100$ to $500$ epochs).
Furthermore, the improvement from $1k$ to $5k$ epochs is only about $2\%$, suggesting that we would need to train for tens of thousands of epochs, even in the optimistic case that agreement improves linearly,
in order to get close to $100\%$ train agreement.

\textbf{The distillation loss surface hypothesis}: 
If we cannot perfectly distill a ResNet-20 on CIFAR-100 with any of the interventions we have discussed so far, we now ask if there is any modification of the problem that \textit{can} produce a high-fidelity student. 

In the self-distillation setting, we do know of at least one set of weights that is optimal w.r.t. the distillation loss --- the teacher's own weights $\theta_t$. 
Letting $\theta_r$ be a random weight initialization, in Figure \ref{main/fig:init_ablation} (a) we examine the effect of choosing the student initialization to be a convex combination of the teacher and random weights, $\theta_s = \lambda \theta_t + (1- \lambda) \theta_r$. 
After being initialized in this way, the student was trained as before. 
In other words $\lambda = 0$ corresponds to a random initialization and $\lambda = 1$ corresponds to initializing the student weights at the final teacher weights.

We find that if the student is initialized far from the teacher ($\lambda \leq 0.25$), the optimizer converges to a
sub-optimal value of the distillation loss, producing a student that significantly disagrees with the teacher.
However at $\lambda=0.375$ there is a sudden change. The final train loss drops to the optimal value and the agreement drastically increases, and the behavior continues for $\lambda > 0.375$.
To further investigate, in Figure \ref{main/fig:init_ablation} (c) we visualize the distillation loss surface for $\lambda \in \{0, 0.25, 0.375\}$ projected on the 2D subspace intersecting $\theta_t$, the initial student weights, and the final student weights.
If the student is initialized far from the teacher ($\lambda \in \{0, 0.25\}$), it converges to a distinct, sub-optimal basin of the loss surface.
On the other hand, when initialized close to the teacher ($\lambda = 0.375$), the student converges to the same basin as the teacher, achieving nearly 100\% agreement. 

\textbf{Is using the initial teacher weights enough for good fidelity?}
If good fidelity can be obtained by initializing the student near the \textit{final} teacher weights, it is possible that similar results could be obtained by initializing the student at the \textit{initial} teacher weights.
In Table \ref{main/tab:initialization_ablation} we compare students distilled from random initializations with those initialized at the initial teacher weights.
In addition to the metrics reported in the rest of the paper, we also include the centered kernel alignment (CKA) \citep{kornblith2019similarity} of the preactivations of each of the teacher and student networks. 
There is a small increase in CKA, indicating that sharing an initialization between teacher and student does increase alignment in activation space, but functionally the students are identical to their randomly initialized counterparts -- there is no observable change in accuracy, agreement, or predictive KL when compared to random initialization.

\begin{table*} 
\centering
\begin{tabular}{lccccc} 
\toprule 
& & & \multicolumn{3}{c}{\textbf{CKA} $(\uparrow)$} \\
\cmidrule(l){4-6}
\textbf{Init.} & \textbf{Agree. $(\uparrow)$} & \textbf{KL $(\downarrow)$} & \textbf{Stage 1} & \textbf{Stage 2} & \textbf{Stage 3}\\ 
\midrule 
Rand. & 77.174 (0.352) & 0.836 (0.016) & 0.939 (0.017) & 0.925 (0.027) & 0.885 (0.011)\\ 

Teach. & 77.098 (0.238) & 0.838 (0.020) & 0.951 (0.017) & 0.937 (0.020) & 0.890 (0.015)\\ 
\bottomrule 
\end{tabular}
\caption{We examine whether fidelity can be improved in the context of ResNet-20 self-distillation on CIFAR-100 if the teacher and student share the same weight initialization. 
All metrics are computed on the test set.
A shared initialization does make the student slightly more similar to the teacher in activation space (measured by CKA), but in function space the results are indistinguishable from randomly initialized students.
We report the mean and standard deviation, estimated from 10 trials. The average teacher accuracy was 70.522 (0.412).
} 
\label{main/tab:initialization_ablation} 
\end{table*}

To summarize, we have at last identified a root cause of the ineffectiveness of all our previous interventions on the knowledge distillation procedure. 
Knowledge distillation is unable to converge to optimal student parameters, even when we know a solution and give the initialization a small head start in the direction of an optimum.
Indeed, while identifiability can be an issue, in order to match the teacher on all inputs, the student has to at least match the teacher on the data used for distillation,
and achieve a near-optimal value of the distillation loss.
Furthermore, the suboptimal convergence of knowledge distillation appears to be a consequence of the optimization dynamics specifically, and not simply initialization bias.
In practice, optimization converges to sub-optimal solutions, leading to poor distillation fidelity.

\section{Discussion}
\label{main/sec:discussion}

Our work provides several new key findings about knowledge distillation:
\begin{itemize}
\item \emph{Good student accuracy does not imply good distillation fidelity:} even outside of self-distillation, the models with the best generalization do not always achieve the best fidelity.
\item \emph{Student fidelity is correlated with calibration when distilling ensembles:} although the highest-fidelity student is not always the most accurate, it is always the best calibrated.
\item \emph{Optimization is challenging in knowledge distillation:} even in cases when the student has sufficient capacity to match the teacher on the distillation data, it is unable to do so.
\item \emph{There is a trade-off between optimization complexity and distillation data quality:} Enlarging the distillation dataset beyond the teacher training data makes it easier for the student to identify the correct solution, but also makes an already difficult optimization problem harder.
\end{itemize}

In standard deep learning, we are saved by not needing to solve the optimization problem well: while it true that our training loss is highly multimodal, properties such as the flatness of good solutions, the inductive biases of the network, and the implicit biases of SGD, often enable good generalization in practice. In knowledge distillation, however, good fidelity is directly aligned with solving what turns out to be an exceptionally difficult optimization problem. 

\section*{Acknowledgements}
The authors would like to thank Gregory Benton, Marc Finzi, Sanae Lotfi, Nate Gruver, and Ben Poole for helpful feedback.
This  research  is  supported by  an  Amazon  Research  Award, NSF  I-DISRE  193471,  NIH  R01DA048764-01A1,  NSF IIS-1910266,  and NSF 1922658NRT-HDR: FUTURE Foundations,  Translation,  and Responsibility for Data Science. Samuel Stanton is also supported by a United States Department of Defense NDSEG fellowship.

\bibliography{references}
\bibliographystyle{apalike}

\appendix

\section*{Appendix}

\textbf{Supplement Outline}:
\begin{itemize}
    \item[A.] Implementation details for all experiments.
    \item[B.] Additional experiments with the teacher ensemble size ablation.
    \item[C.] Experiments addressing spurious explanations for poor student fidelity.
\end{itemize}

\section{Implementation details}
\label{supp/sec:implementation}

Here we briefly describe key implementation details to reproduce our experiments.  Data augmentation details are given in \ref{supp/subsec:aug_procedures}, followed by architecture details in \ref{supp/subsec:architectures}, and finally
training details are provided in \ref{supp/subsec:training}. The reader is encouraged to consult the included code for closer inspection.

\subsection{Data augmentation procedures}
\label{supp/subsec:aug_procedures}

Some of the data augmentation procedures we consider attempt to generate data that is close to the train data distribution (standard augmentations, GAN, mixup). Others (random noise, out-of-domain data) produce data for distillation that the teacher would never encounter during normal supervised training. In particular, we compare the following augmentation procedures:

\textbf{Baseline augmentations} \quad As a baseline, we use the same data augmentation strategy that was used to train the teachers during distillation: we apply random horizontal flips ($p=0.5$) and random shifts via pad and random-crop with a 4 pixel pad width.
In all of the configurations we consider in this section we use this set of augmentations along with other strategies, unless stated otherwise.

\textbf{Conventional image transformations} \quad Standard data augmentations used in computer vision \citep{marcel2010torchvision}: random rotations by up to $20$ degrees,  random vertical flips, color jitter and all possible combinations.

\textbf{Mixup} \quad Mixup is an effective regularization technique originally proposed to increase generalization and robustness of deep networks \citep{zhang2017mixup, tokozume2018between}.
Instead of training on original dataset, the network is trained on convex combination of images with targets mixed in the same way.
We adapt mixup to knowledge distillation as follows: on each iteration we construct random pairs of inputs $\vec x$, $\vec x'$ from the training set and mix them as $\lambda \cdot \vec x + (1 - \lambda) \cdot \vec x'$,
where the coefficient $\lambda$ is sampled uniformly on $[0, 1]$\footnote{
Note that unlike in the original mixup procedure we are only mixing the inputs and we use the predictions of the teacher on the mixed inputs as the target for the student.}.

\textbf{Synthetic GAN-generated images}\quad We use a Spectral Normalization GAN (SN-GAN) trained on CIFAR-100 \citep{miyato2018spectral} to generate synthetic data for distillation. We used the same pretrained SN-GAN ($\mathrm{FID}=74.2617$, $\mathrm{IS}=6.6023$) for all experiments. 
Our synthetic augmentation procedure was the following: for each minibatch of real training data, we concatenated synthetic images sampled from a pretrained SN-GAN at a ratio of 1 synthetic image to 4 real images.

\textbf{Random noise} \quad 
To observe the effect of unnatural images in the distillation dataset we augment with images sampled pixel-wise from uniform $[0,1]^d$. During distillation each image in a minibatch is randomly resampled with probability $p=0.2$.

\textbf{Out-of-domain data} \quad Finally, we consider using images from the SVHN dataset \citep{netzer2011reading} which is semantically unrelated to the target CIFAR-100 dataset.

We use the \texttt{torchvision.transoforms} package \citep{pytorch} to implement the augmentations from the Baseline Augmentations and Conventional Image Transformations categories:
\begin{itemize}
    \item Horizontal flips: \texttt{torchvision.transforms.RandomHorizontalFlip()}
    \item Random shifts: \texttt{torchvision.transforms.RandomCrop(size=<input\_size>, padding=4)}
    \item Vertical flips: \texttt{torchvision.transforms.RandomVerticalFlip()}
    \item Color jitter: \texttt{torchvision.transforms.ColorJitter(brightness=0.2, contrast=0.2, saturation=0.2, hue=0.2)}
    \item Random rotations: \texttt{torchvision.transforms.RandomRotation(degrees=20)}
\end{itemize}

\subsection{Network architectures}
\label{supp/subsec:architectures}

\paragraph{Image classifiers}
For experiments on CIFAR-100 we used preactivation ResNets with batchnorm, skip connections \citep{he2016deep}, and the standard three-stage macro-structure, varying the number of layers in each stage (i.e. the depth of the network). For all choices of depth we used the same number of filters in each stage (16, 32, and 64, respectively). 
In Section \ref{supp/subsec:distilling_vgg} we use a VGG-16 network without batch-normalization, with implementation directly adapted from \url{https://github.com/pytorch/vision/blob/master/torchvision/models/vgg.py}. 
For experiments on MNIST and EMNIST we used a 5-layer LeNet \citep{lecun1989backpropagation}. 
For ImageNet we used ResNet50 networks for both teacher and students.

\paragraph{Image generators}

\begin{figure*}
\centering
\includegraphics[width=0.24\textwidth]{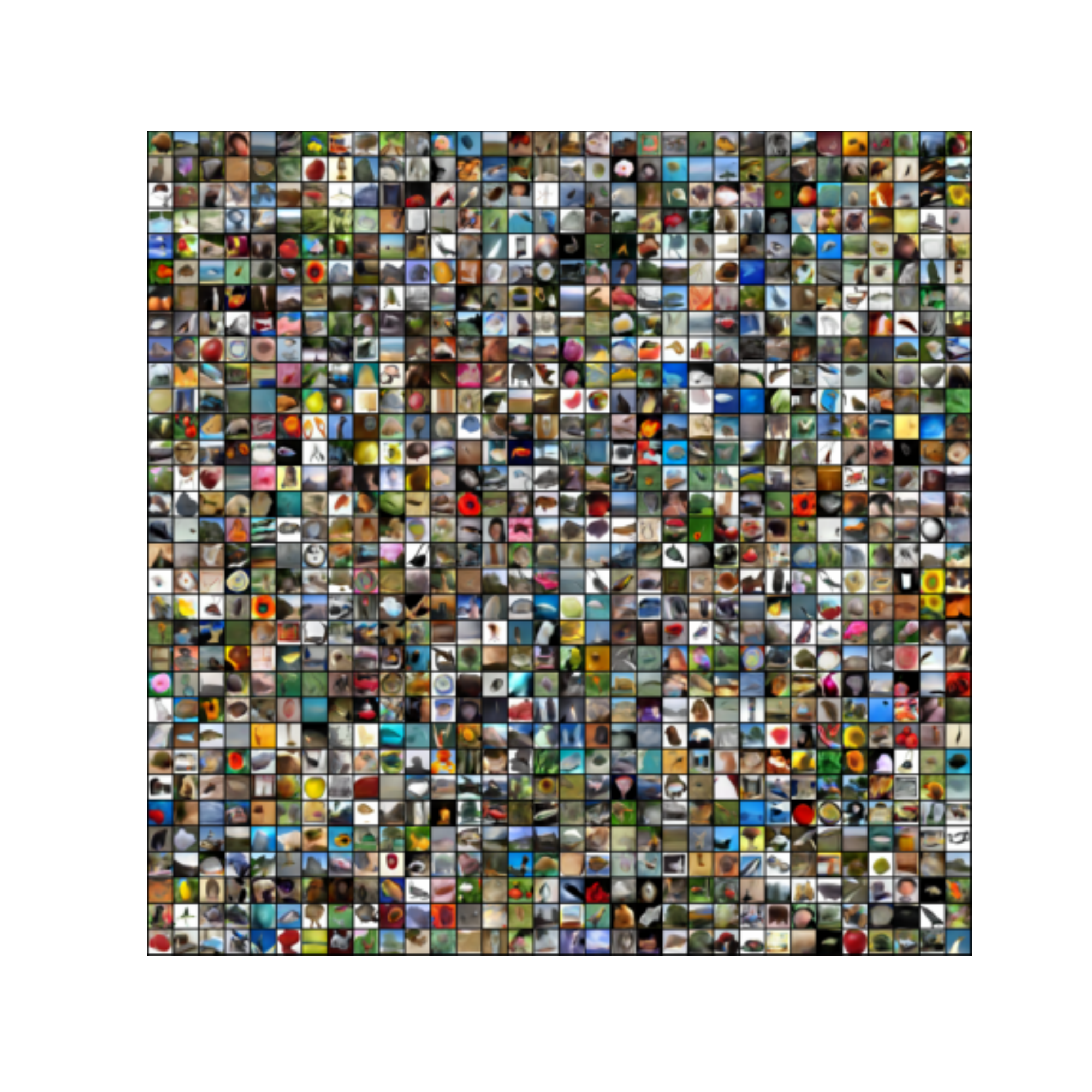}
\hfill
\includegraphics[width=0.24\textwidth]{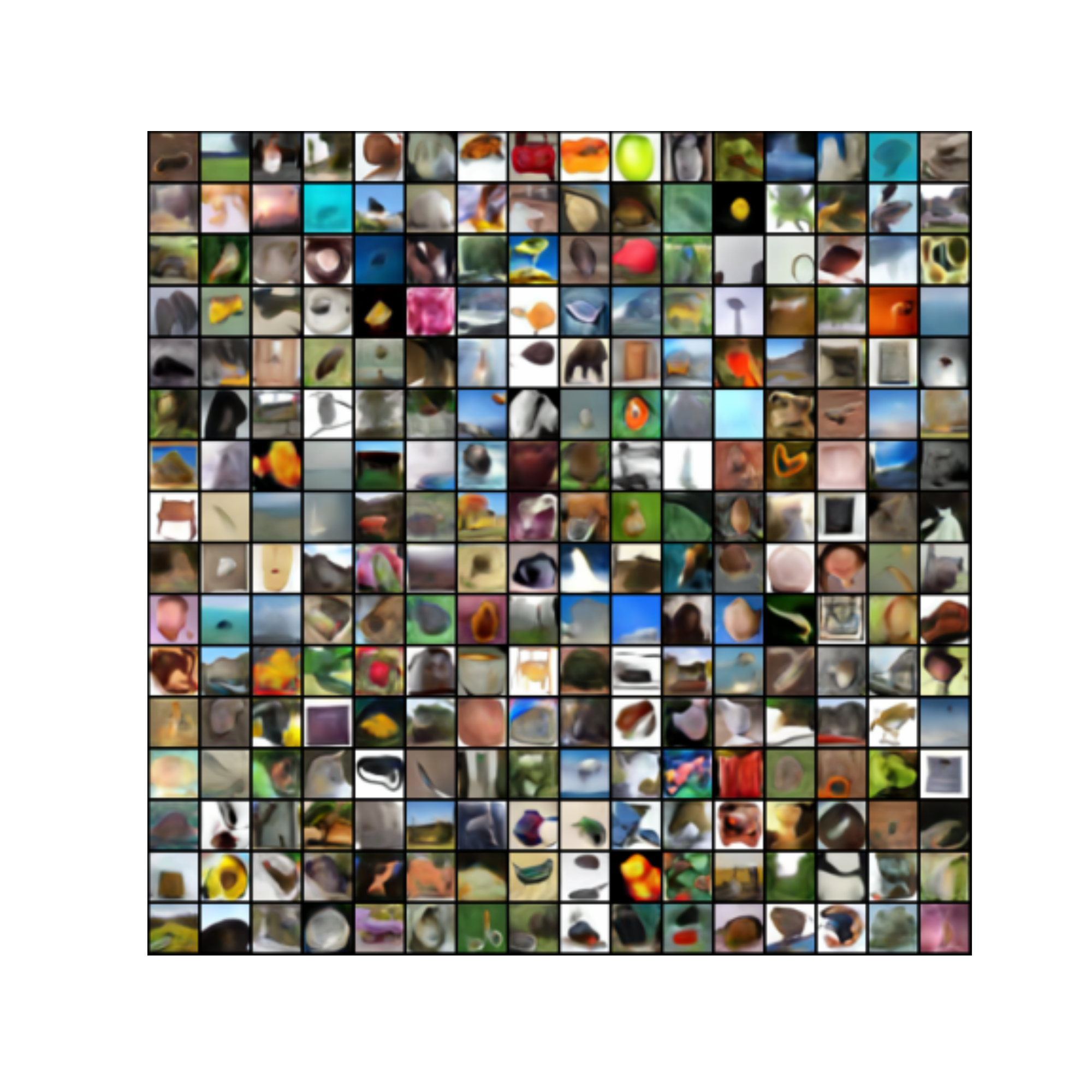}
\includegraphics[width=0.24\textwidth]{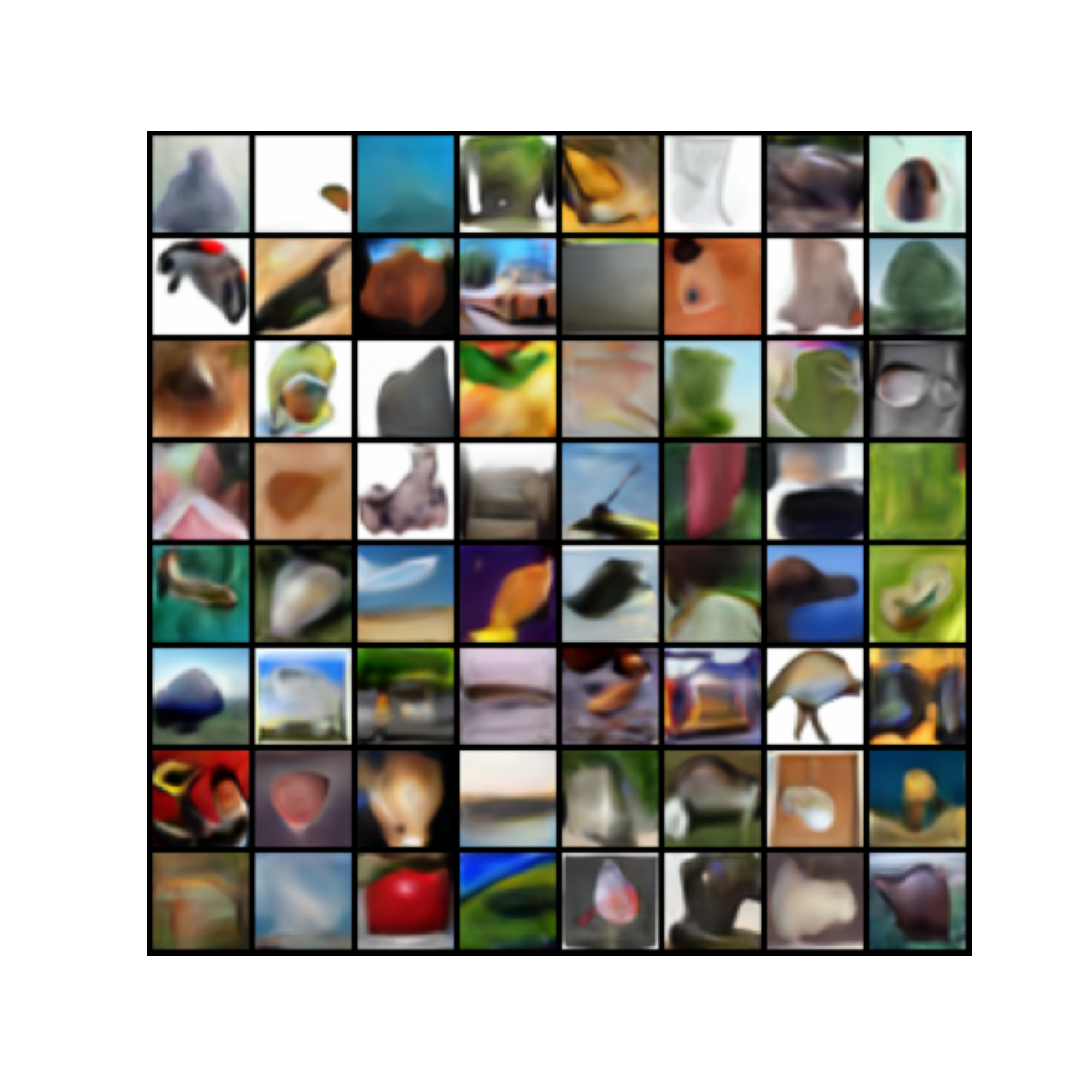}
\hfill
\includegraphics[width=0.24\textwidth]{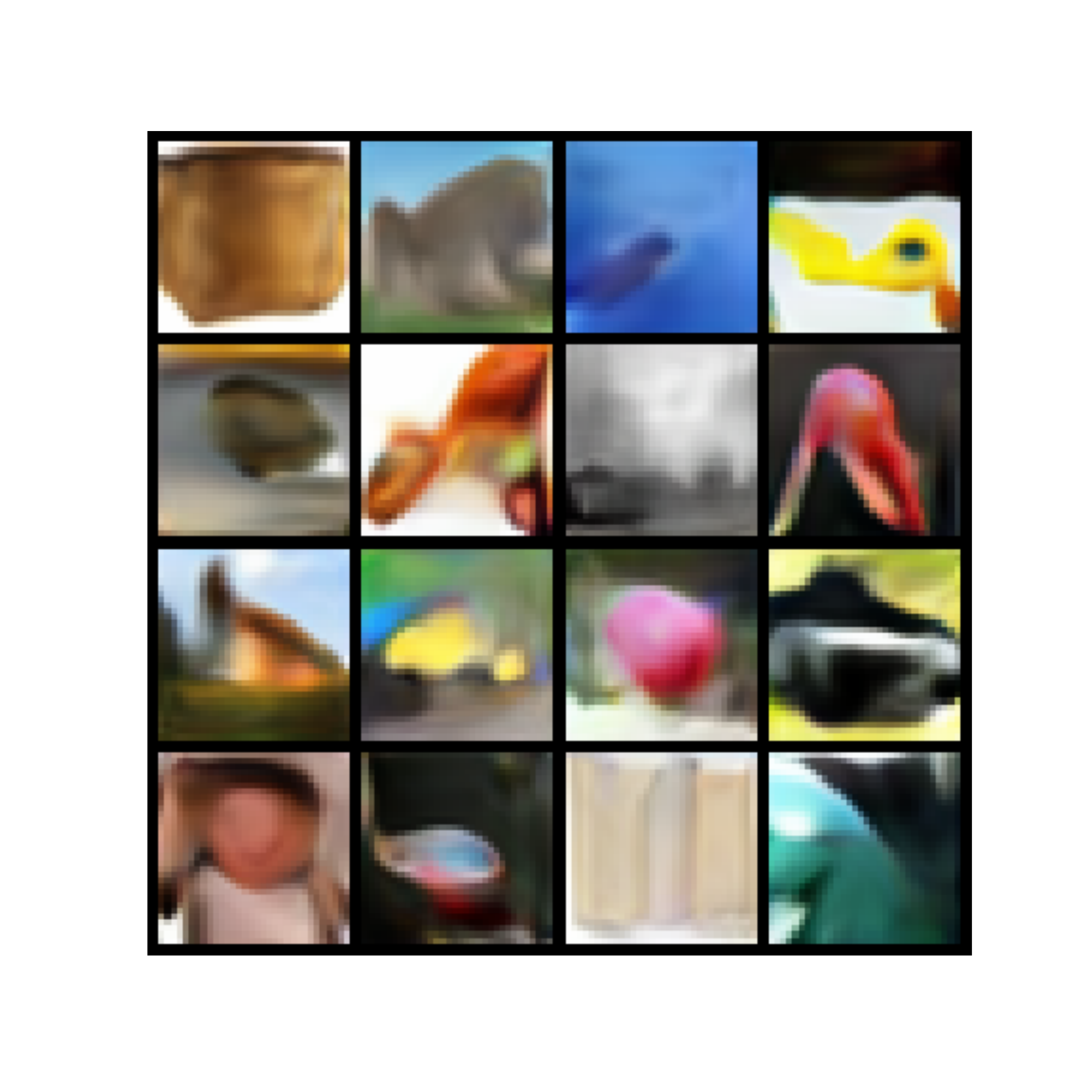}
\caption{Sample images from our SN-GAN ($\mathrm{FID}=74.2617$, $\mathrm{IS}=6.6023$) trained on CIFAR-100.}
\label{supp/fig:gan_samples}
\end{figure*}

We used the standard three-stage ResNet architectures for the SN-GAN generator and discriminator from \citet{miyato2018spectral}. The generator latent dimension was 128, and each generator stage had 256 filters. The discriminator had 128 filters in each stage. Sample images are shown in Figure \ref{supp/fig:gan_samples}.

\paragraph{Text classifier} In section \ref{supp/subsec:imdb} on IMDB sentimental analysis problem, we use a bidirectional LSTM recurrent neural network with $2$ layers, embedding size $128$ and LSTM cell size $70$.

\subsection{Training procedures}
\label{supp/subsec:training}

In our experiments, we consider pre-activation ResNet networks with depths $20$, $56$ and $110$
\citep{he2016deep}. We evaluate on MNIST/EMNIST \citep{cohen2017emnist} and CIFAR-100 \citep{krizhevsky2009learning}, focusing primarily on the latter. We chose CIFAR-100 rather than CIFAR-10 because increasing the problem difficulty increases the gap in performance between a single model and an ensemble, making significant trends more apparent.
We independently train each network in the teacher-ensemble to minimize $\mathcal{L}_{\mathrm{NLL}}$ for 200 epochs, and we distill each student by training it to minimize $\mathcal{L}_{\mathrm{KD}}$ for 300 epochs (i.e. we take $\alpha$ in $\mathcal{L}_s$ to be 0). Note that in the literature one typically sees $\alpha > 0$. We have chosen $\alpha = 0$ so that our objective reflects our aim of producing the highest fidelity student possible. We use an SGD optimizer with initial learning rate $5\times10^{-2}$ and cosine annealing learning rate schedule.
We produce augmented datasets for distillation by sampling images from a set of specified sources, including rotations or color jitter applied to ground truth images, uniform `white noise' images, or synthetic GAN-generated images. Unless specified otherwise, the only augmentations applied when training the teacher were the standard random horizontal flip ($p=0.5$) and padded random crop (4 pixel pad width), regardless of the choice of distillation dataset. 

\paragraph{Teacher image classifiers:}
The teacher models were trained through the standard empirical cross-entropy loss for 200 epochs with a batch size of 256 using SGD with momentum ($0.9$ momentum weight) and weight decay of $1.0 \times 10^{-4}$. We used a cosine annealing learning rate schedule with $\eta_{\mathrm{max}} = 0.1$, $\eta_{\mathrm{min}} = 0$. For data augmentation we used random horizontal flips ($p=0.5$) and random crops (padding width 4).

\paragraph{Student image classifiers:}
Our student models were distilled through the temperature-scaled teacher-student cross-entropy with varying temperatures $\tau$ for 300 epochs, with a batch size of 128 using SGD with Nesterov momentum ($0.9$ momentum weight) and weight decay of $1.0 \times 10^{-4}$. We used a cosine annealing learning rate schedule with $\eta_{\mathrm{max}} = 5.0 \times 10^{-2}$, $\eta_{\mathrm{min}} =$ $1.0 \times 10^{-6}$. For details on the data augmentation procudures we considered, the reader is directed to Appendix \ref{supp/subsec:aug_procedures}.

\paragraph{Image generators:}
For synthetic image generation we trained SN-GAN models with the hinge discriminator loss from \citet{miyato2018spectral}. We trained the generator for 100K gradient steps with a batch size of 128. For each generator step, we took 5 discriminator steps. We used Adam ($\beta_1=0, \beta_2=0.9$) and a linearly decayed learning rate
$\eta_{\mathrm{max}} =$ $2.0 \times 10^{-4}$, $\eta_{\mathrm{min}} =$ $1.0 \times 10^{-6}$. We used random horizontal flips ($p=0.5$) as data augmentation for the discriminator. To evaluate FID and IS scores, we used 5K samples from the generator and the pretrained PyTorch Inception-v3 networks\footnote{\url{https://pytorch.org/docs/stable/torchvision/models.html?highlight=inception\#torchvision.models.inception_v3}}. For the discriminator and generator architectures the reader is referred to Appendix \ref{supp/subsec:architectures}.

\paragraph{Text classifiers:}
For text classification on IMDB dataset, we train LSTM networks for $100$ epochs with learning rate $10^{-2}$, weight decay $10^{-3}$ and batch size $100$ sequences. For the data loader, we use $100$ as maximum sequence length and filter out tokens in the vocabulary that are present less than $10$ times. 

\paragraph{ImageNet experiments:} we trained the teachers with weight decay $10^{-4}$ and did not use weight decay for training the students.
We trained both the teachers and the students for 90 epochs using the SGD optimizer with momentum 0.9 and a cosine decay learning rate schedule with a linear learning rate ramp-up for 5 epochs to the initial value of $0.1$.
We used a batch size of 1024.

%\clearpage

\section{Additional understanding experiments}

In this section we include additional experimental results that were not included in the main text in the interest of clarity, but are still noteworthy to those seeking a deep understanding of the behavior of knowledge distillation.

\subsection{Understanding the effect of teacher capacity on the distillation labels}
\label{supp/subsec:understanding_ensemble_variation}

In this subsection, we explore the qualitative effect of teacher ensemble size, network depth, and distillation temperature on the predictive distributions on train and test, to get a better understanding of what the students are being asked to emulate.

\begin{figure}[h]
\centering
\begin{tabular}{cccccc}
\hspace{-0.35cm}\includegraphics[width=0.16\textwidth]{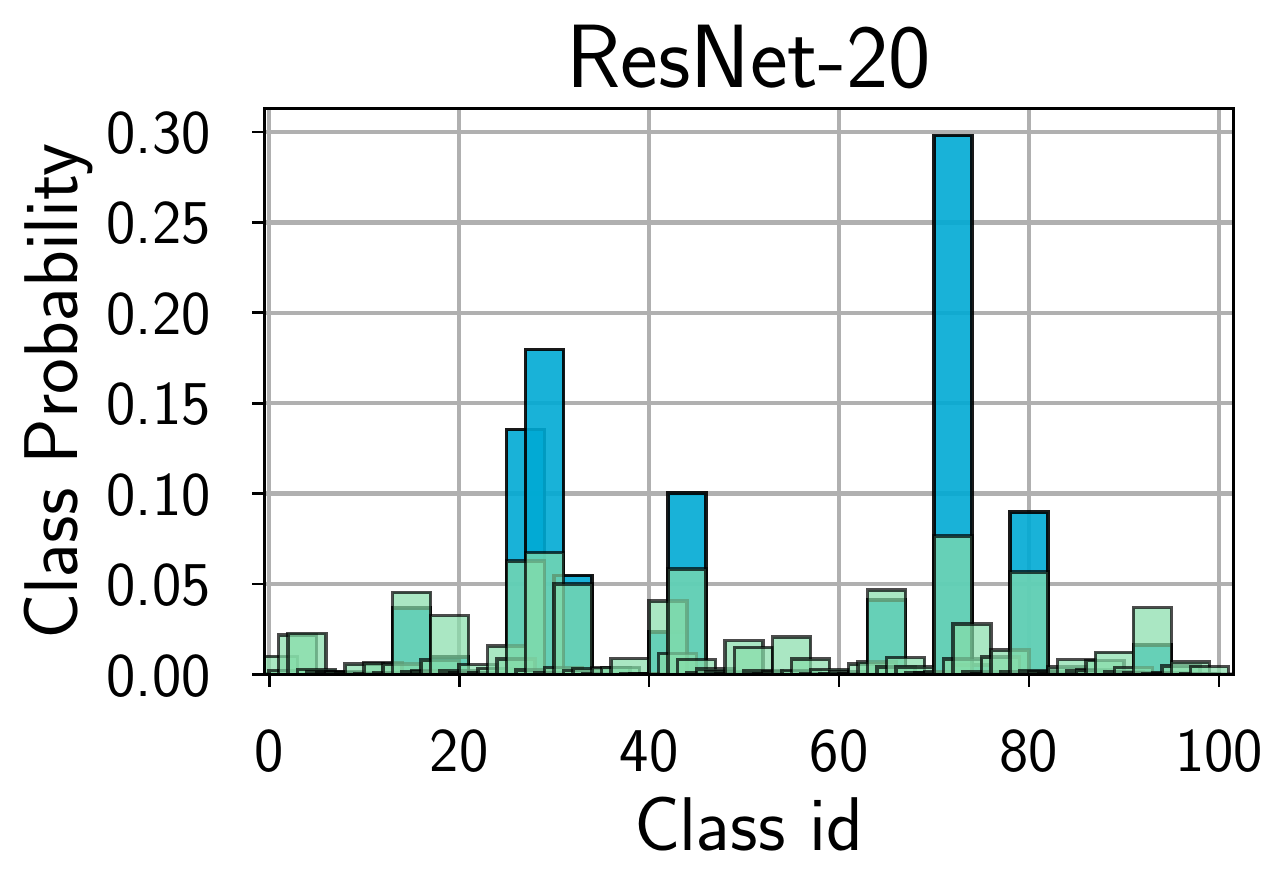} &
\hspace{-0.35cm}\includegraphics[width=0.16\textwidth]{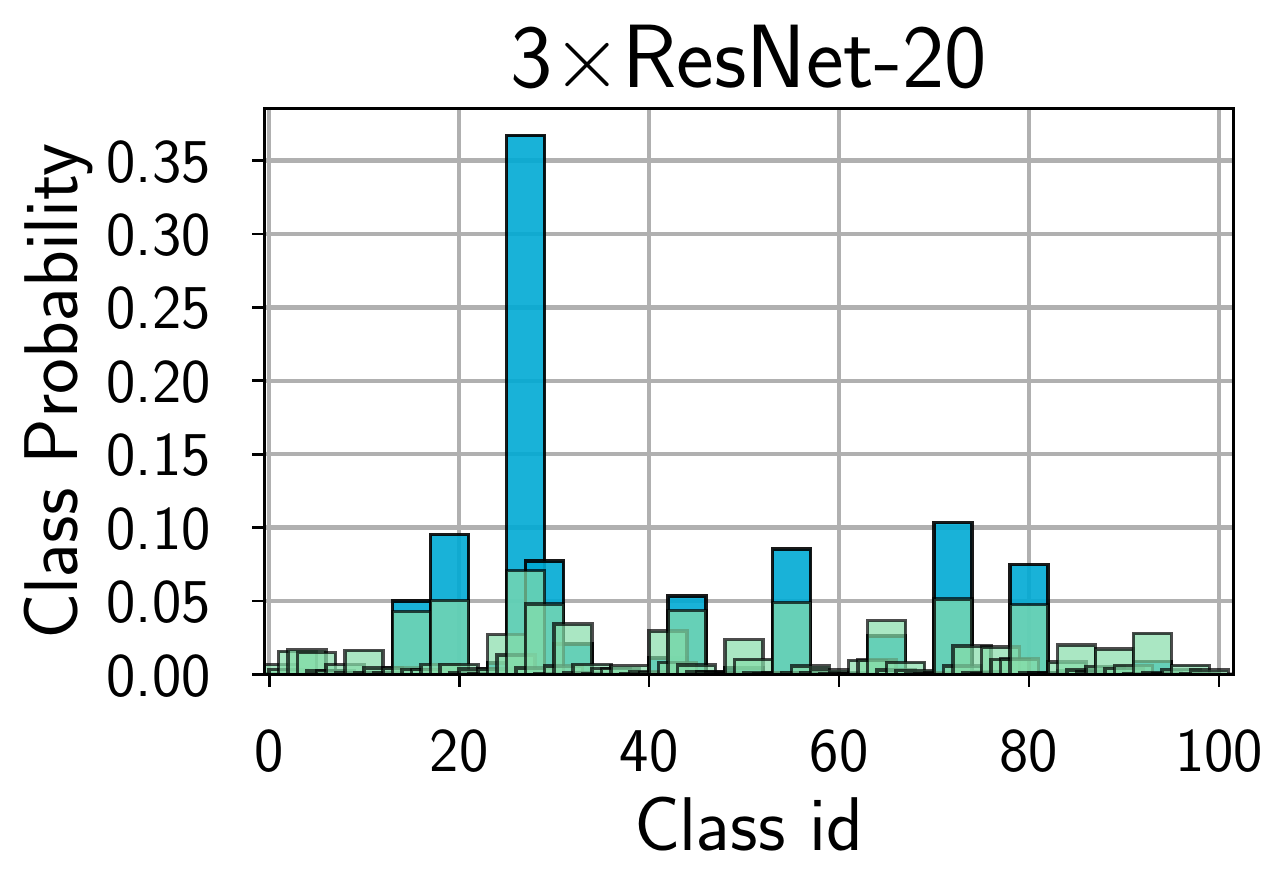} &
\hspace{-0.35cm}\includegraphics[width=0.16\textwidth]{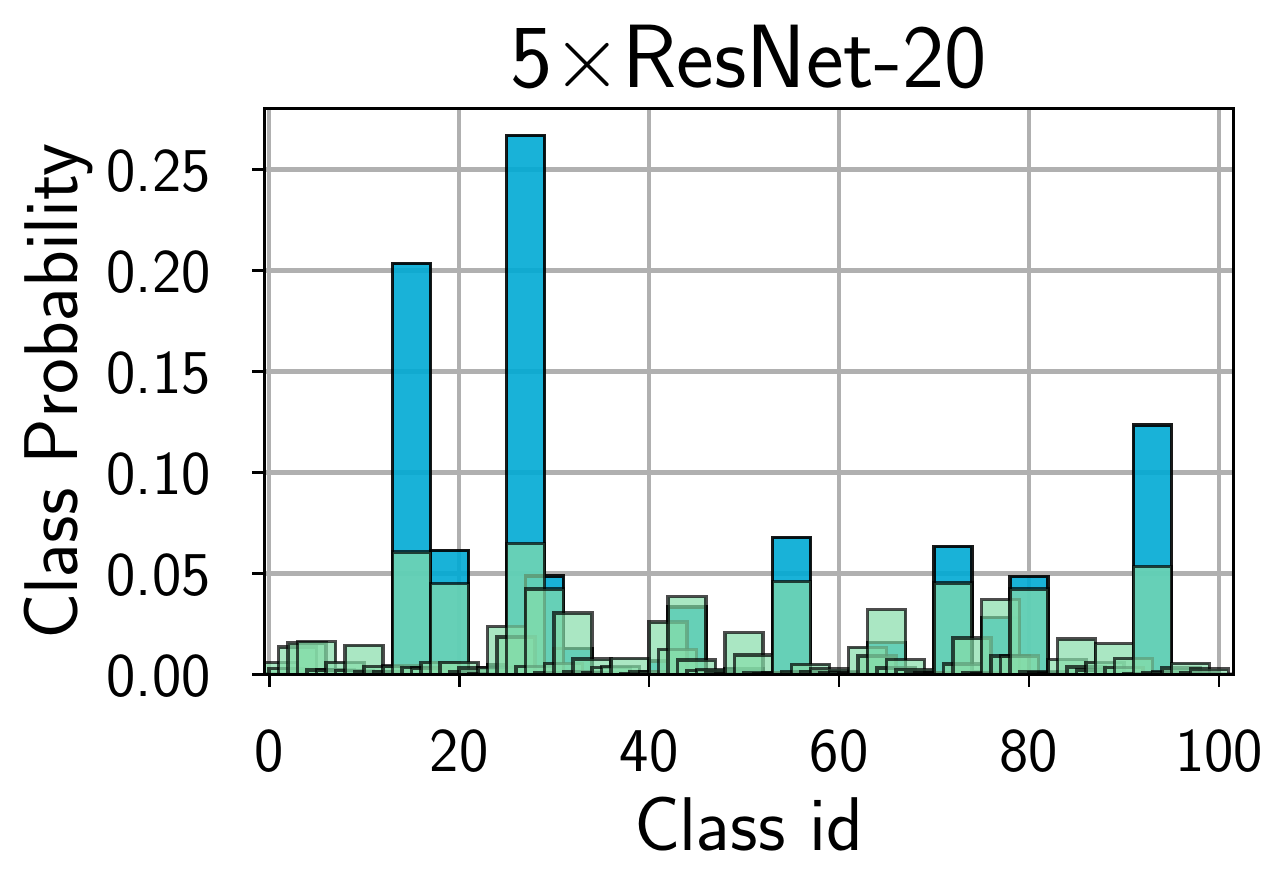} &
\quad
\hspace{-0.35cm}\includegraphics[width=0.16\textwidth]{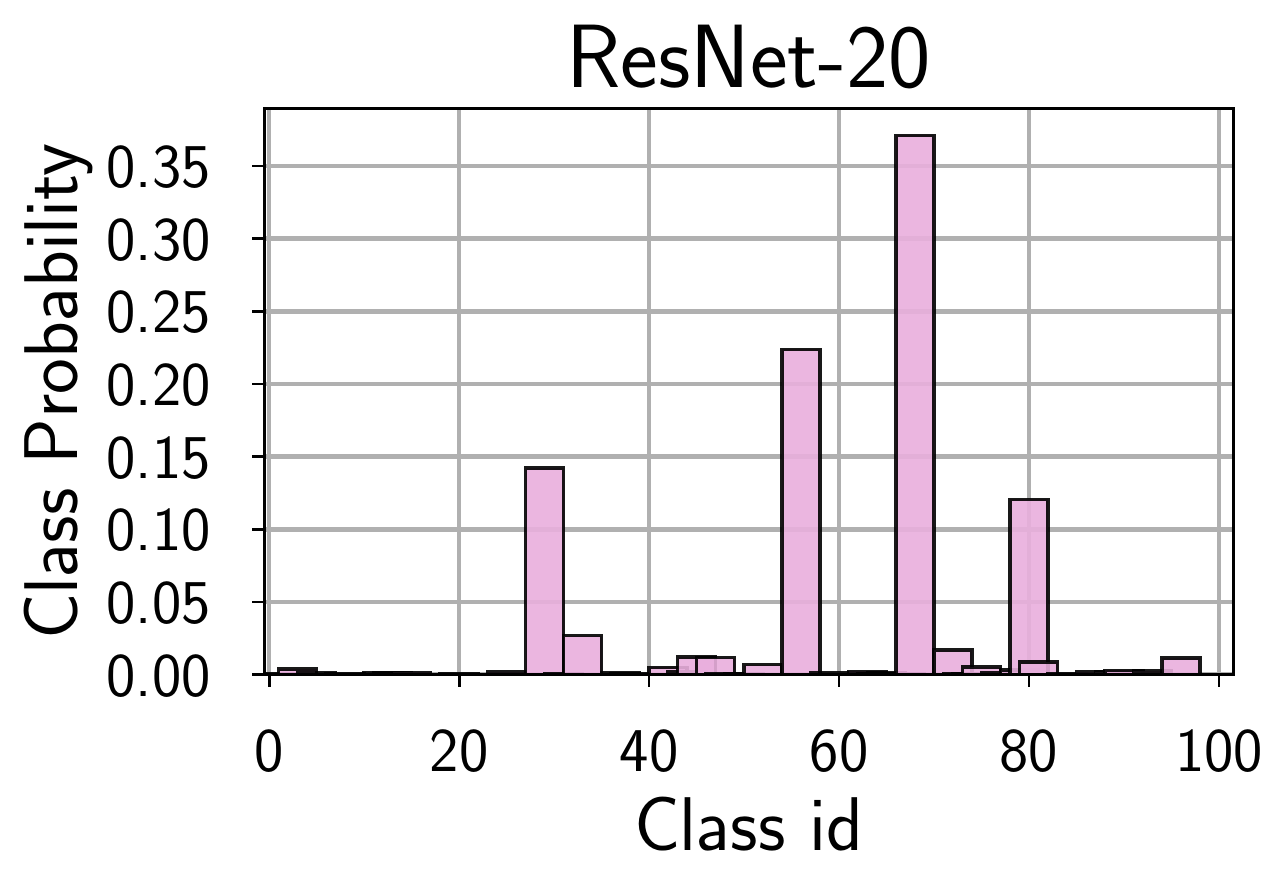} &
\hspace{-0.35cm}\includegraphics[width=0.16\textwidth]{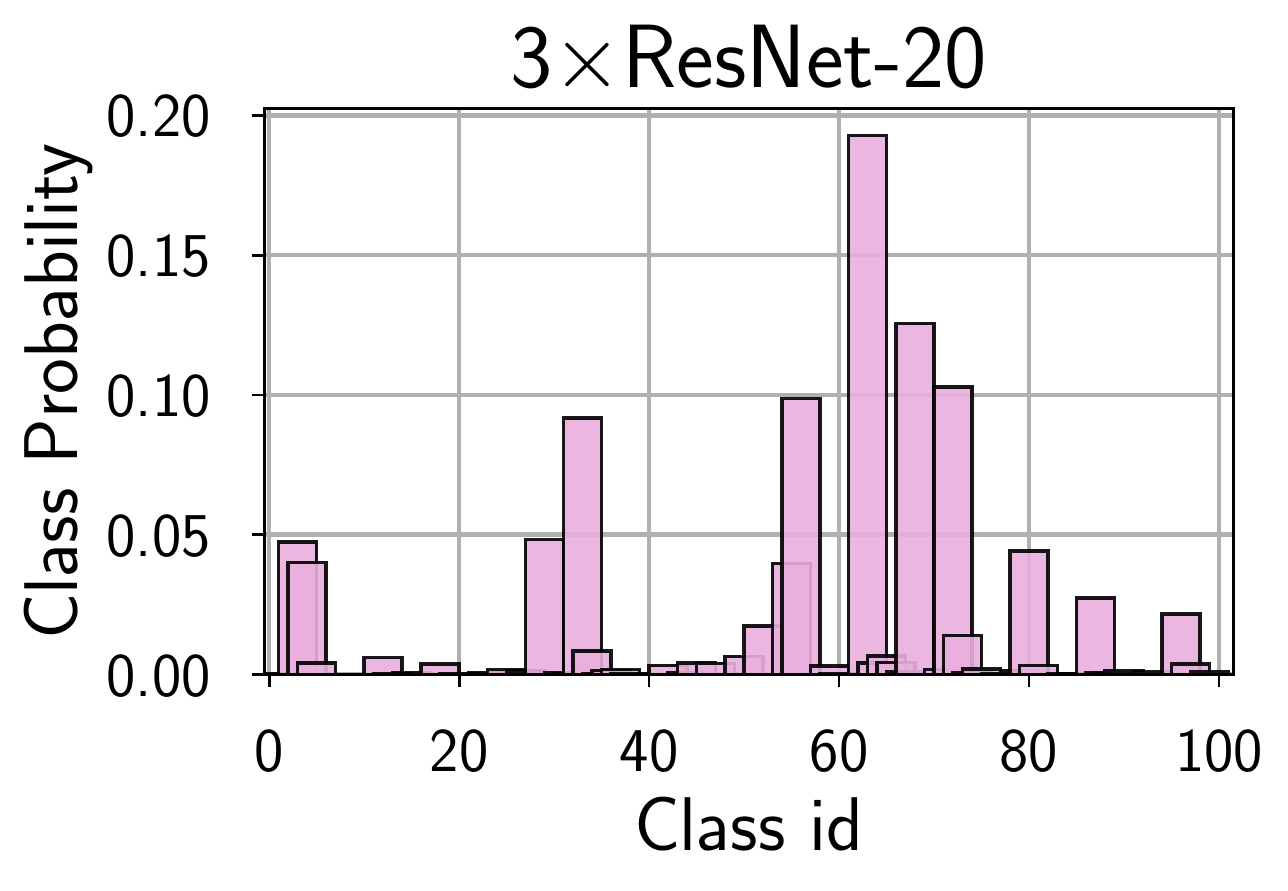} &
\hspace{-0.35cm}\includegraphics[width=0.16\textwidth]{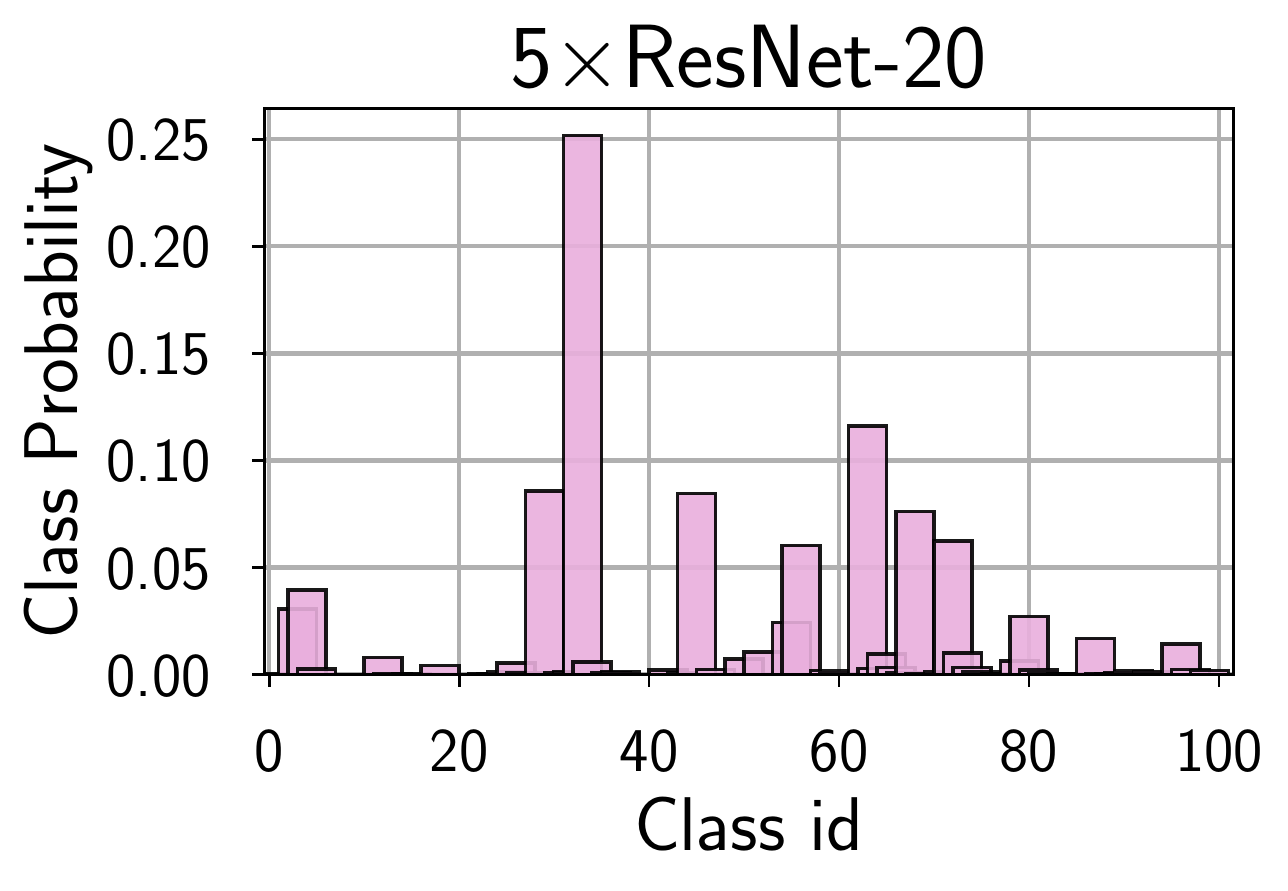}
\\
\hspace{-0.35cm}\includegraphics[width=0.16\textwidth]{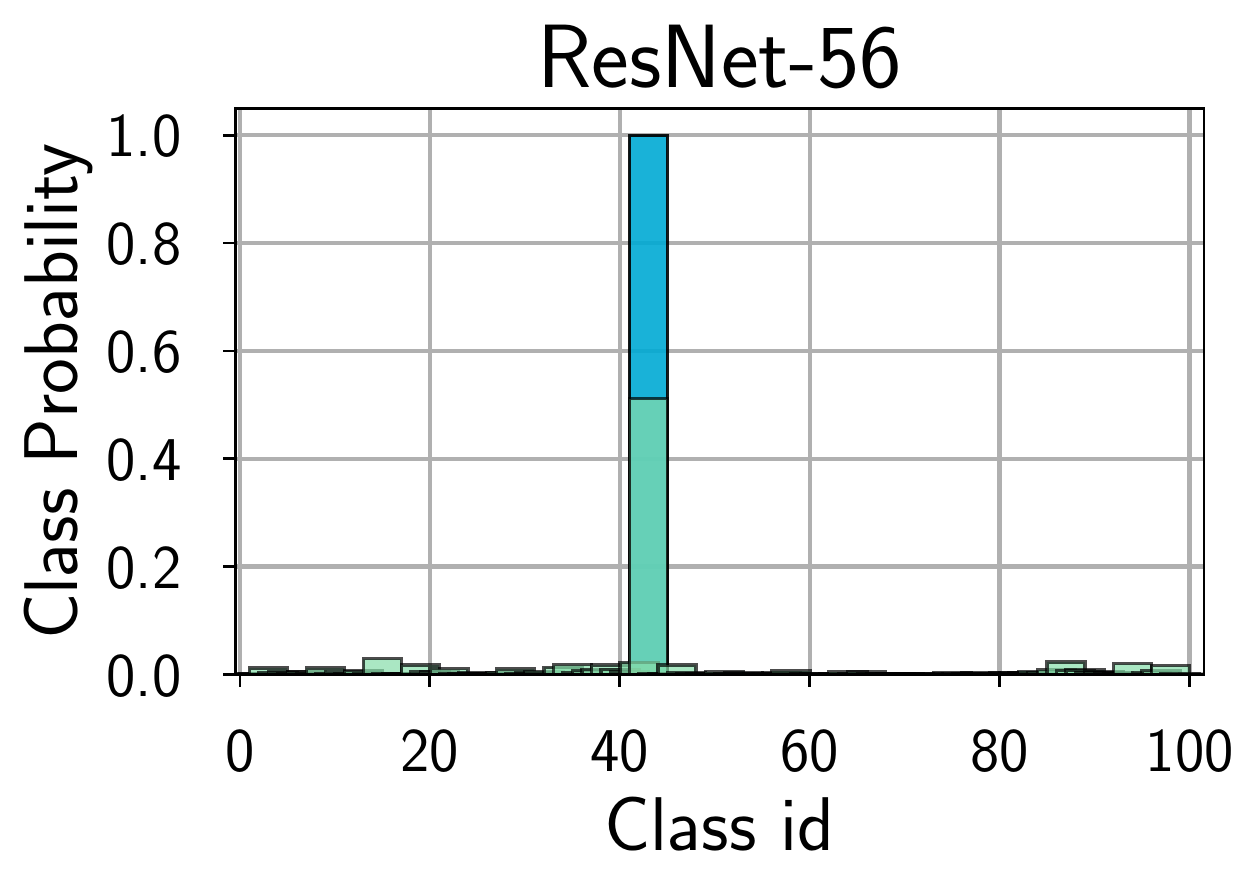} &
\hspace{-0.35cm}\includegraphics[width=0.16\textwidth]{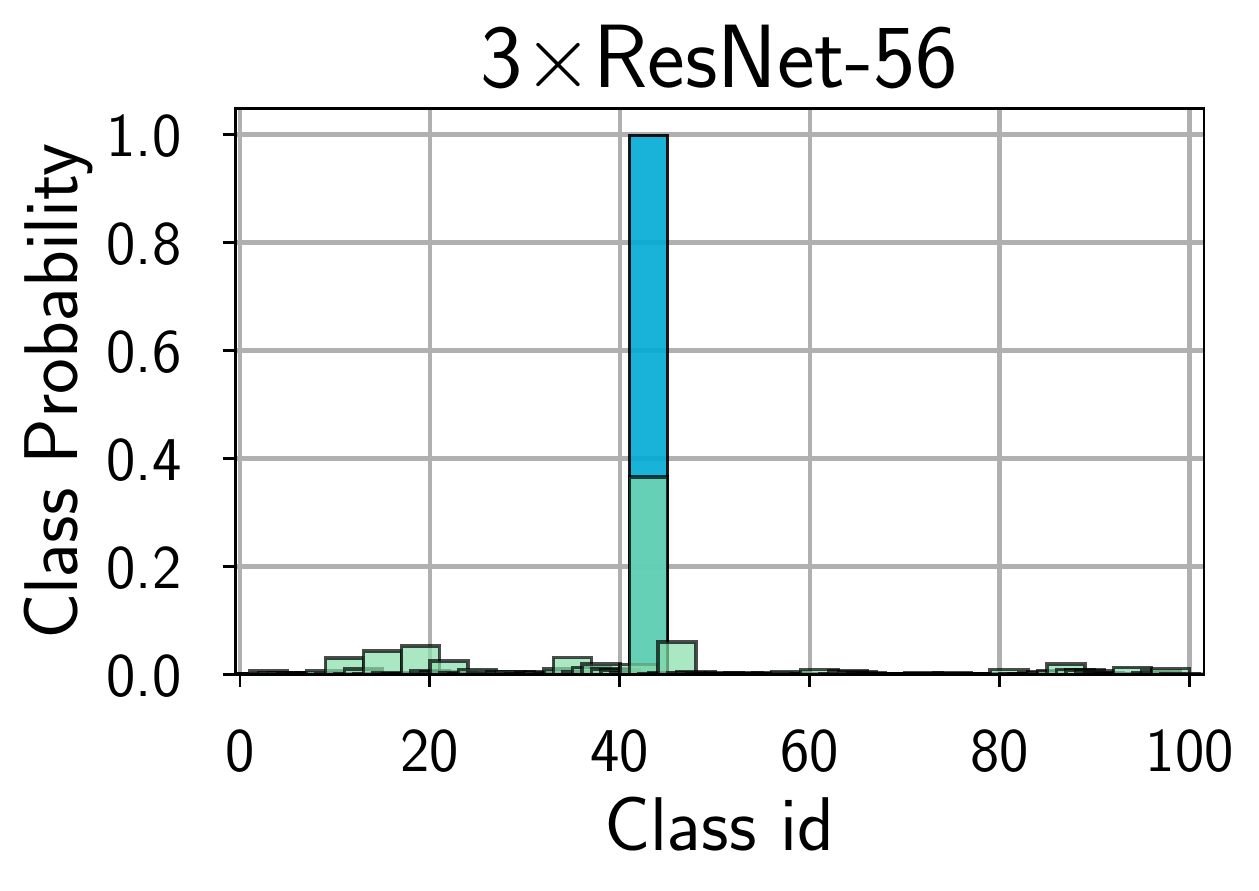} &
\hspace{-0.35cm}\includegraphics[width=0.16\textwidth]{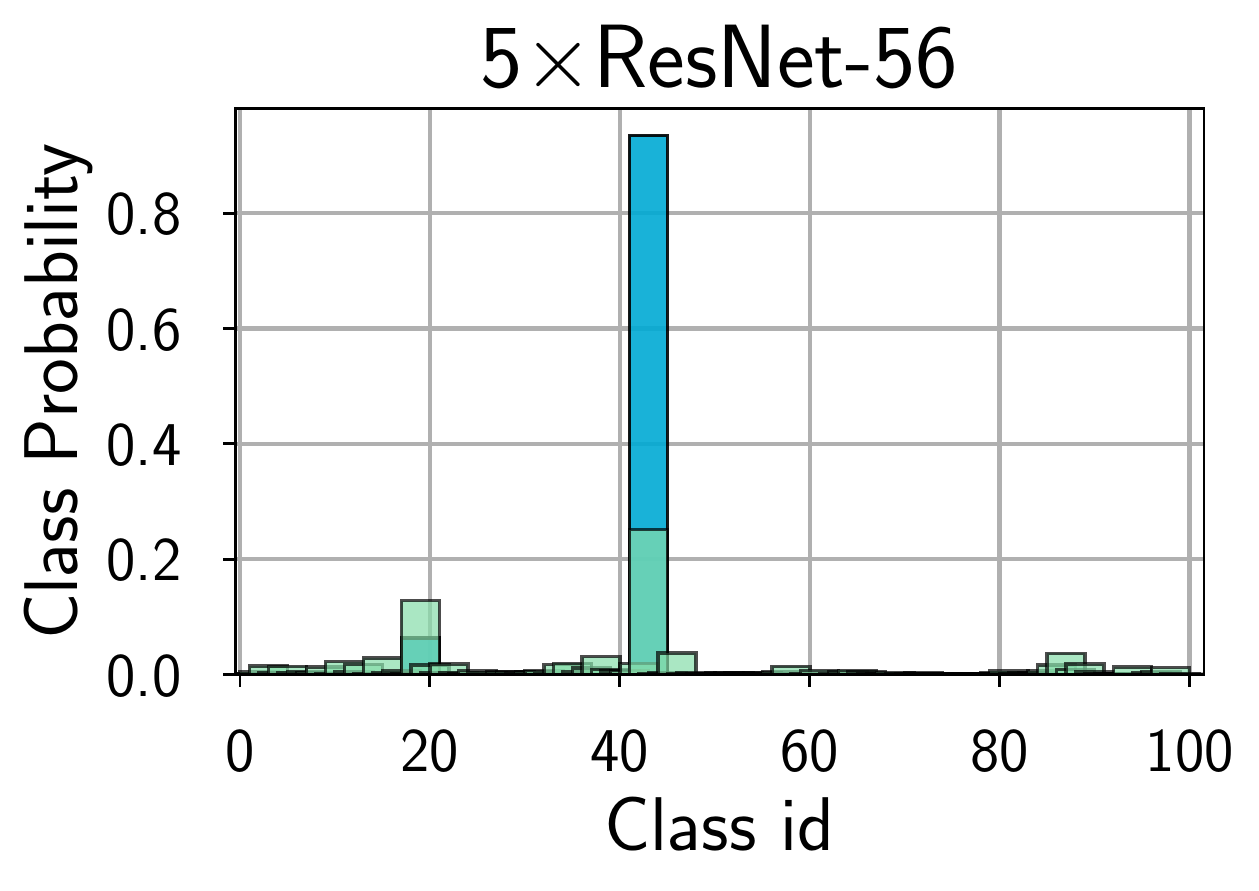} &
\quad
\hspace{-0.35cm}\includegraphics[width=0.16\textwidth]{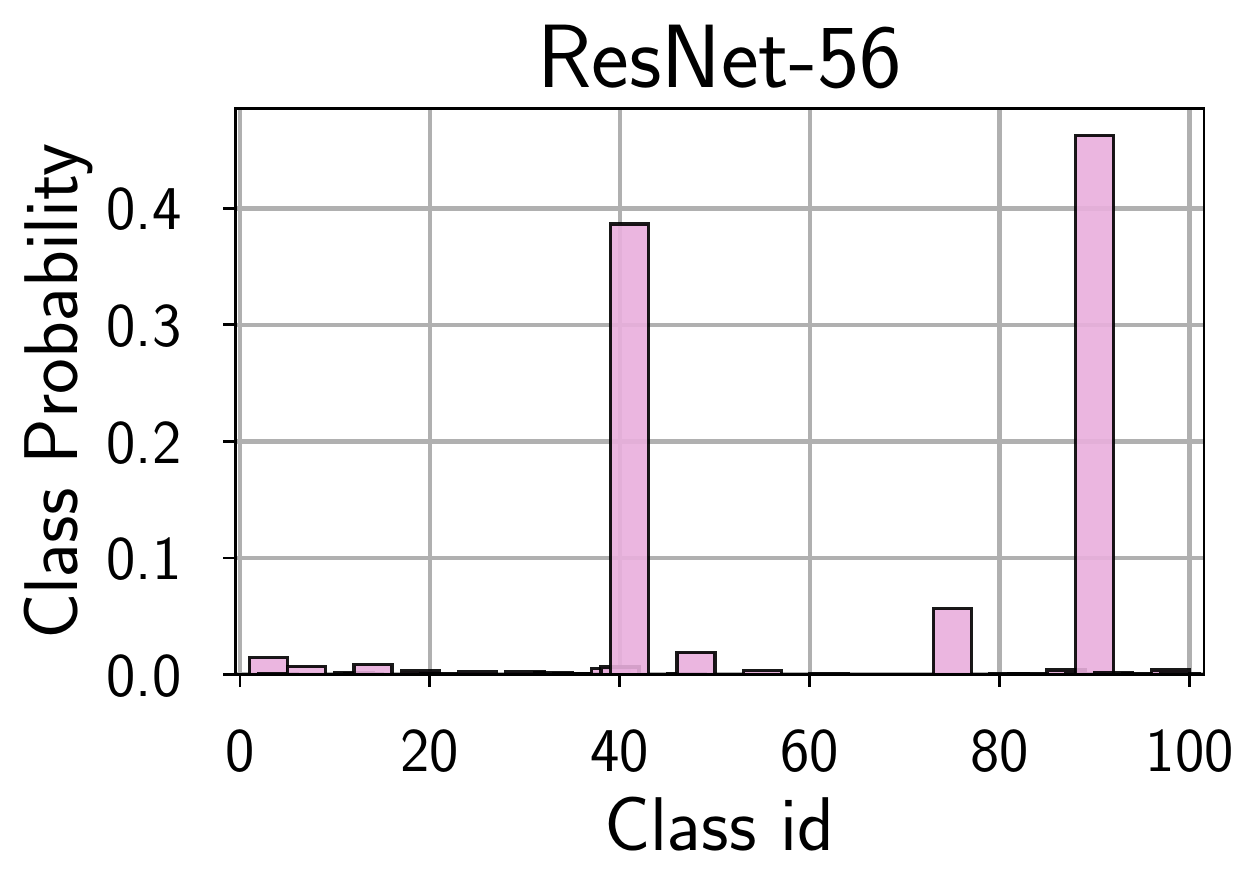} &
\hspace{-0.35cm}\includegraphics[width=0.16\textwidth]{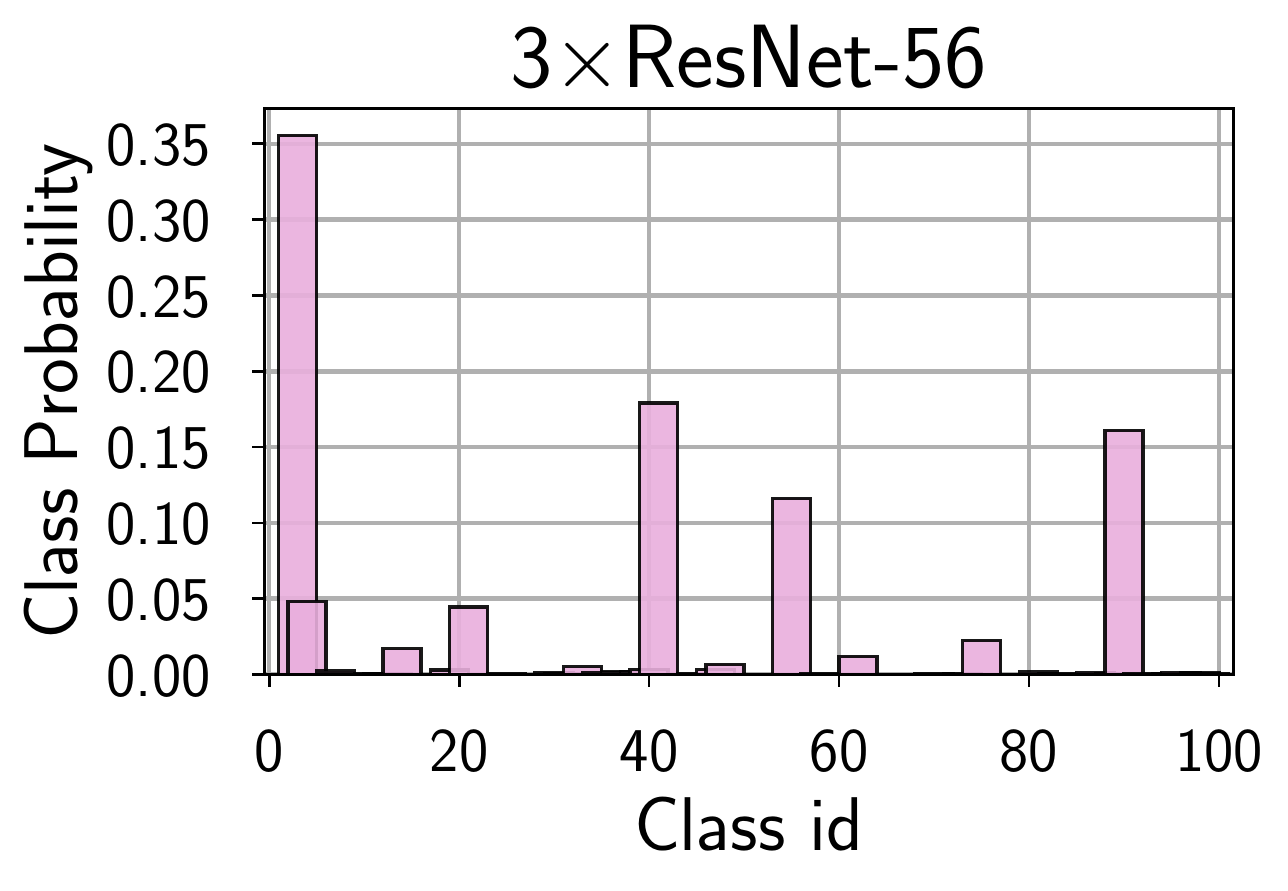} &
\hspace{-0.35cm}\includegraphics[width=0.16\textwidth]{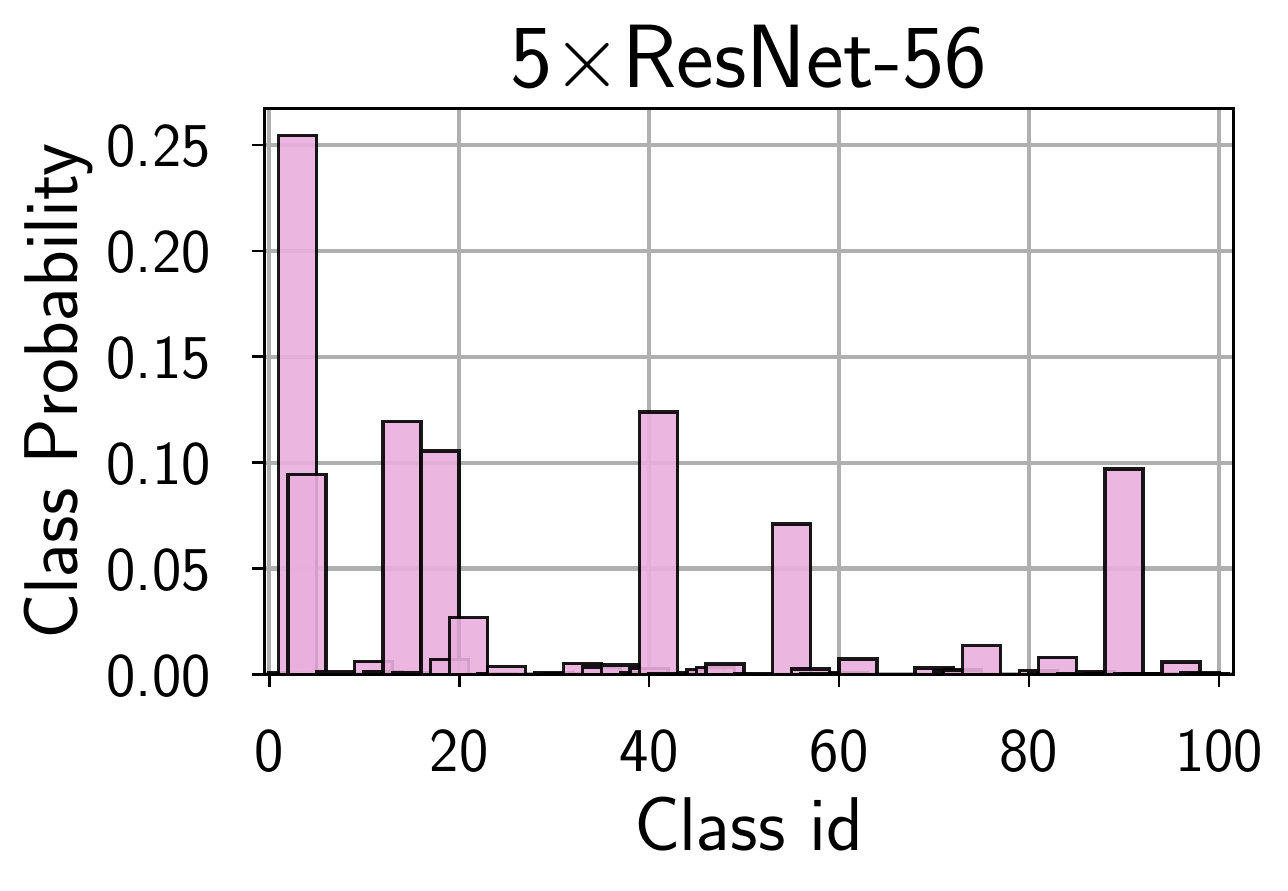}
\\
\hspace{-0.35cm}\includegraphics[width=0.16\textwidth]{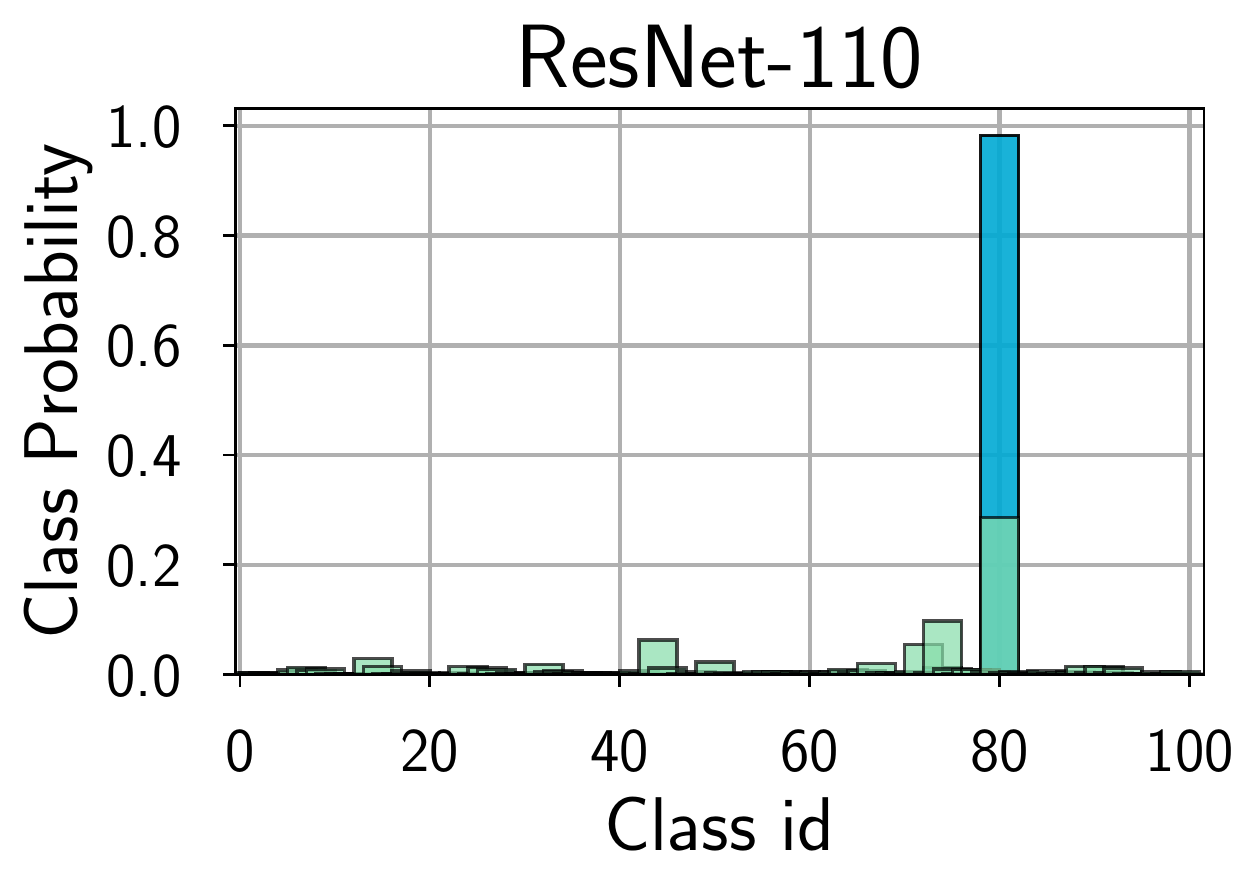} &
\hspace{-0.35cm}\includegraphics[width=0.16\textwidth]{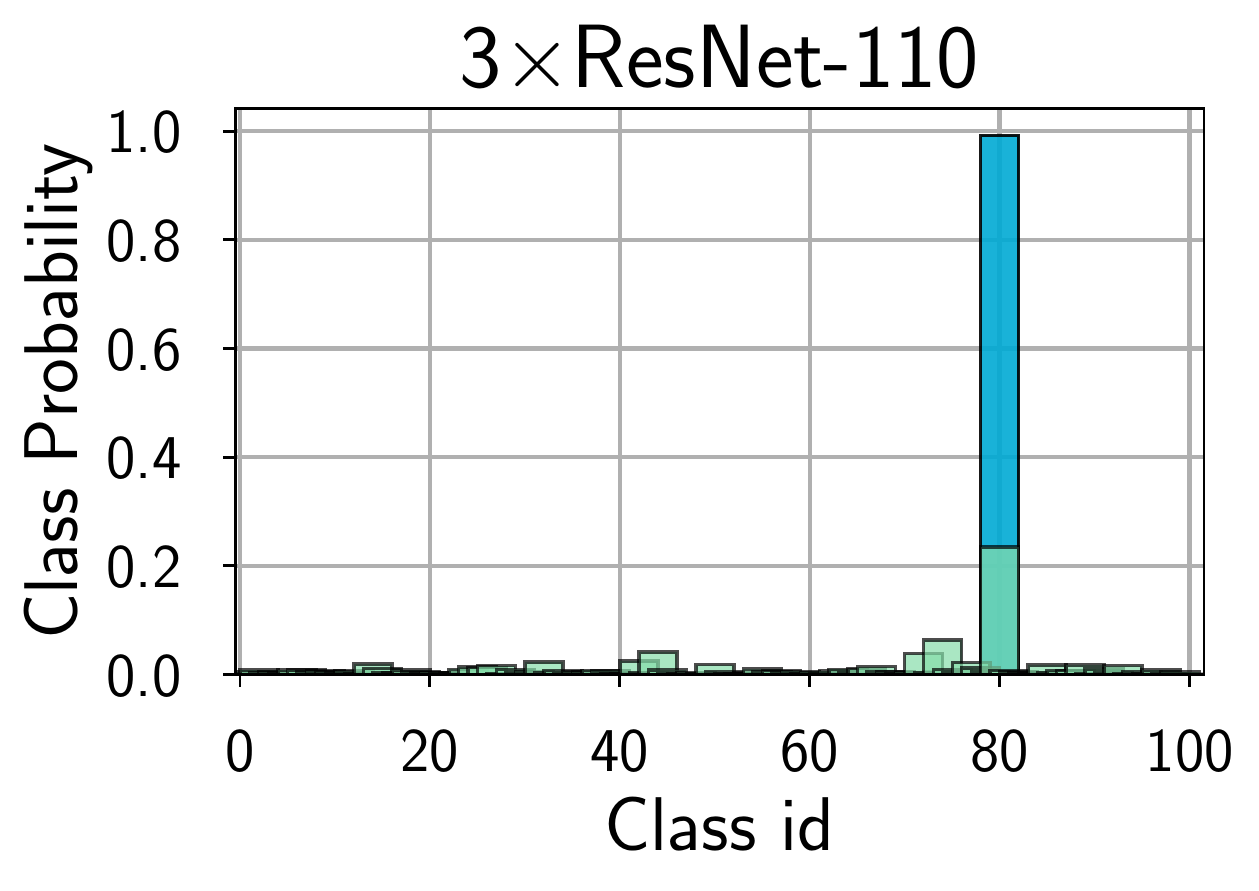} &
\hspace{-0.35cm}\includegraphics[width=0.16\textwidth]{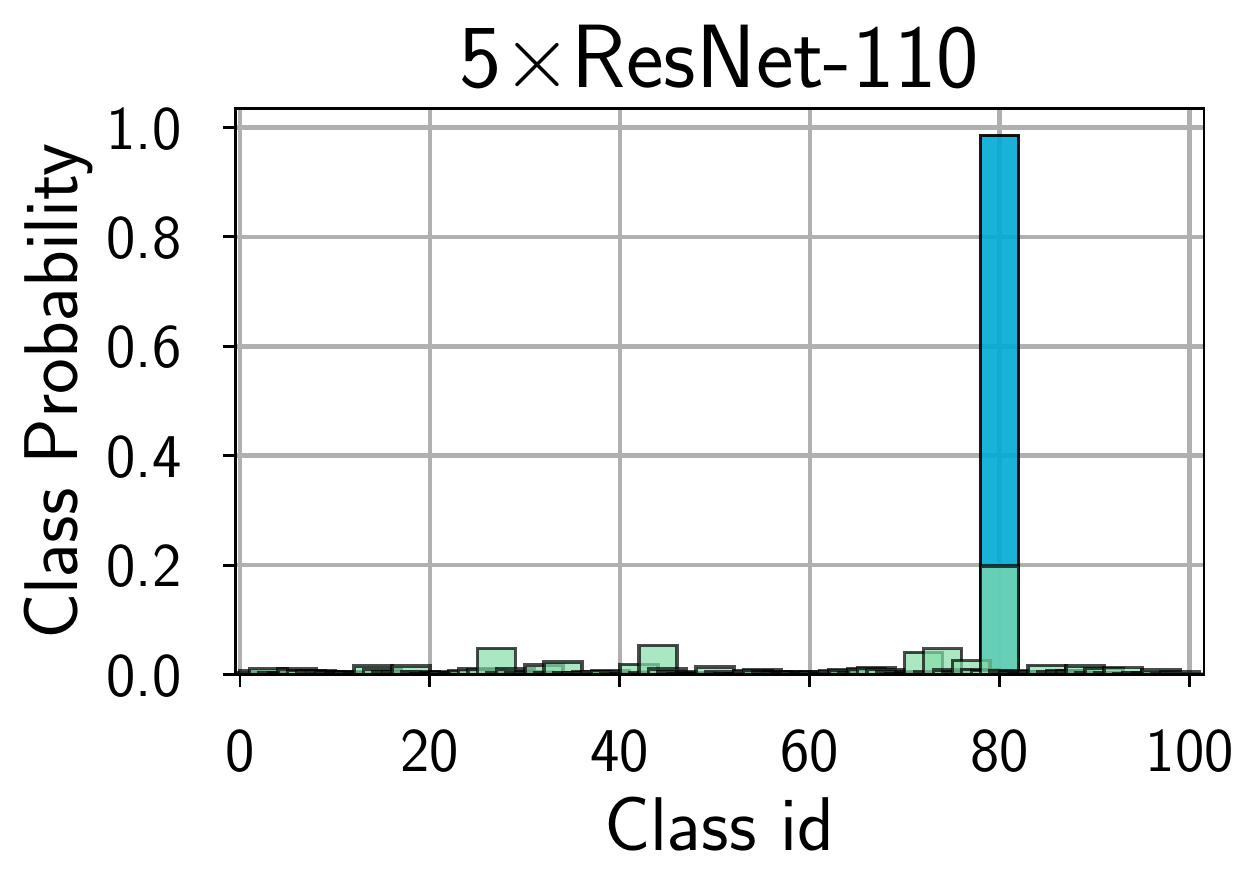} &
\quad
\hspace{-0.35cm}\includegraphics[width=0.16\textwidth]{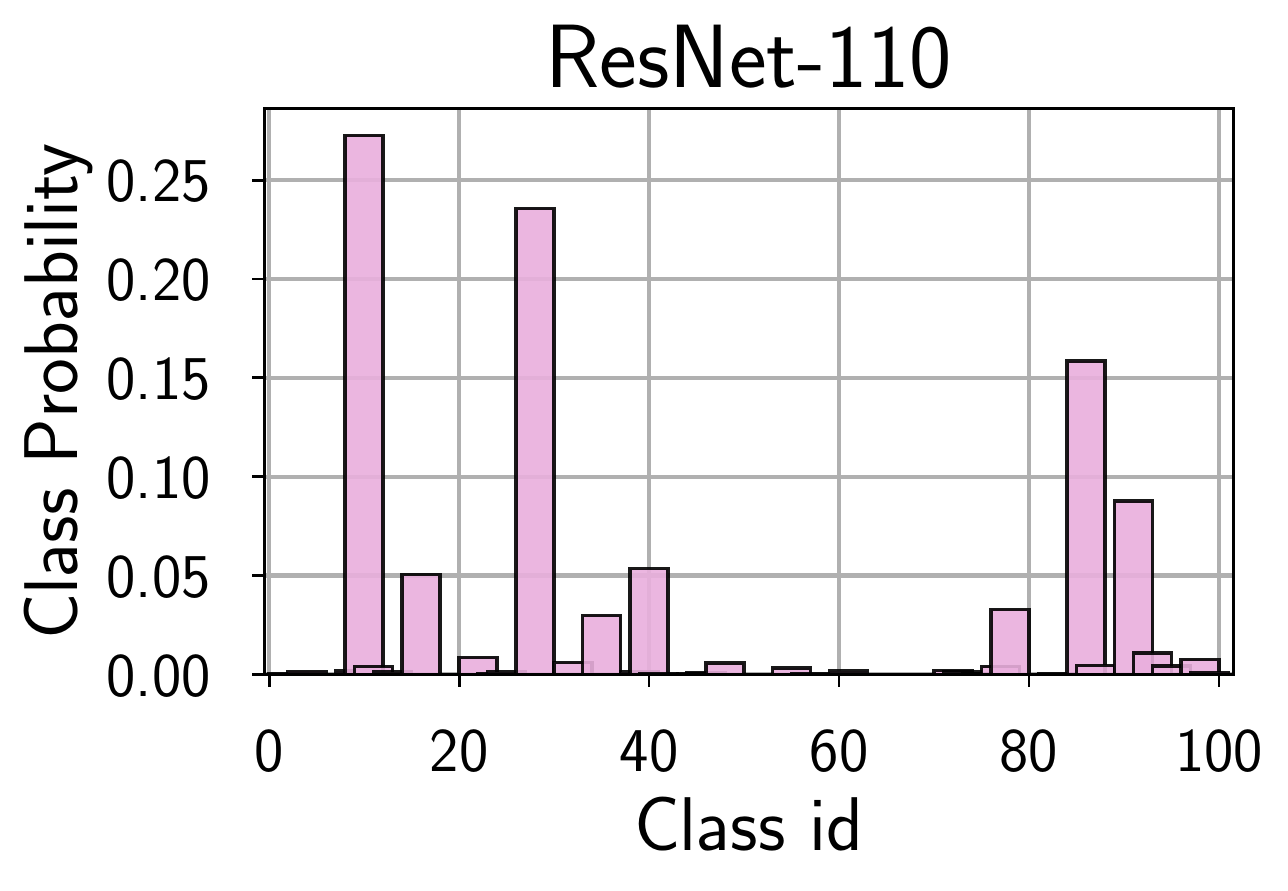} &
\hspace{-0.35cm}\includegraphics[width=0.16\textwidth]{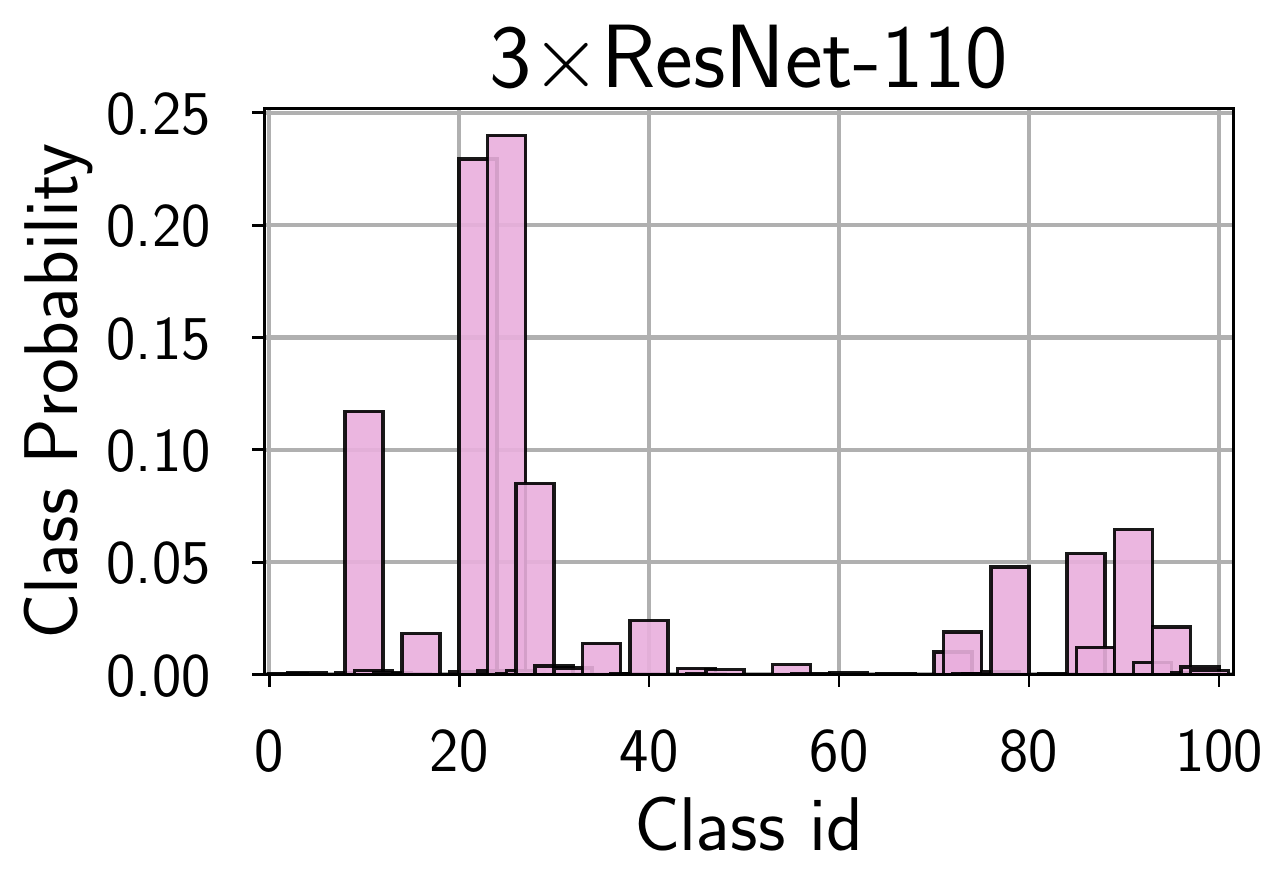} &
\hspace{-0.35cm}\includegraphics[width=0.16\textwidth]{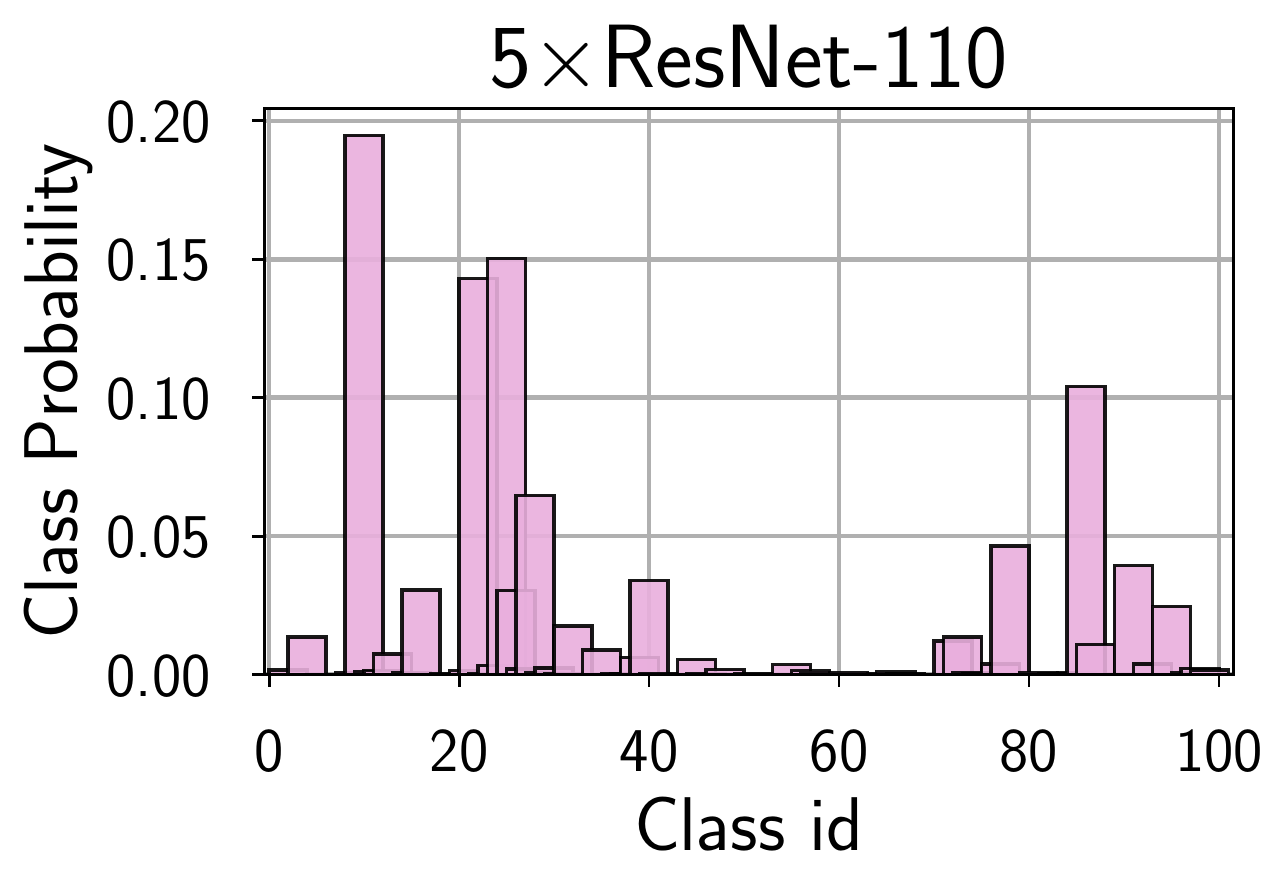}
\\
\multicolumn{3}{c}{\small (a) Train Data} &
\multicolumn{3}{c}{\small (b) Test Data}
\end{tabular}
\caption{Teacher predictive distributions for example images from CIFAR-100 train \textbf{(left)} and test \textbf{(right)}. For train examples we show the distribution when $\tau=1$ in blue and the tempered distribution when $\tau=4$ in green. For test examples we only show $\tau=1$. Each row corresponds to a different teacher depth, and the column corresponds to the number of ensemble components.}
\label{supp/fig:teacher_predictions}
\end{figure}

Since deep ensemble components are typically large networks that achieve almost 100\% accuracy on train with very high confidence, it is tempting to assume that each ensemble component conveys effectively the same information when used for distillation. 
One consequence of that assumption would be that adding ensemble components would produce little or no improvement in the student if the distillation was performed on train. 
In fact, we find that although the component networks are indeed very confident on the train, there is sufficient variation in their predictive distributions for the student to benefit significantly.
In Figure \ref{supp/fig:teacher_predictions} we provide examples of teacher logits, varying the ensemble size, network depth, and temperature, comparing examples from CIFAR-100 train and test.

As discussed in Section \ref{main/subsec:kd_background} in order to benefit from soft teacher labels, one must choose $\tau$ large enough that the student directs some capacity towards mimicking the smaller teacher logits (Figure \ref{main/fig:subsample_split}, bottom). If $\tau$ is chosen too small (e.g. $\tau = 1$), then the student distilled from a 3-component ResNet56 ensemble is no better than a student distilled from a single network. The improvements in student performance and fidelity taper off fairly quickly as ensemble components are added.

\begin{figure}[h]
    \centering
    \includegraphics[width=0.25\textwidth]{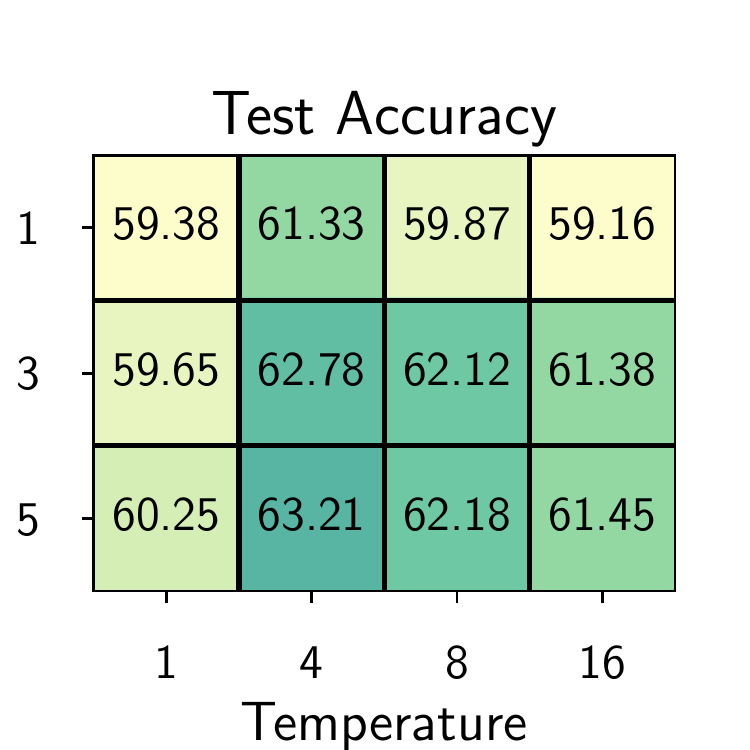}
    %\hfill
    \quad
    \includegraphics[width=0.25\textwidth]{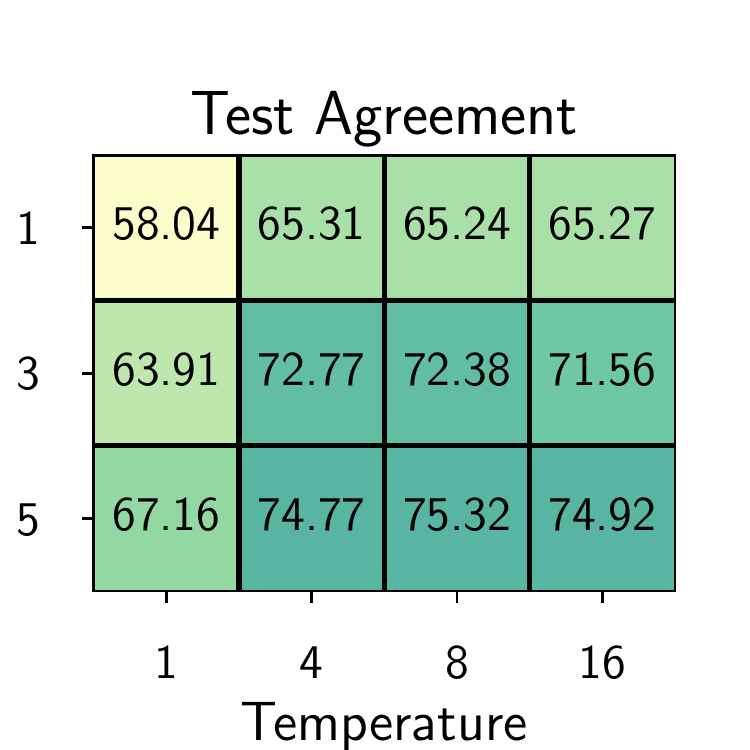}
    \caption{Subsampled CIFAR-100 experiment performed with ResNet20 networks. ResNet20 networks are much less confident on train than ResNet56 networks. As a result increasing the ensemble size will improve the student even with a small temperature setting $\tau=1$.}
    \label{main/fig:resnet_20_a_b}
\end{figure}

The correct choice of $\tau$ depends on the level of confidence the teacher has on train. ResNet56 networks achieve nearly 100\% accuracy on train with high confidence, so a temperature like $\tau=4$ works well. 
When ResNet20 networks are used (networks which are not capable of perfectly fitting CIFAR-100), we see that lower temperatures can be used, although $\tau = 4$ still outperforms other choices (Figure \ref{main/fig:resnet_20_a_b}). 
The reason lower temperatures work with ResNet20 teachers on CIFAR-100 is because ResNet20 networks do not attain 100\% accuracy on train, so the teacher logits are much less sharply peaked (see also Figure \ref{supp/fig:teacher_predictions}, top row).

\begin{figure}[h]
\centering
\includegraphics[width=0.45\textwidth]{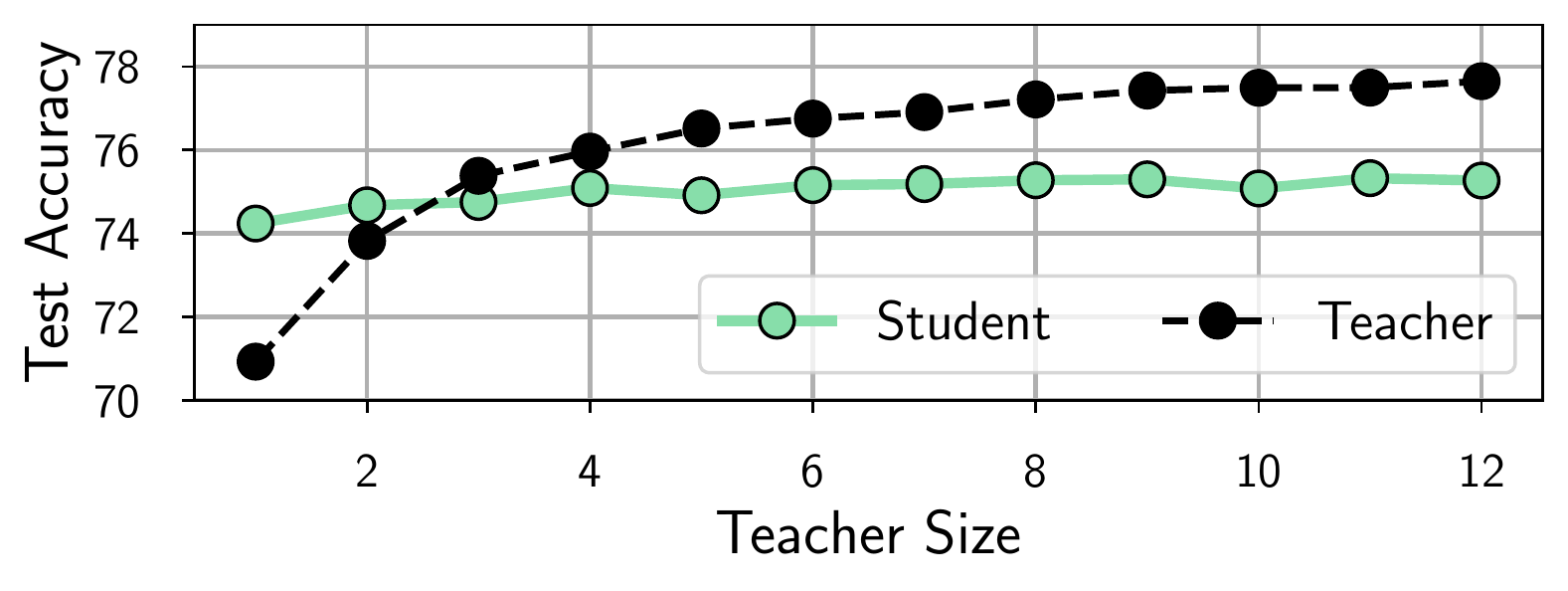}
\quad
\includegraphics[width=0.45\textwidth]{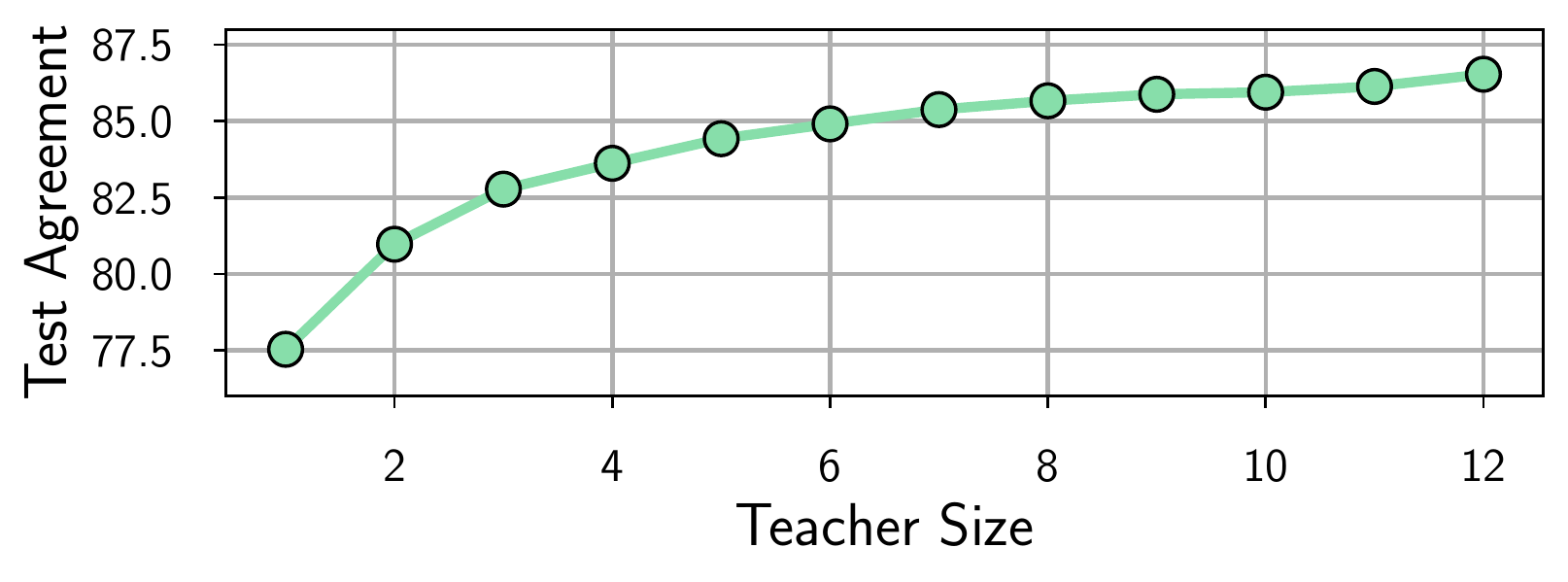}
\caption{The effect on accuracy \textbf{(left)} and agreement \textbf{(right)} of the number of models ($m$) in the teacher ensemble ($\alpha=0, \; \tau=1$). Student accuracy quickly saturates as $m$ increases, despite continuing improvements in teacher accuracy. The teacher-student agreement continues to improve after the accuracy has saturated.}
\label{supp/fig:num_teacher_components}
\end{figure}

\subsection{Understanding the effect of teacher ensemble size on distillation}
\label{supp/subsec:ensemble_size_ablation}

In Figure \ref{supp/fig:num_teacher_components} we demonstrate the effect of increasing the number ($m$) of teacher ensemble ResNet56 components on test accuracy and agreement. 
In the main text we only considered teacher ensembles with up to 5 components -- here we provide results for up to 12 components. 
Although it is plausible that ensembles with more components would have more complex predictive distributions that would be difficult for a single student to match, in reality we see the exact opposite. 
Deep ensembles with \textit{more} components are easier to emulate (indicated by higher agreement). 
One possible explanation is that adding more ensemble components smooths the logits of unlikely classes, making the distribution easier to match. 
Closer investigation into this phenomenon could potentially yield insights into how to improve distillation fidelity in general. 

In agreement with the results on self-distillation \citep{furlanello2018born, mobahi2020self}, 
we see that the student is more accurate than the teacher when $m=1$.
However the accuracy of the student does not substantially improve as we increase the number $m$ of models in the teacher ensemble past 4, even though both the accuracy of the teacher and the teacher-student agreement continue to increase monotonically with $m$.

\subsection{Detailed results for distillation with heavy data augmentation}
\label{supp/subsec:detailed_aug_results}

\begin{figure}[h]
\vspace{-6mm}
\centering
\begin{tabular}{ccccc}
\includegraphics[height=0.12\textwidth]{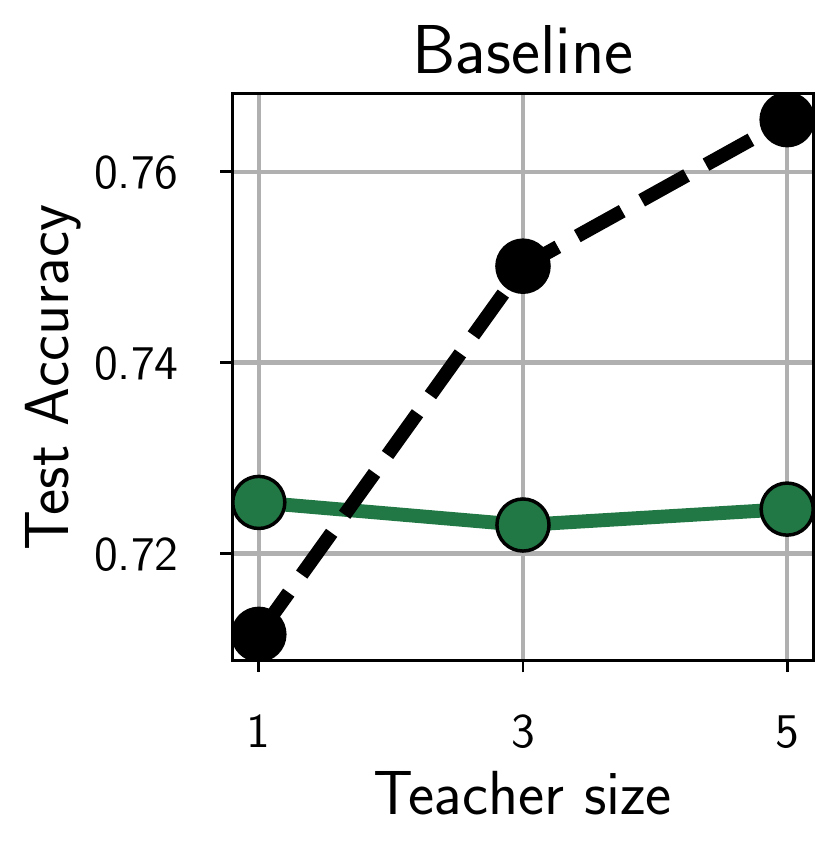} &
\includegraphics[height=0.12\textwidth]{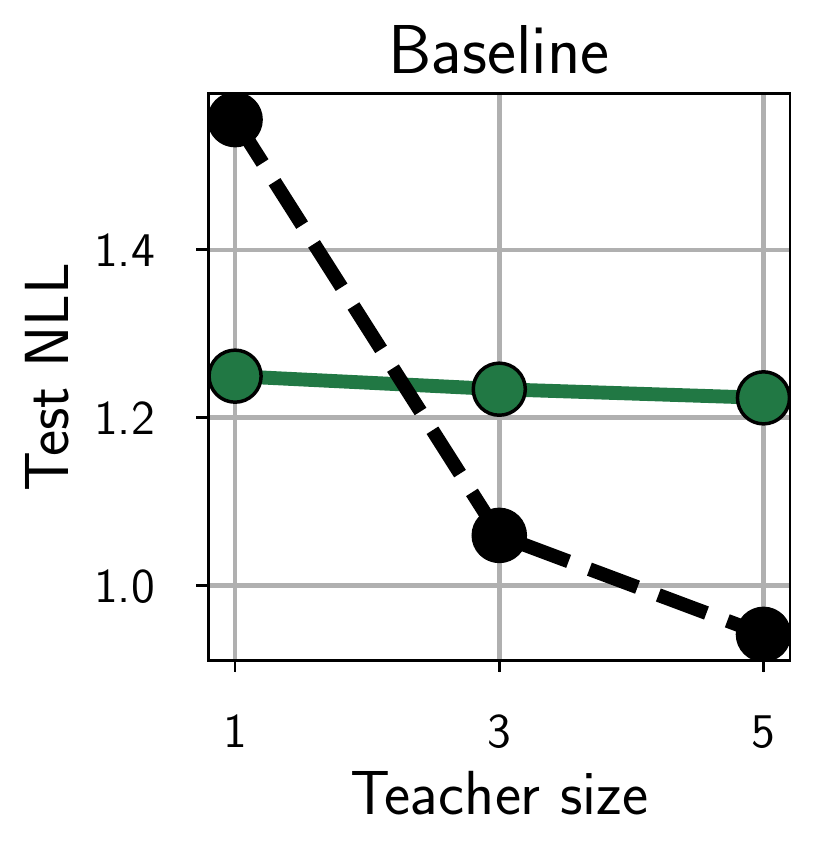} &
\includegraphics[height=0.12\textwidth]{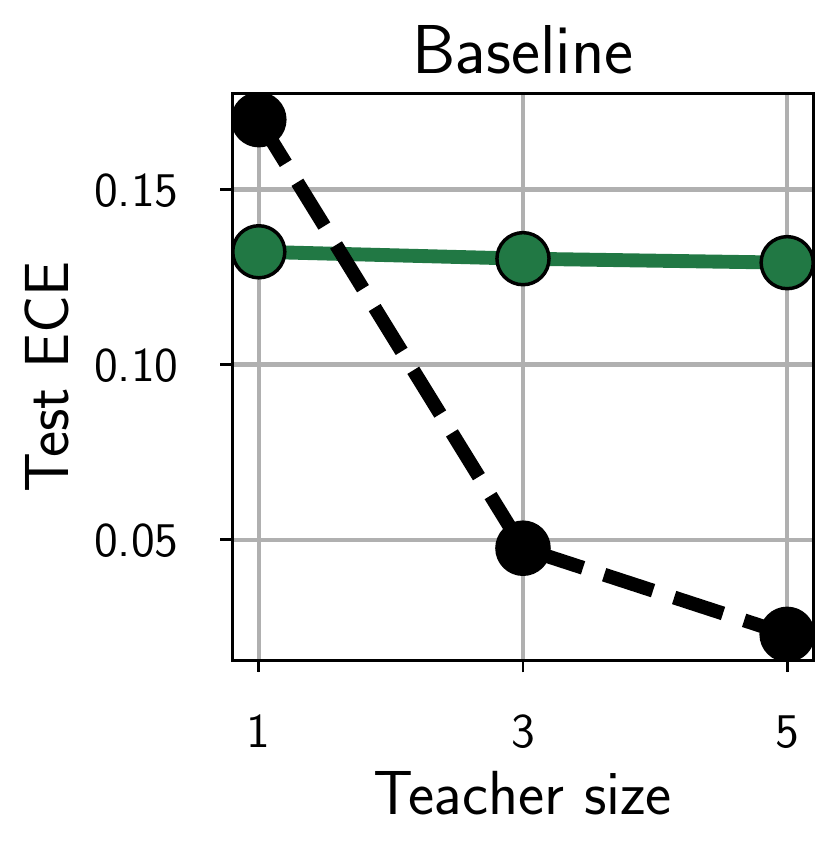} &
\includegraphics[height=0.12\textwidth]{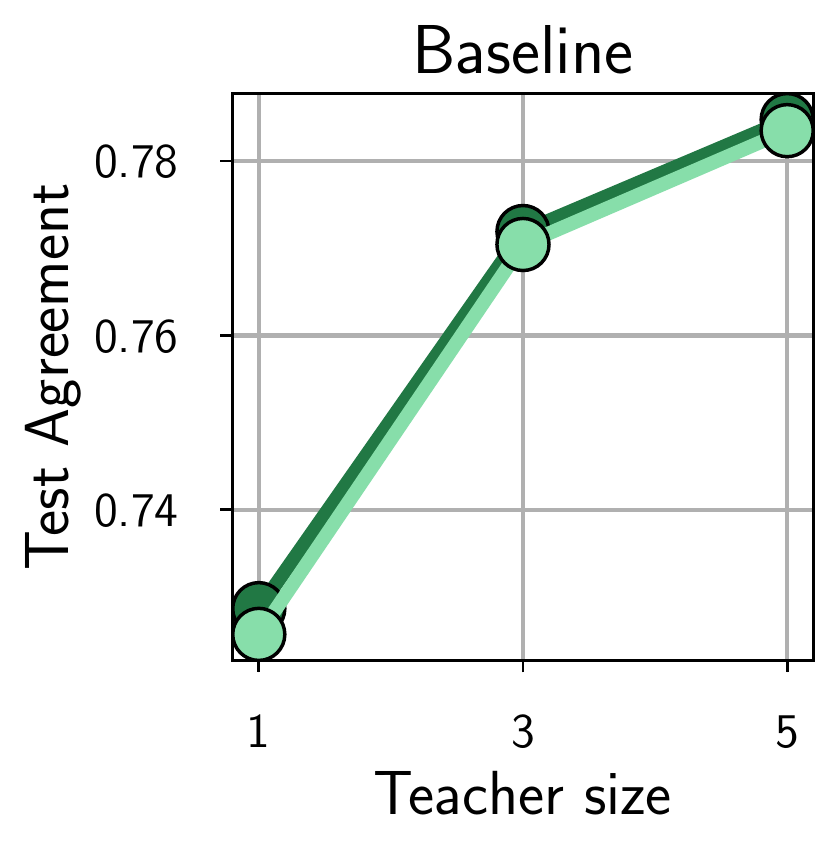} &
\includegraphics[height=0.12\textwidth]{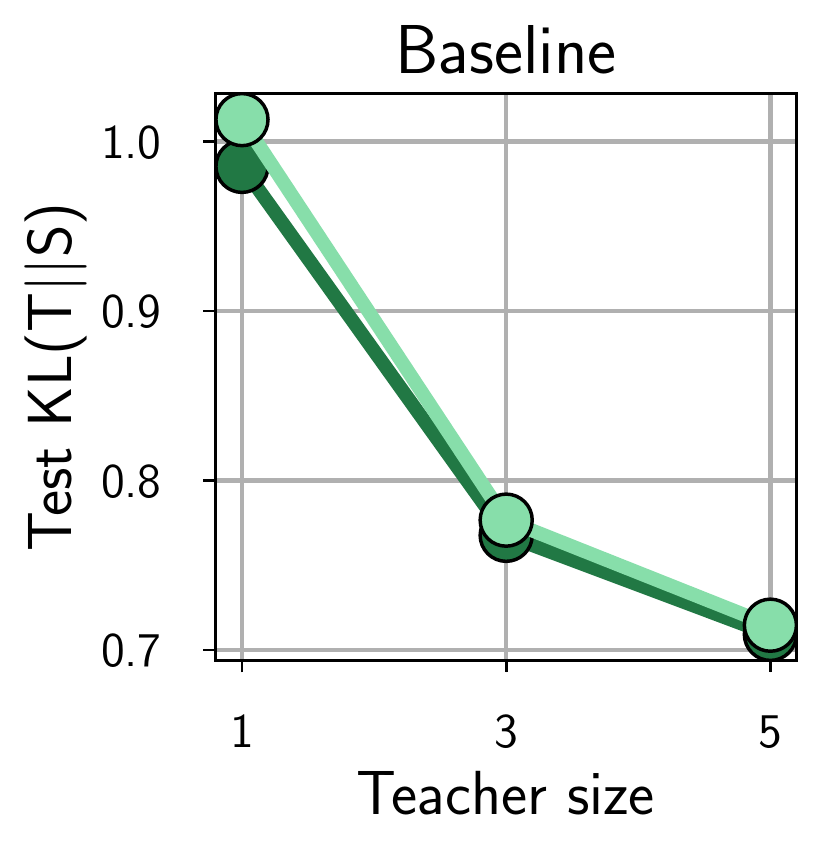}
\\[-0.2cm]
\includegraphics[height=0.12\textwidth]{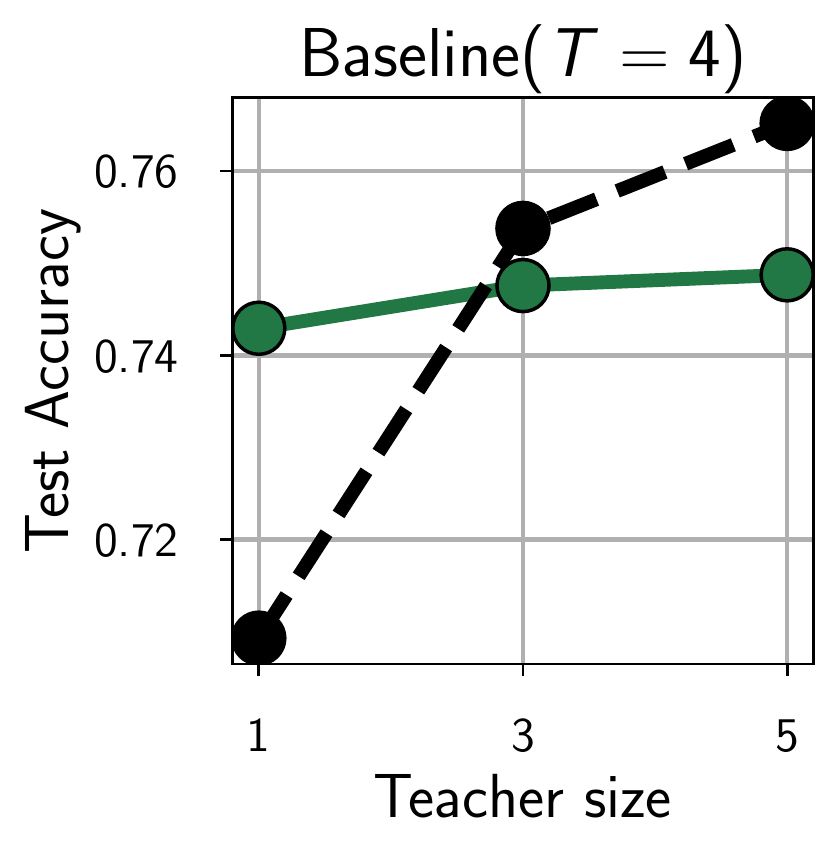} &
\includegraphics[height=0.12\textwidth]{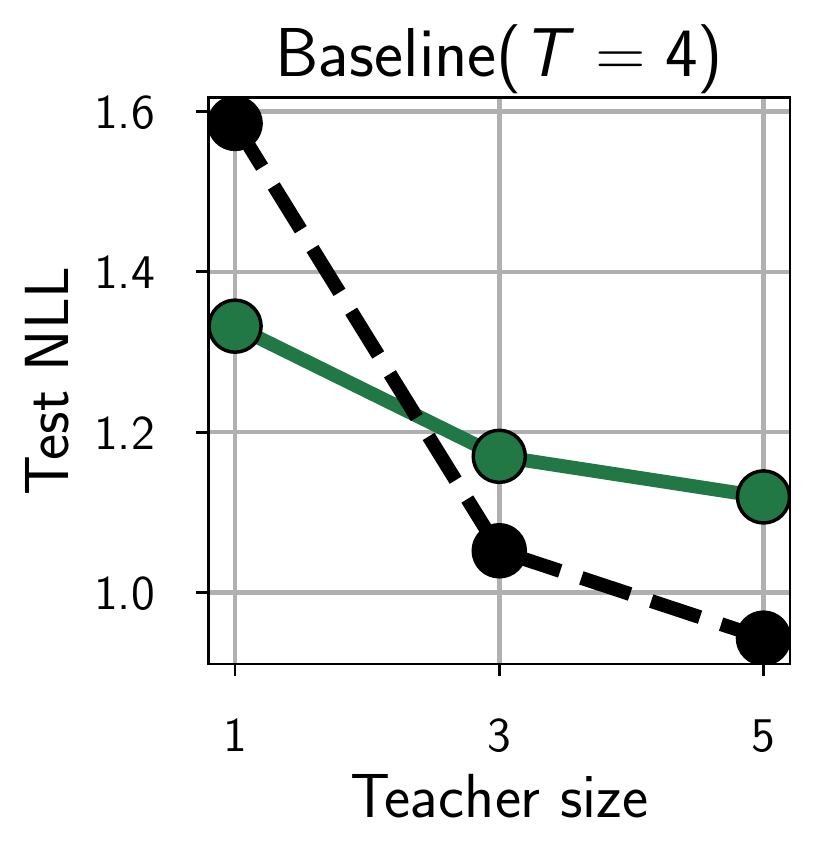} &
\includegraphics[height=0.12\textwidth]{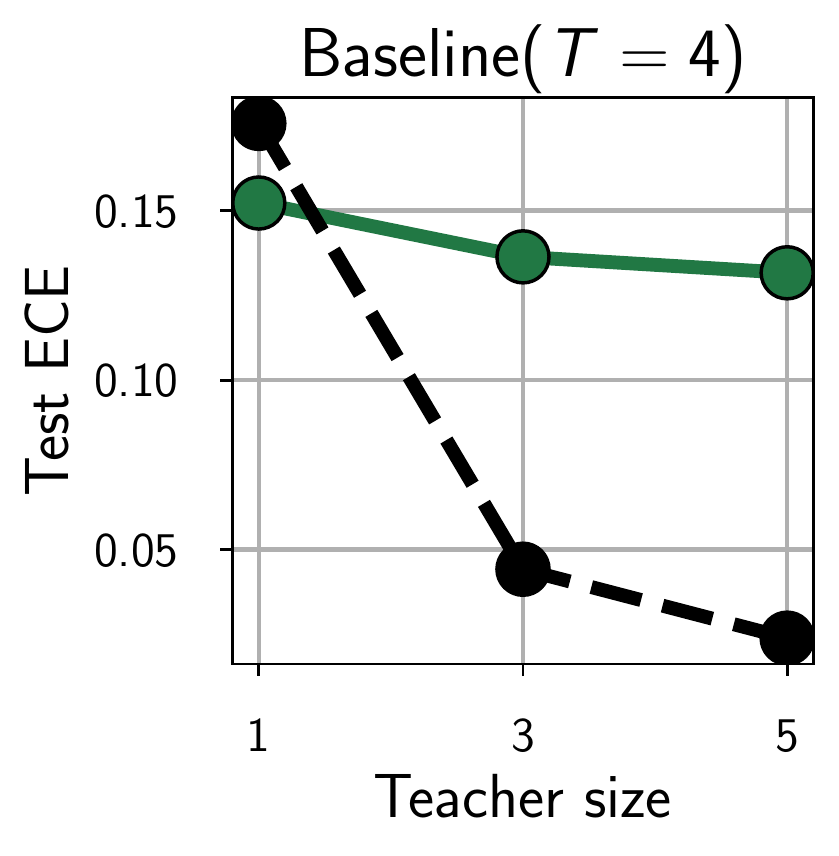} &
\includegraphics[height=0.12\textwidth]{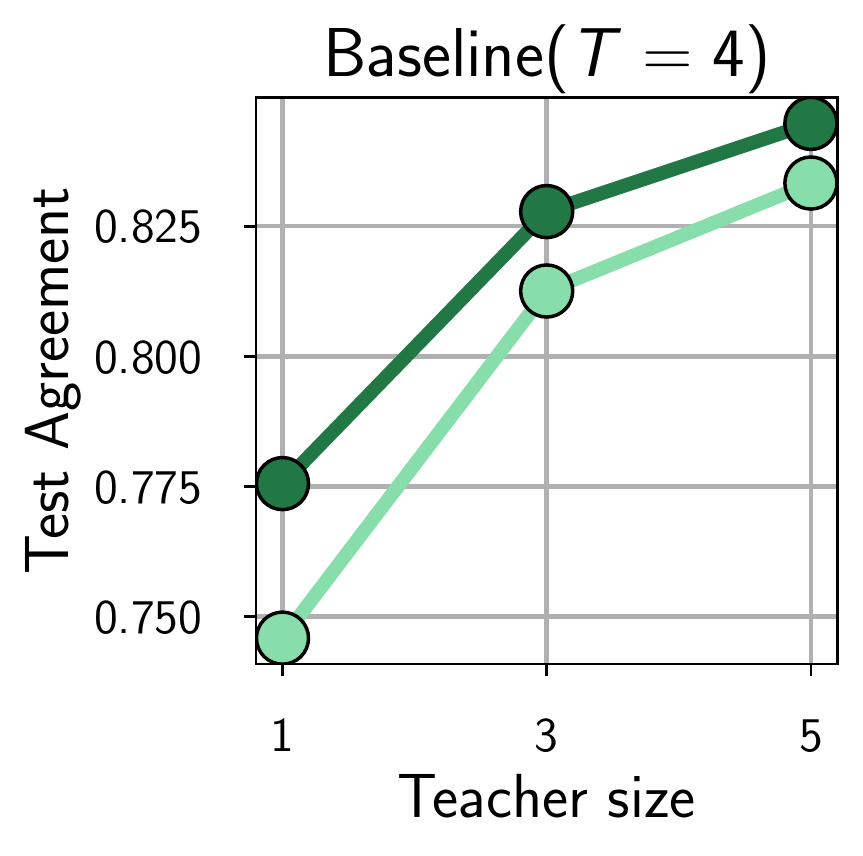} &
\includegraphics[height=0.12\textwidth]{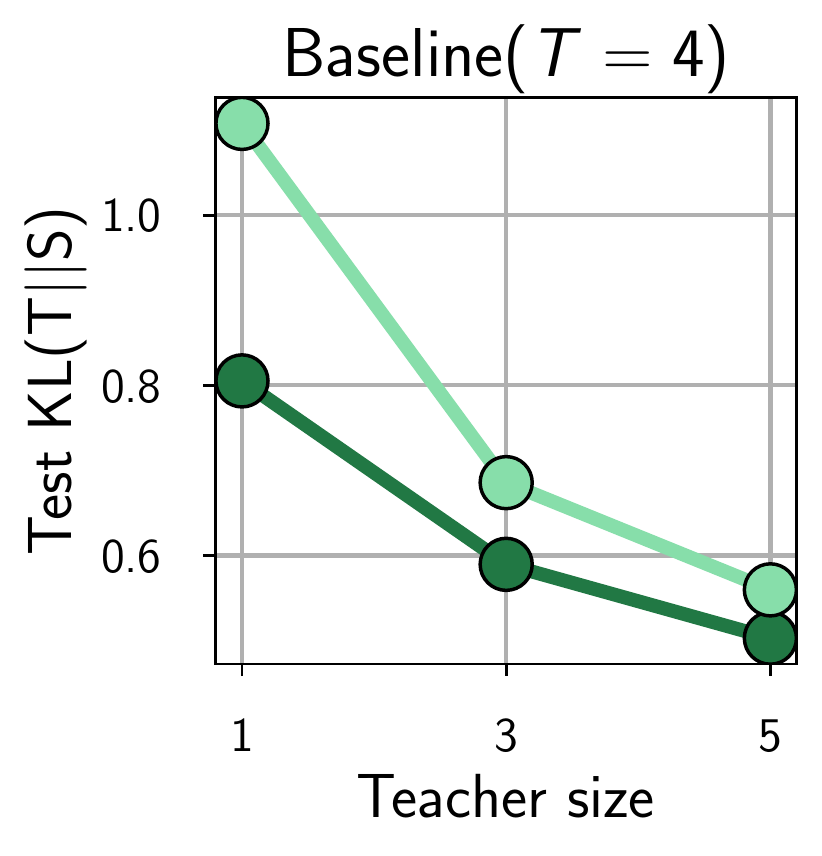}
\\[-0.2cm]
\includegraphics[height=0.12\textwidth]{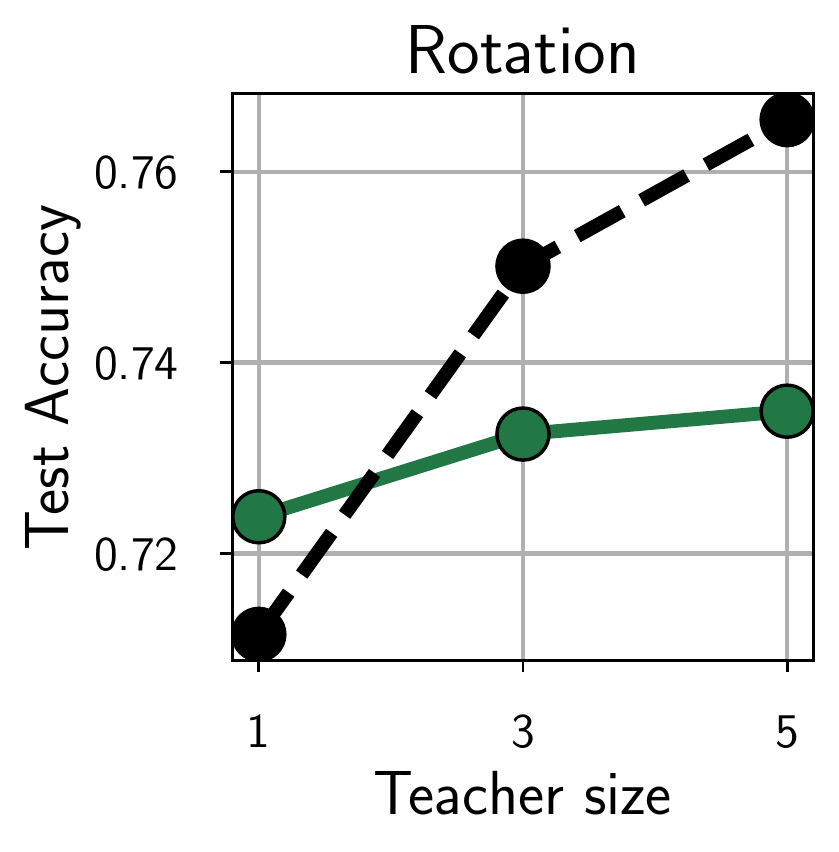} &
\includegraphics[height=0.12\textwidth]{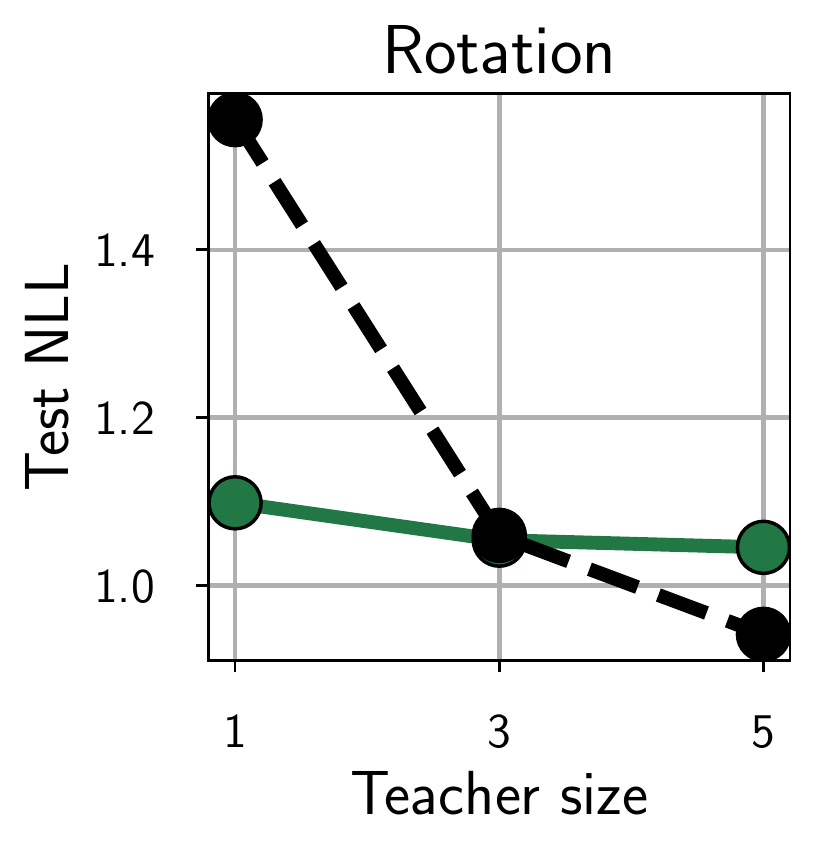} &
\includegraphics[height=0.12\textwidth]{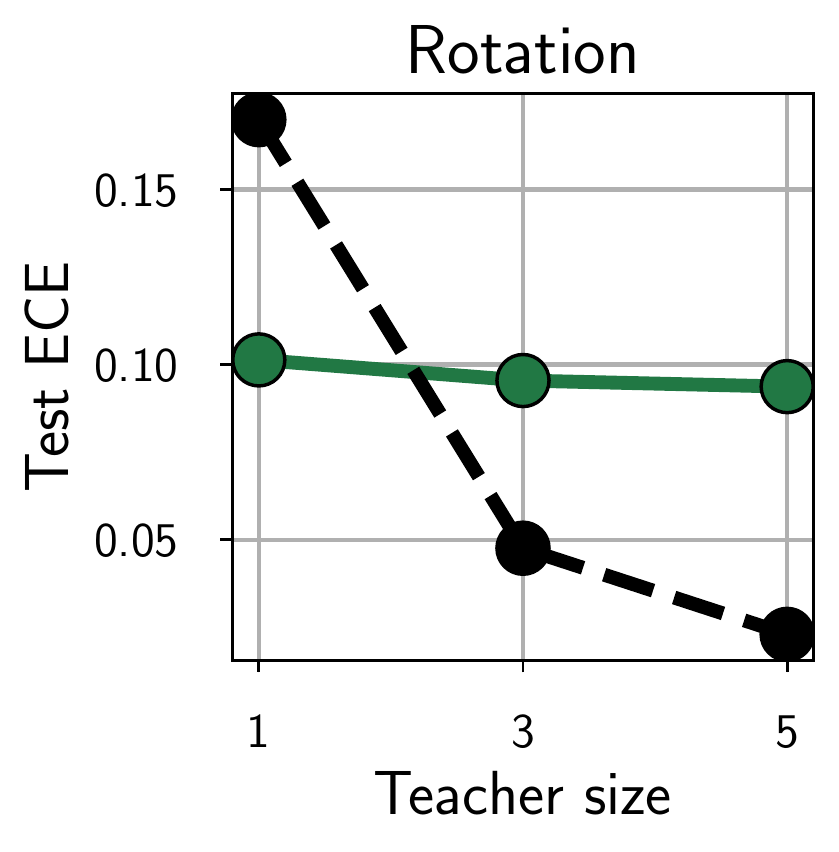} &
\includegraphics[height=0.12\textwidth]{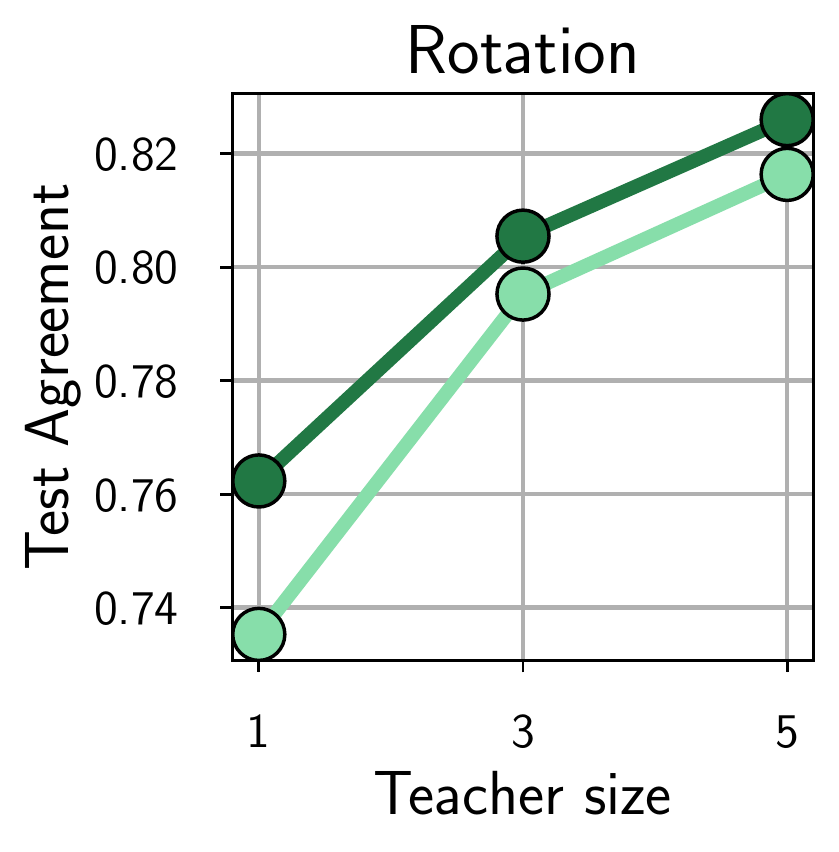} &
\includegraphics[height=0.12\textwidth]{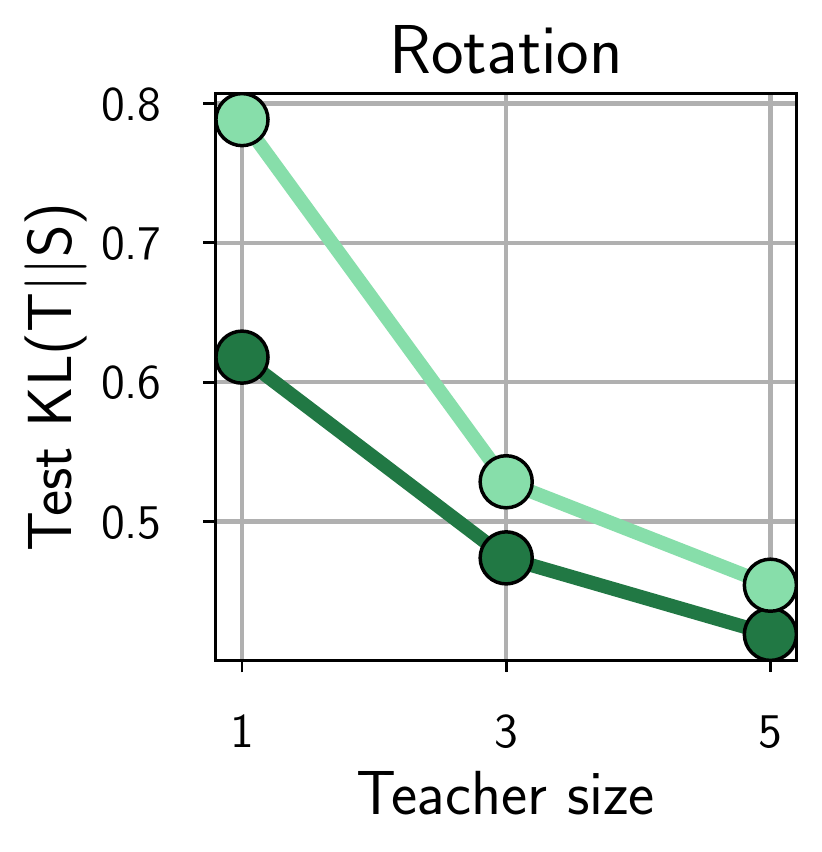}
\\[-0.2cm]
\includegraphics[height=0.12\textwidth]{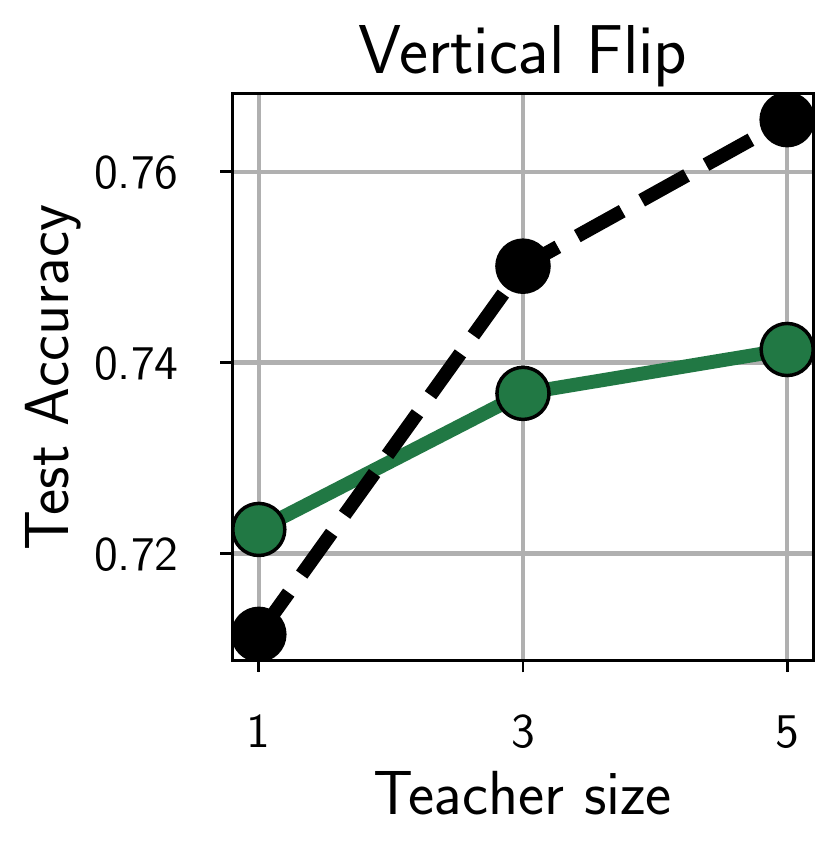} &
\includegraphics[height=0.12\textwidth]{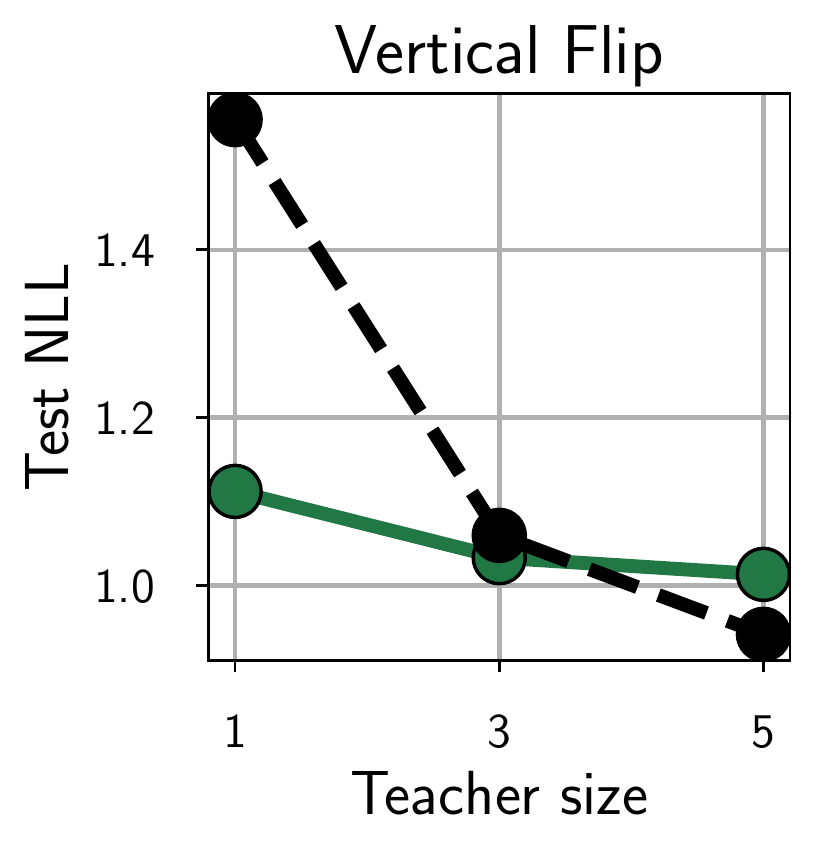} &
\includegraphics[height=0.12\textwidth]{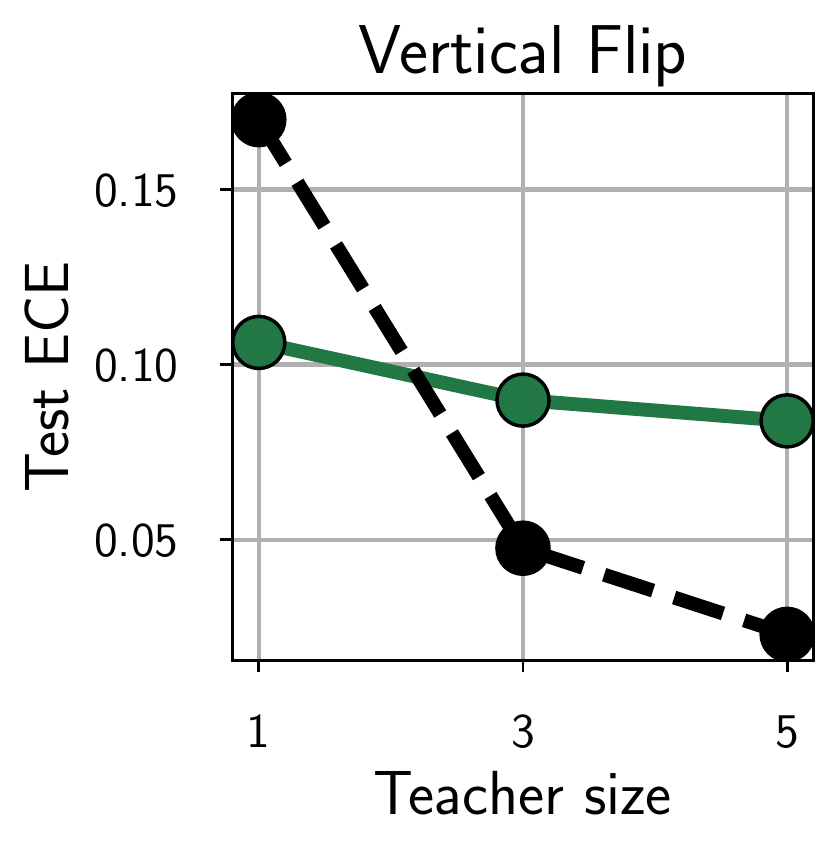} &
\includegraphics[height=0.12\textwidth]{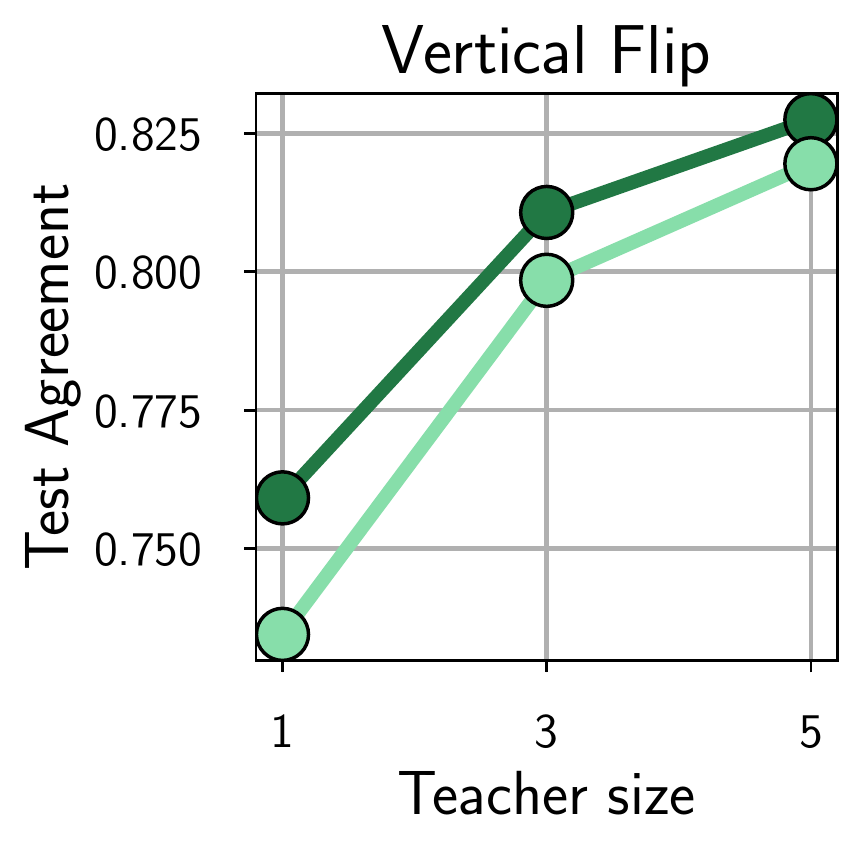} &
\includegraphics[height=0.12\textwidth]{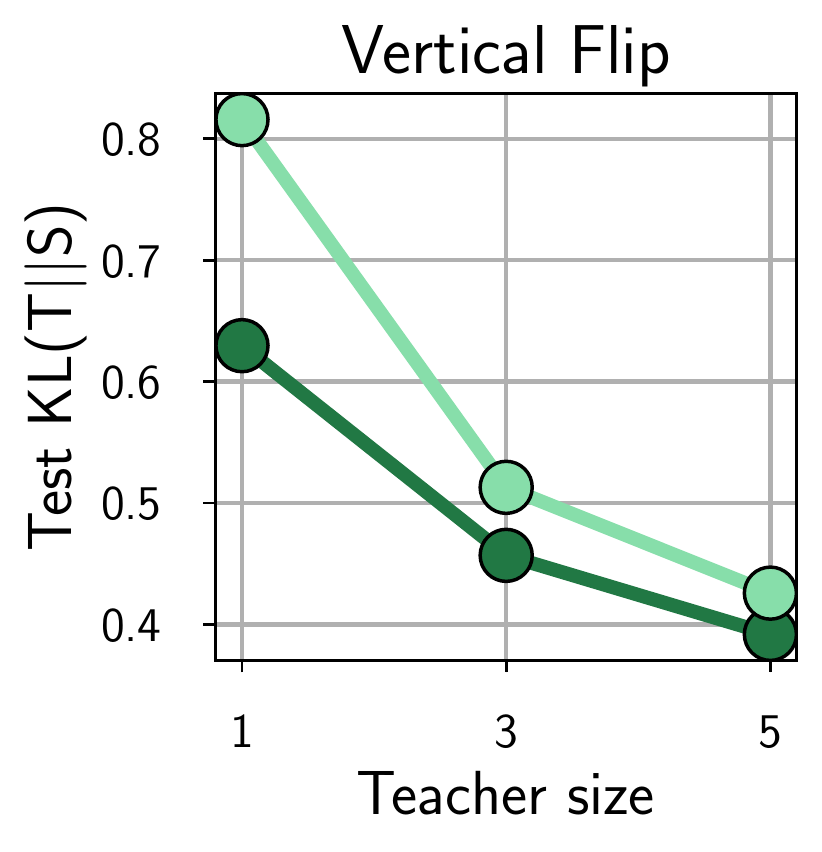}
\\[-0.2cm]
\includegraphics[height=0.12\textwidth]{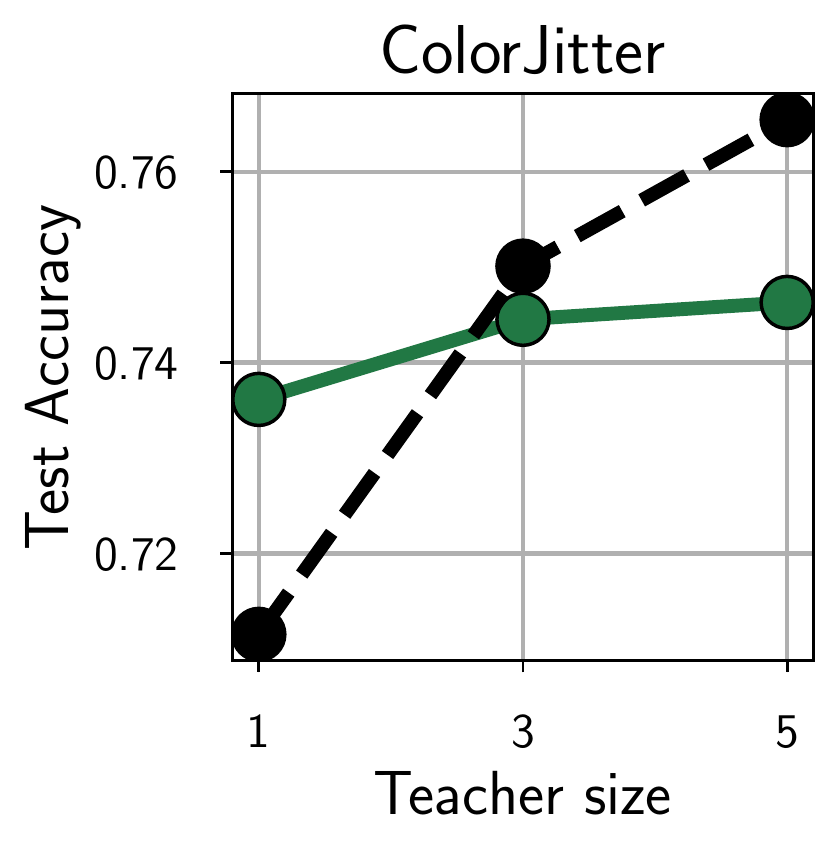} &
\includegraphics[height=0.12\textwidth]{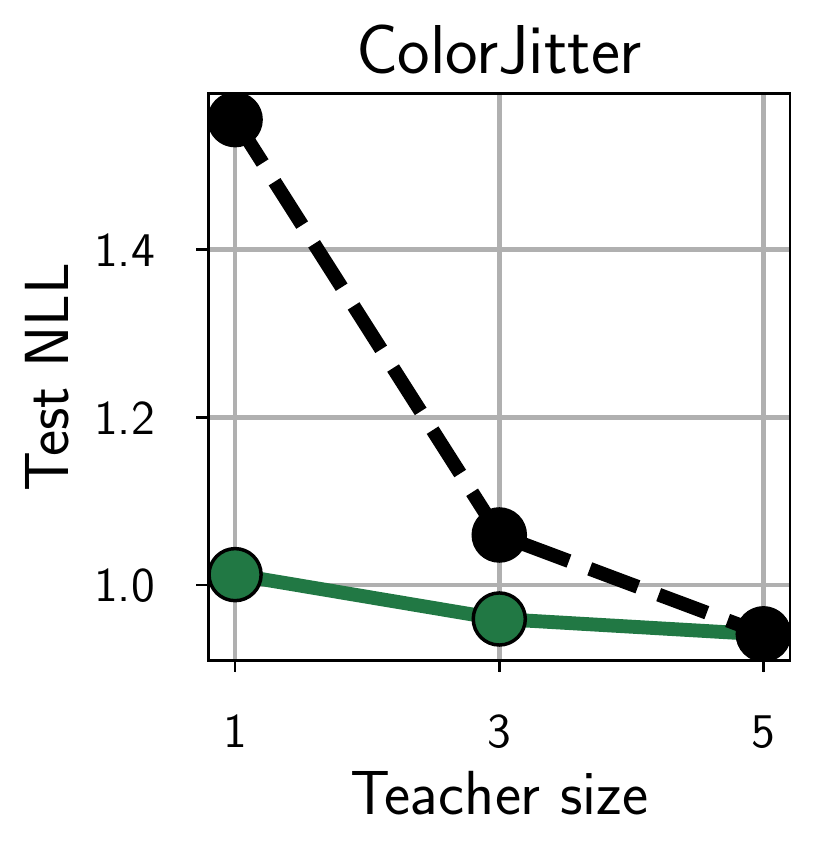} &
\includegraphics[height=0.12\textwidth]{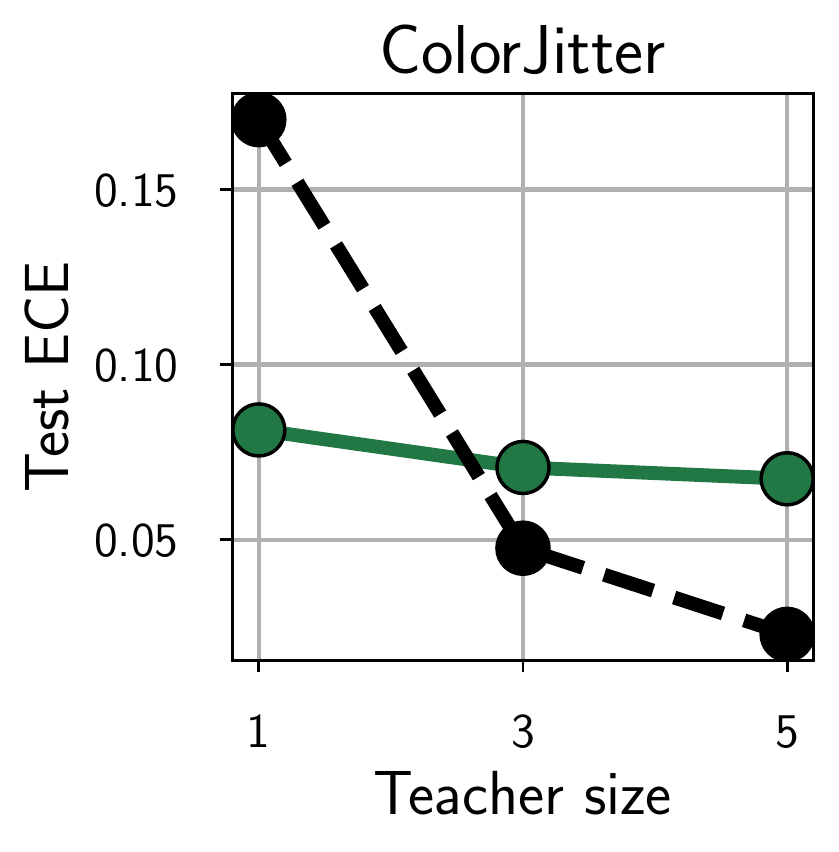} &
\includegraphics[height=0.12\textwidth]{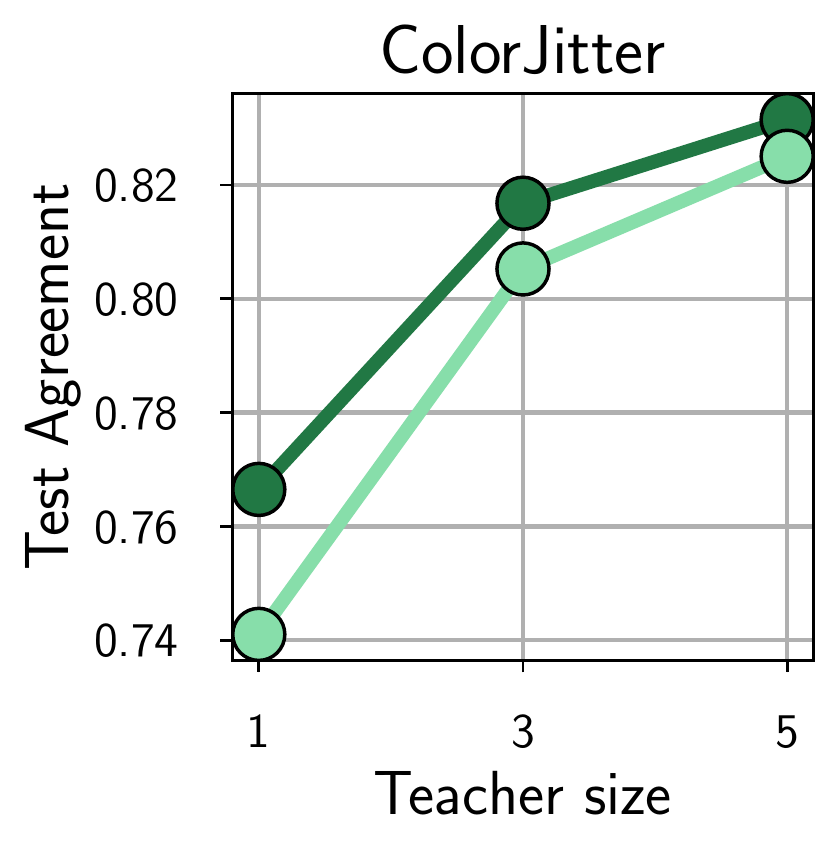} &
\includegraphics[height=0.12\textwidth]{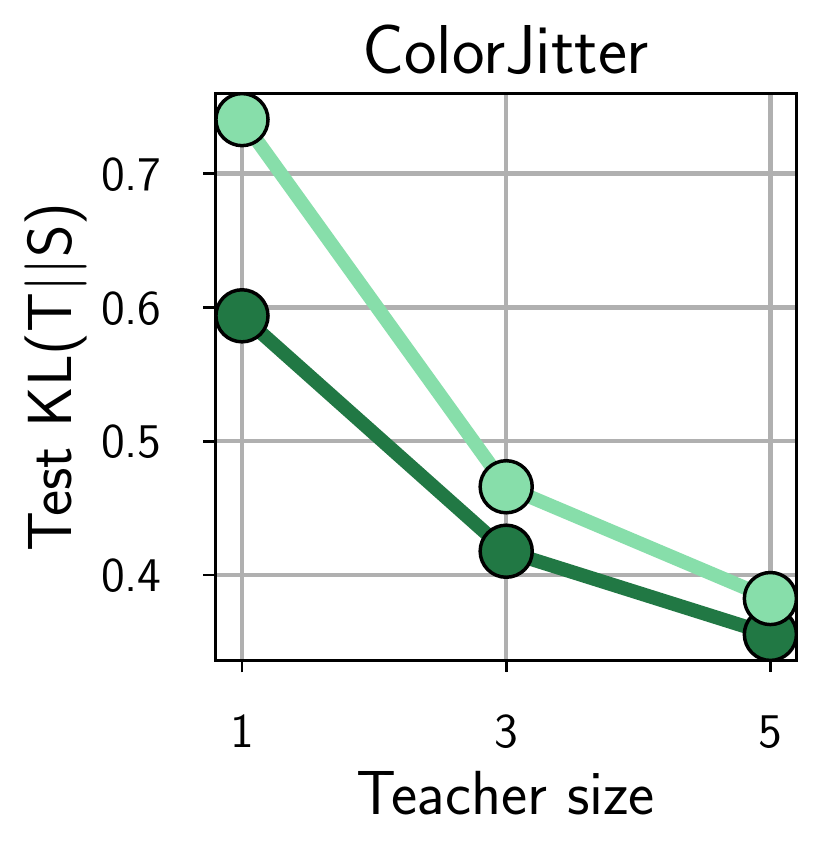}
\\[-0.2cm]
\includegraphics[height=0.12\textwidth]{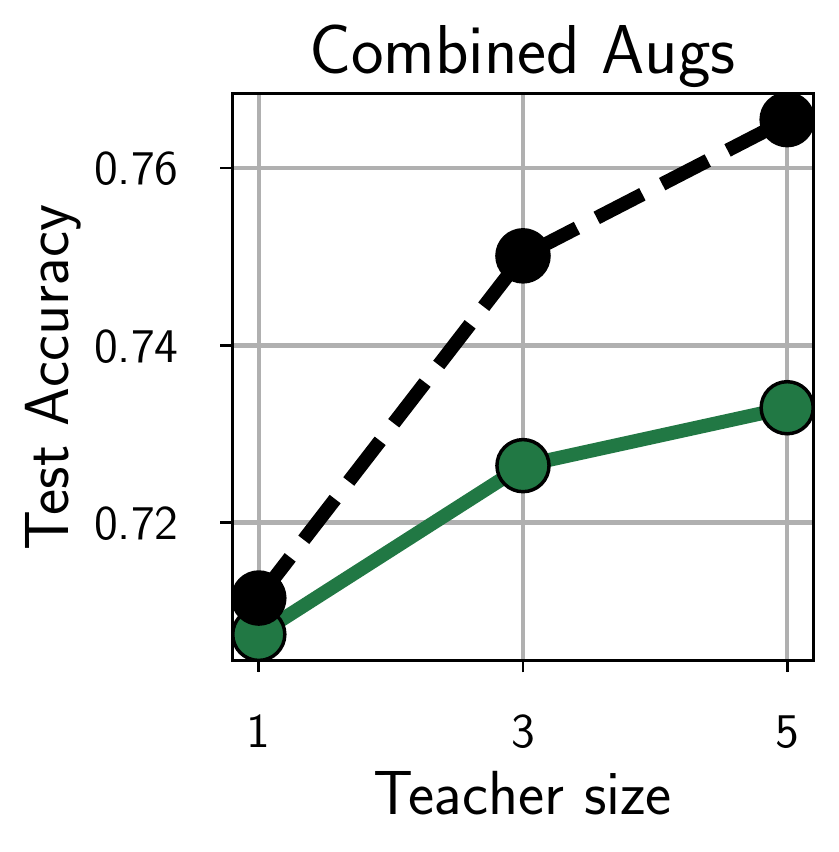} &
\includegraphics[height=0.12\textwidth]{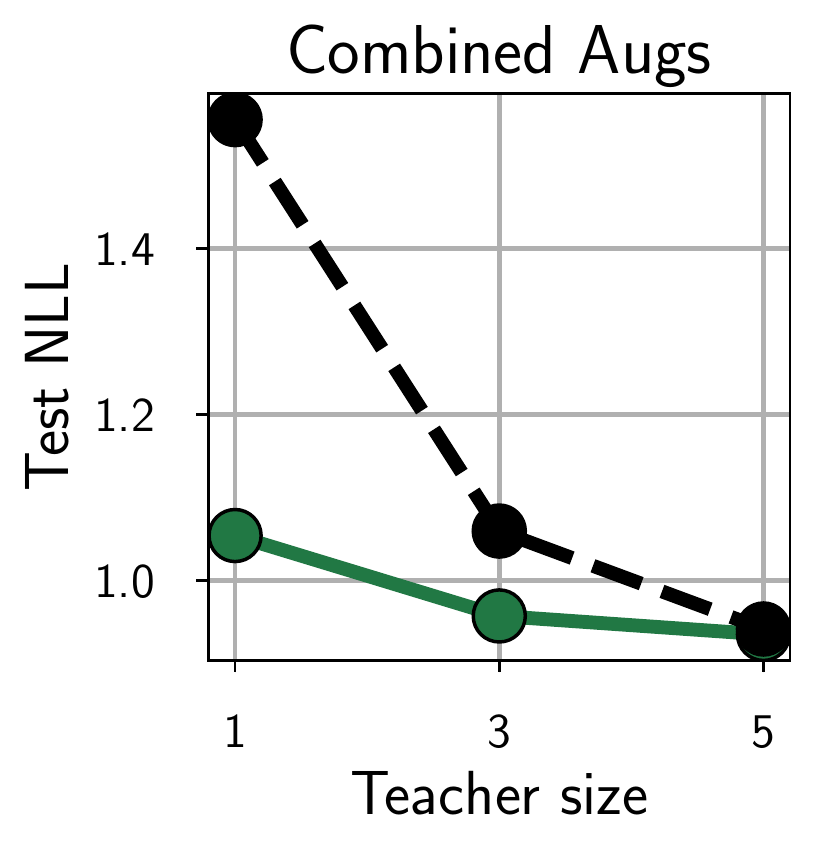} &
\includegraphics[height=0.12\textwidth]{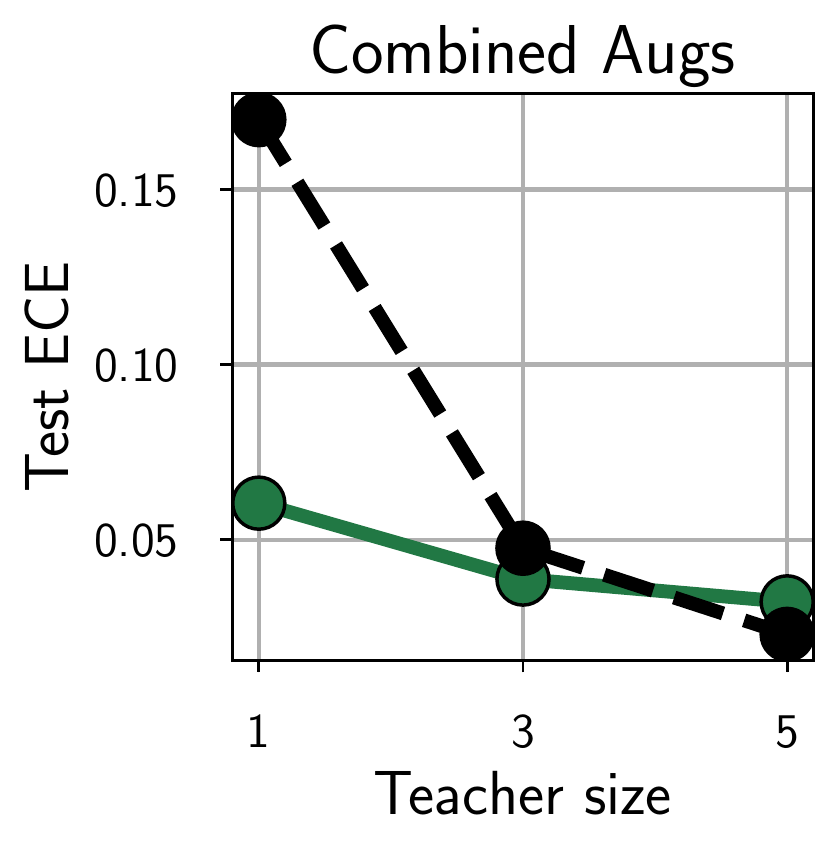} &
\includegraphics[height=0.12\textwidth]{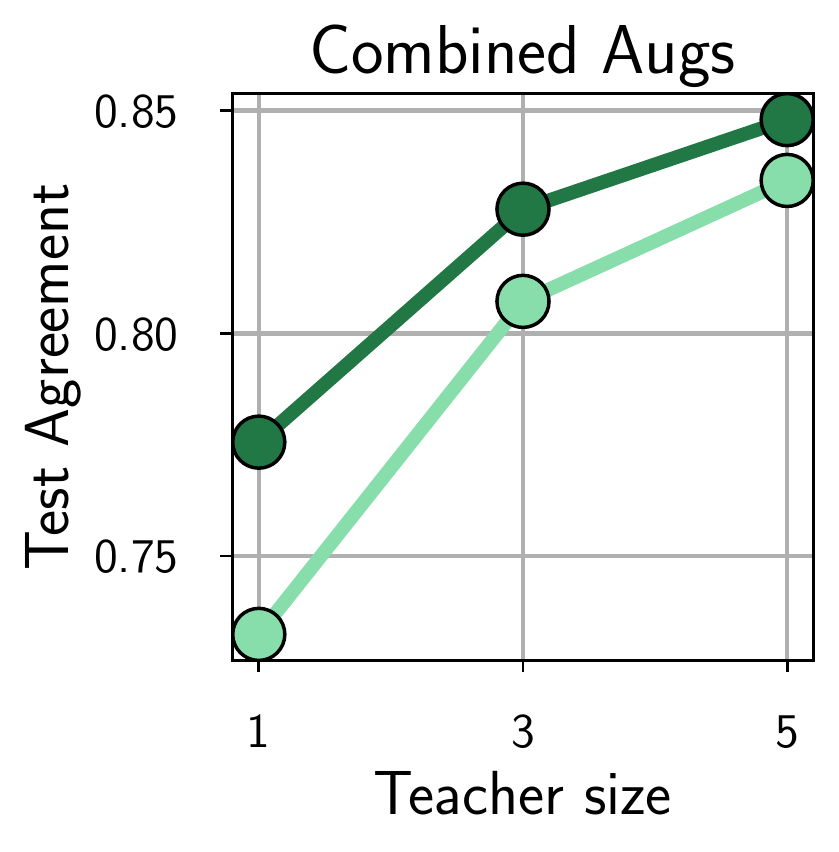} &
\includegraphics[height=0.12\textwidth]{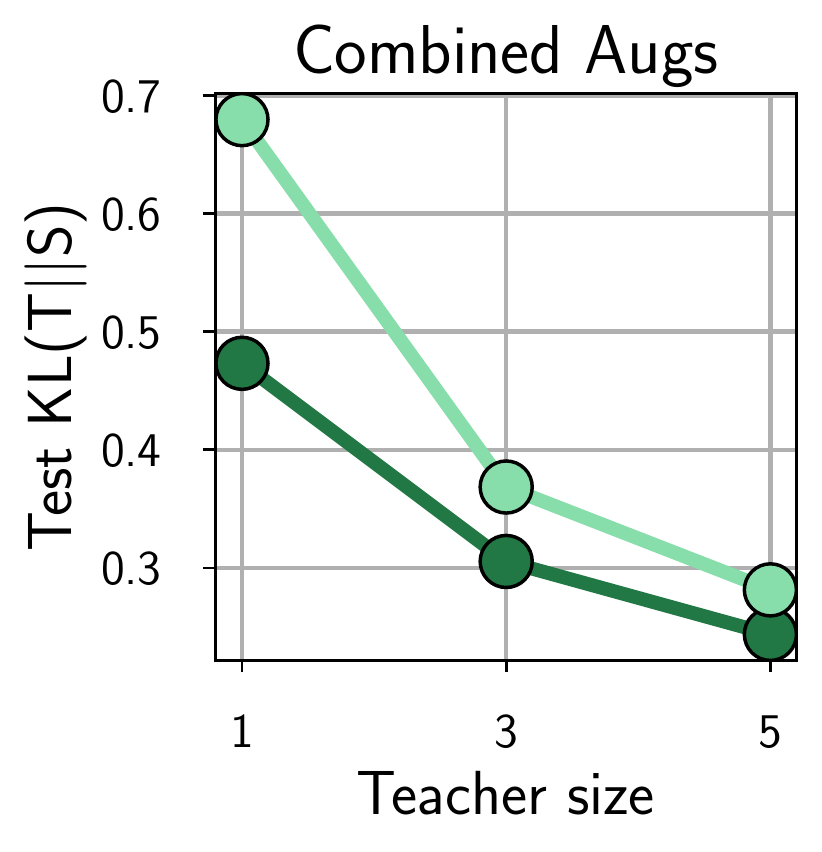}
\\[-0.2cm]
\includegraphics[height=0.12\textwidth]{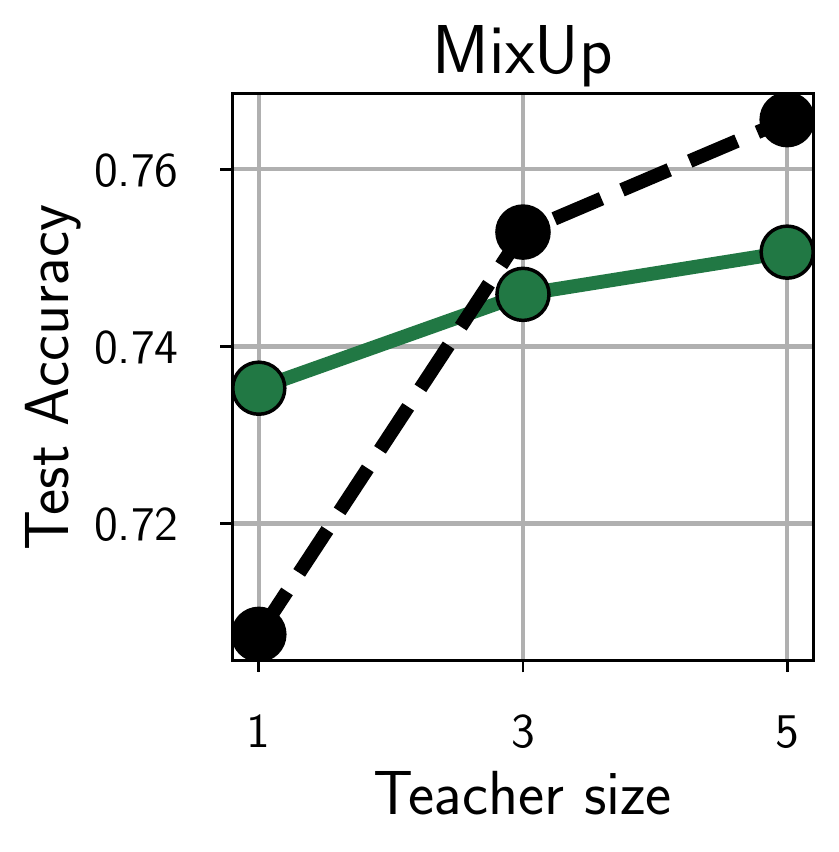} &
\includegraphics[height=0.12\textwidth]{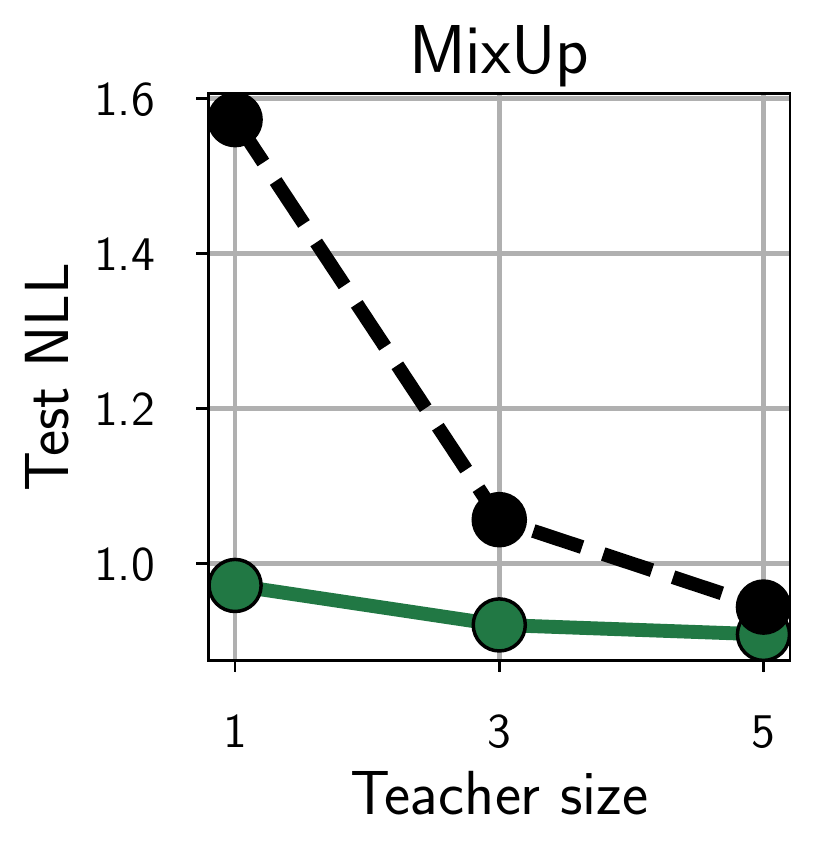} &
\includegraphics[height=0.12\textwidth]{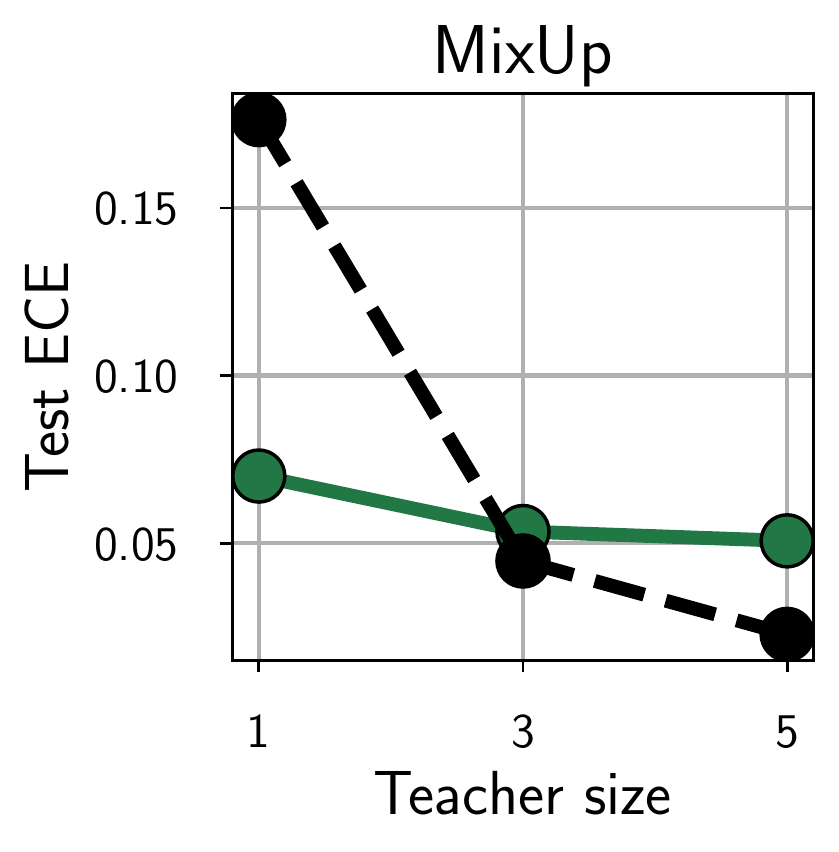} &
\includegraphics[height=0.12\textwidth]{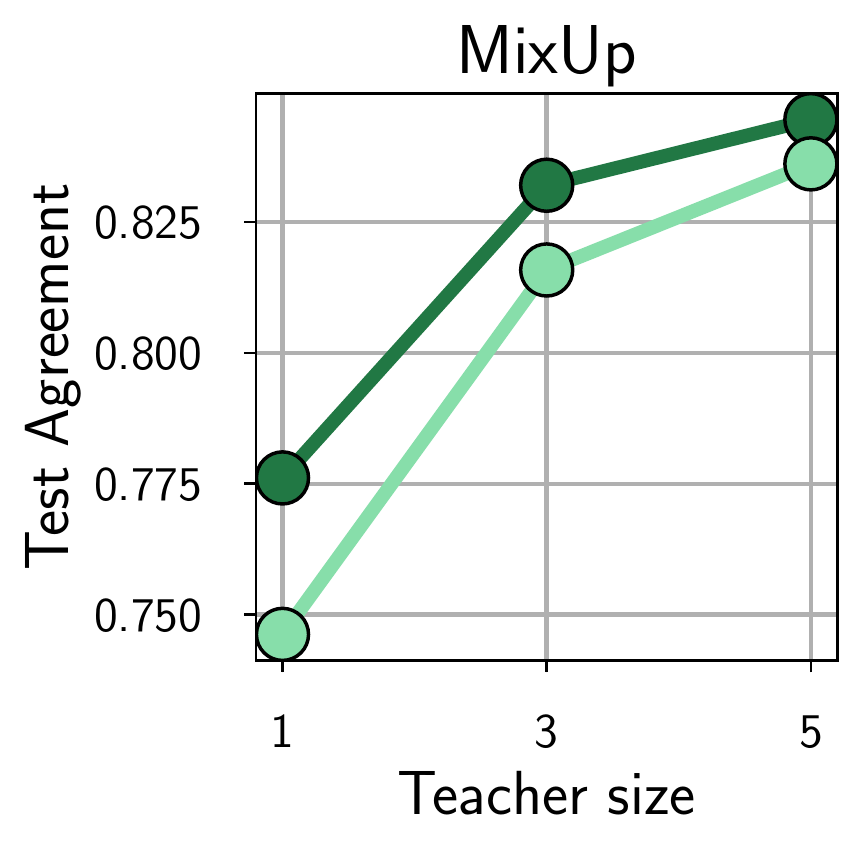} &
\includegraphics[height=0.12\textwidth]{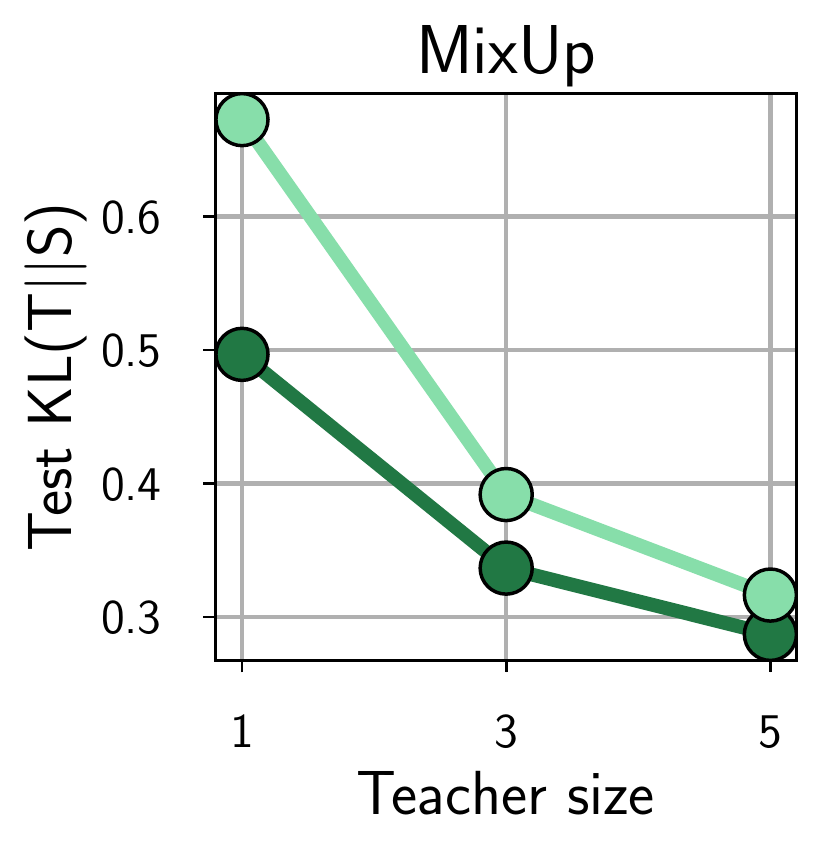}
\\[-0.2cm]
\includegraphics[height=0.12\textwidth]{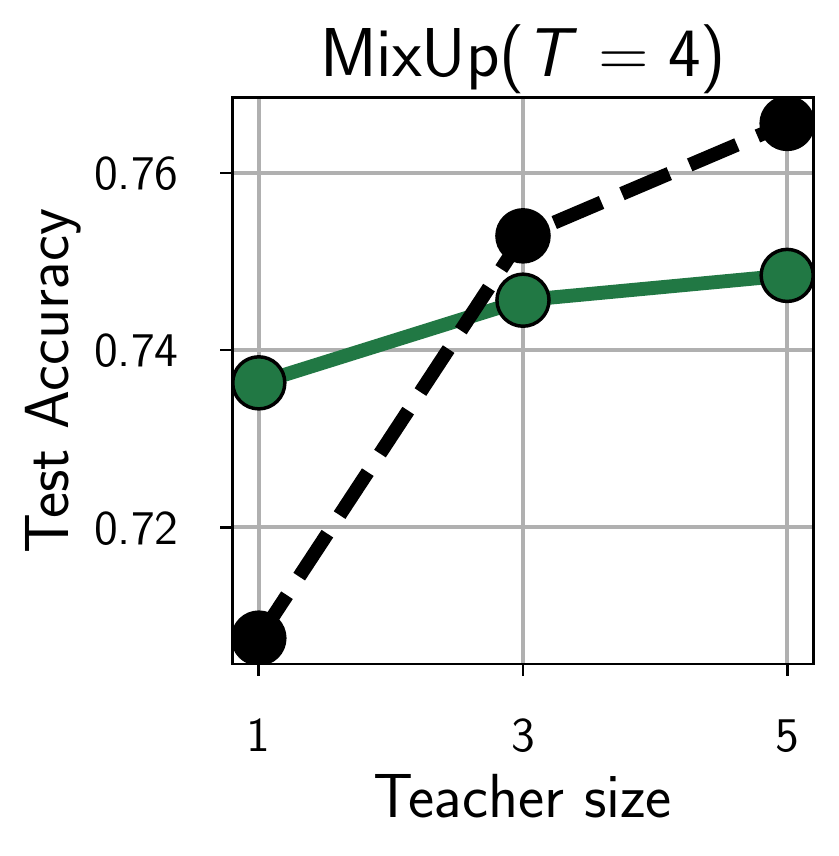} &
\includegraphics[height=0.12\textwidth]{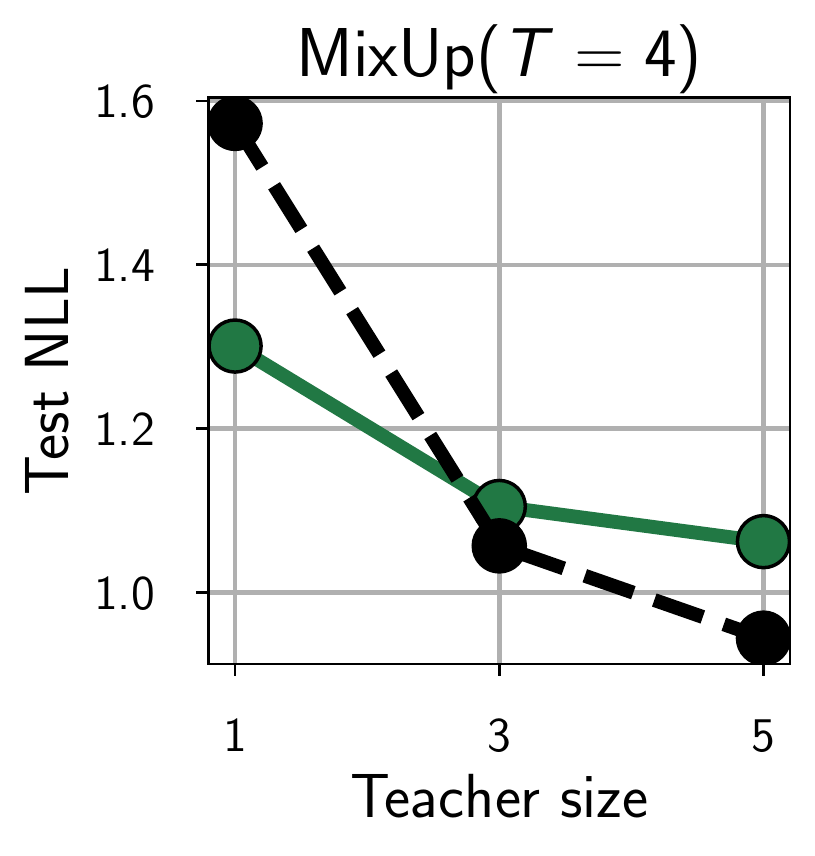} &
\includegraphics[height=0.12\textwidth]{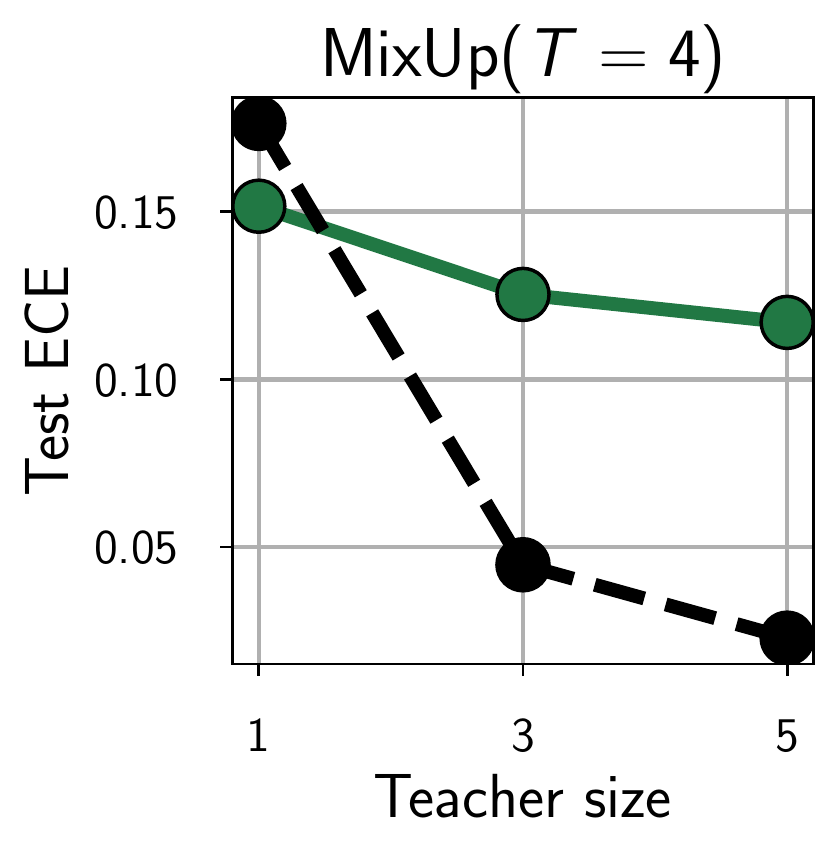} &
\includegraphics[height=0.12\textwidth]{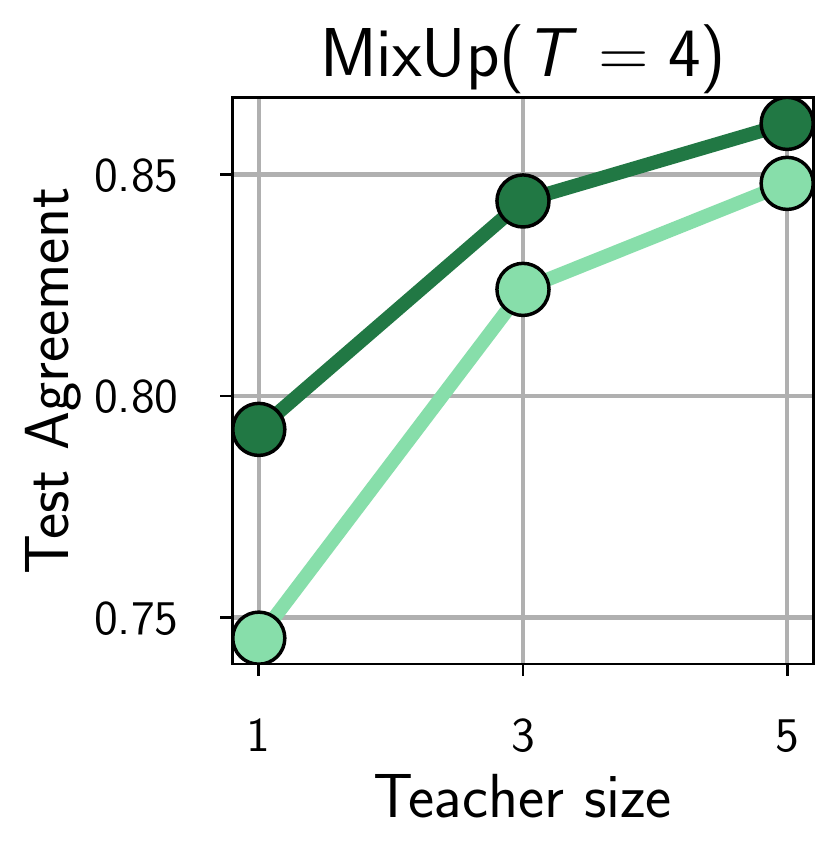} &
\includegraphics[height=0.12\textwidth]{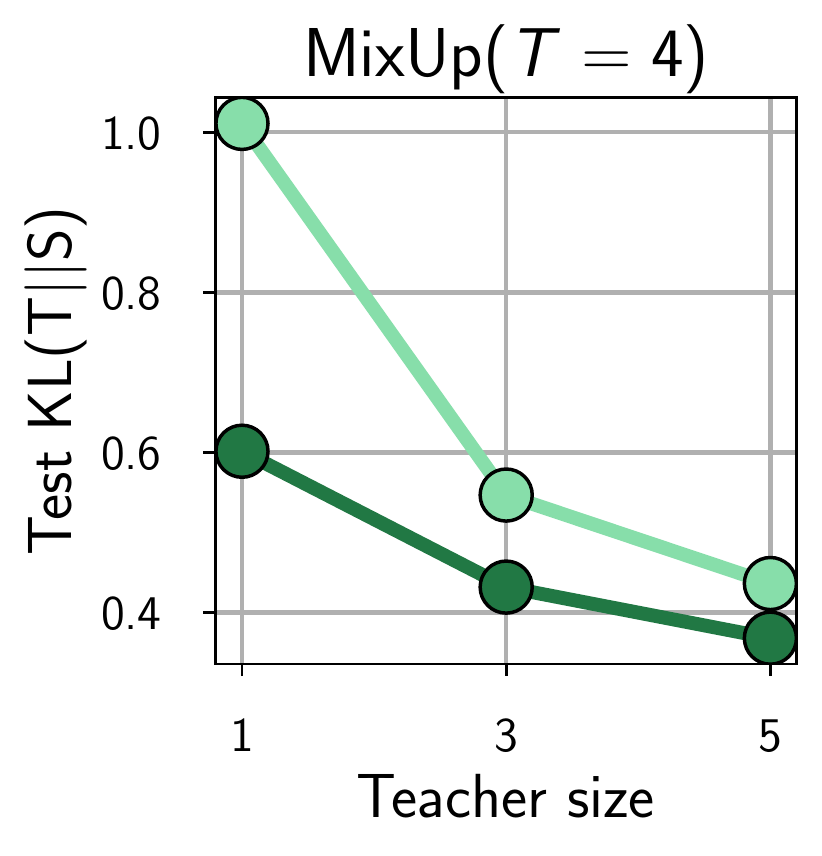}
\\[-0.2cm]
\includegraphics[height=0.12\textwidth]{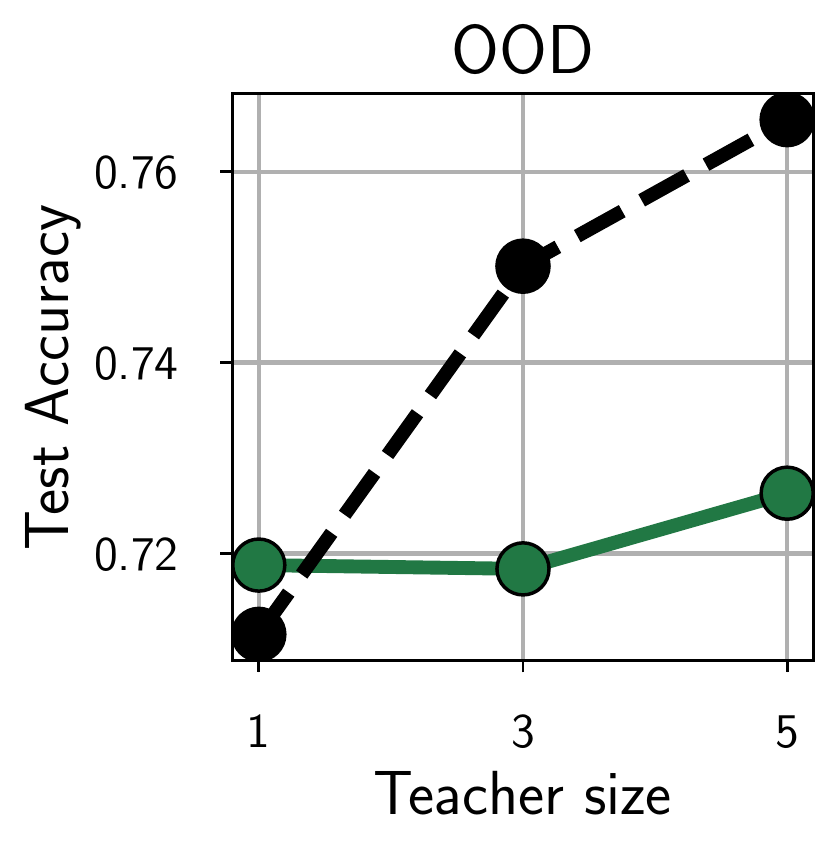} &
\includegraphics[height=0.12\textwidth]{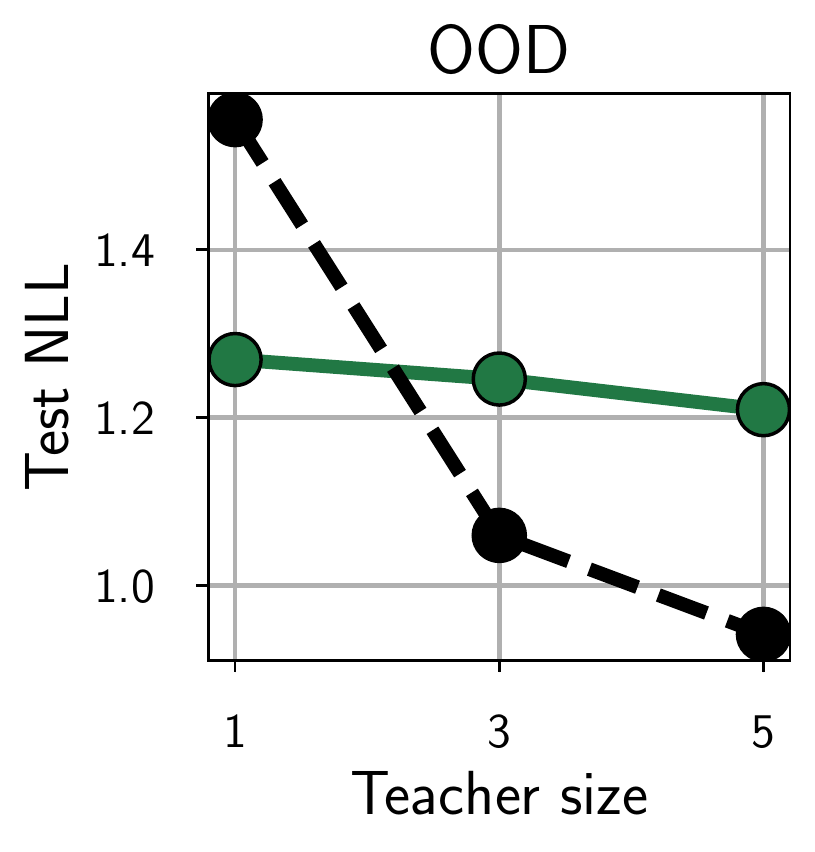} &
\includegraphics[height=0.12\textwidth]{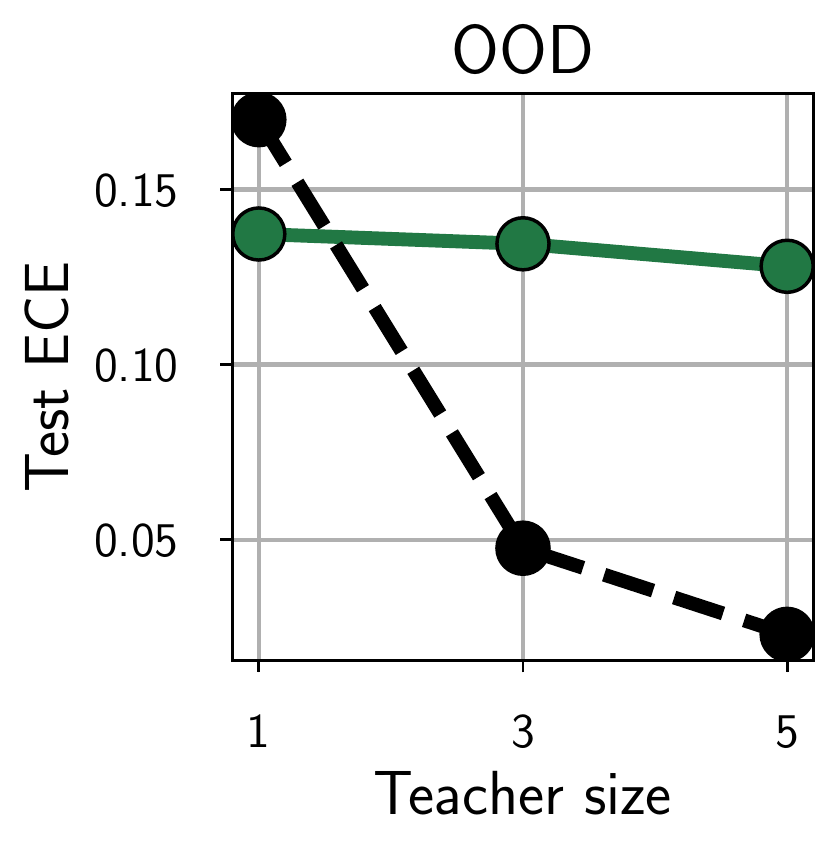} &
\includegraphics[height=0.12\textwidth]{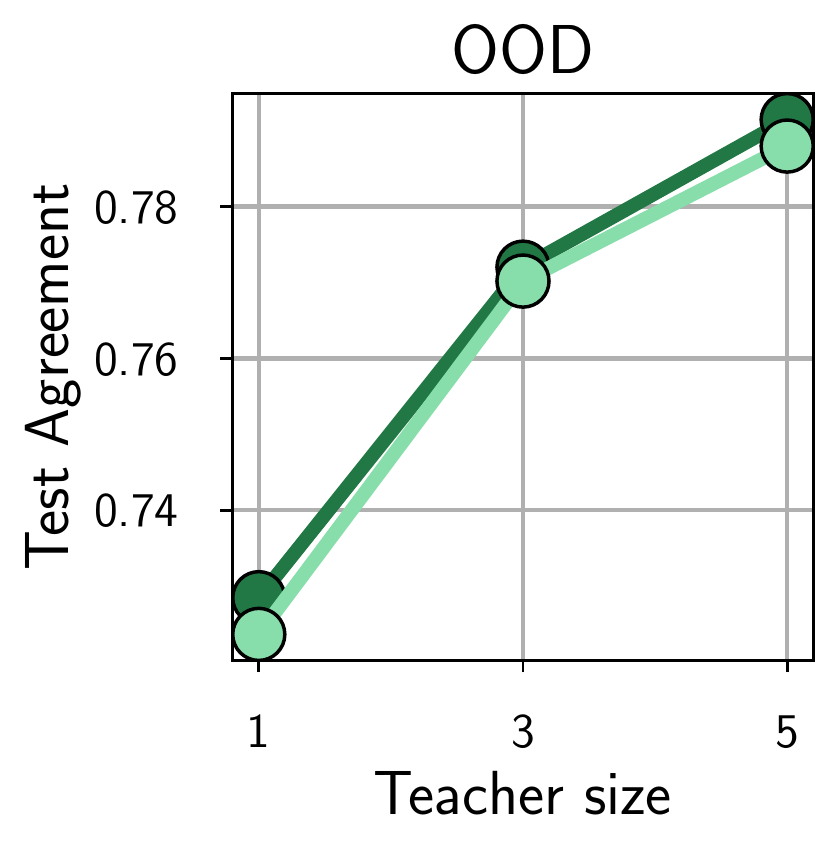} &
\includegraphics[height=0.12\textwidth]{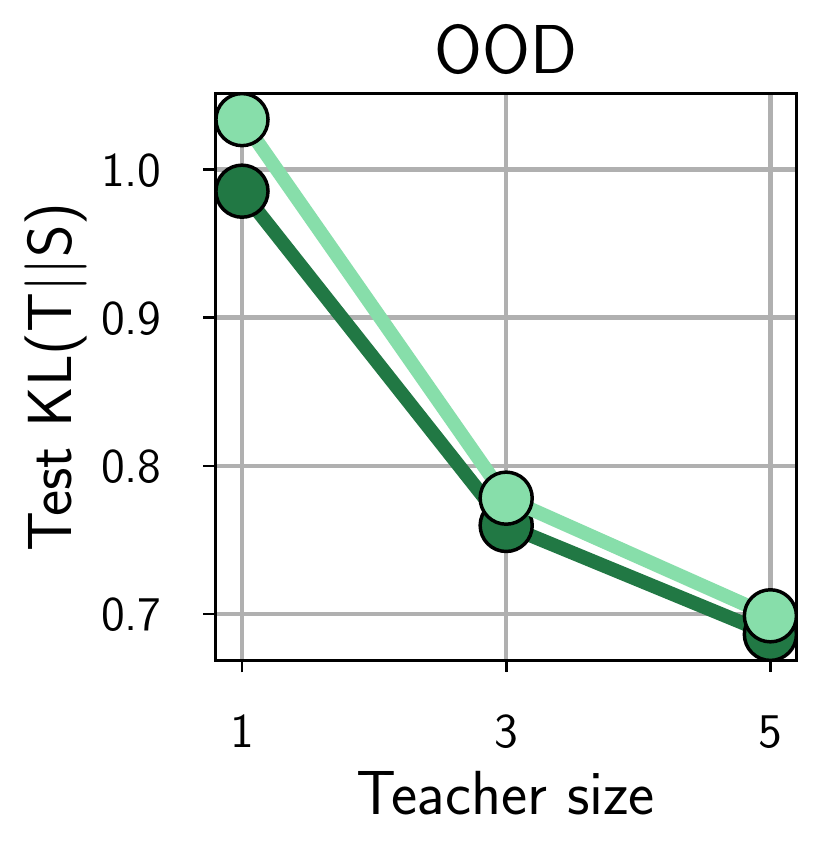}
\\[-0.2cm]
\includegraphics[height=0.12\textwidth]{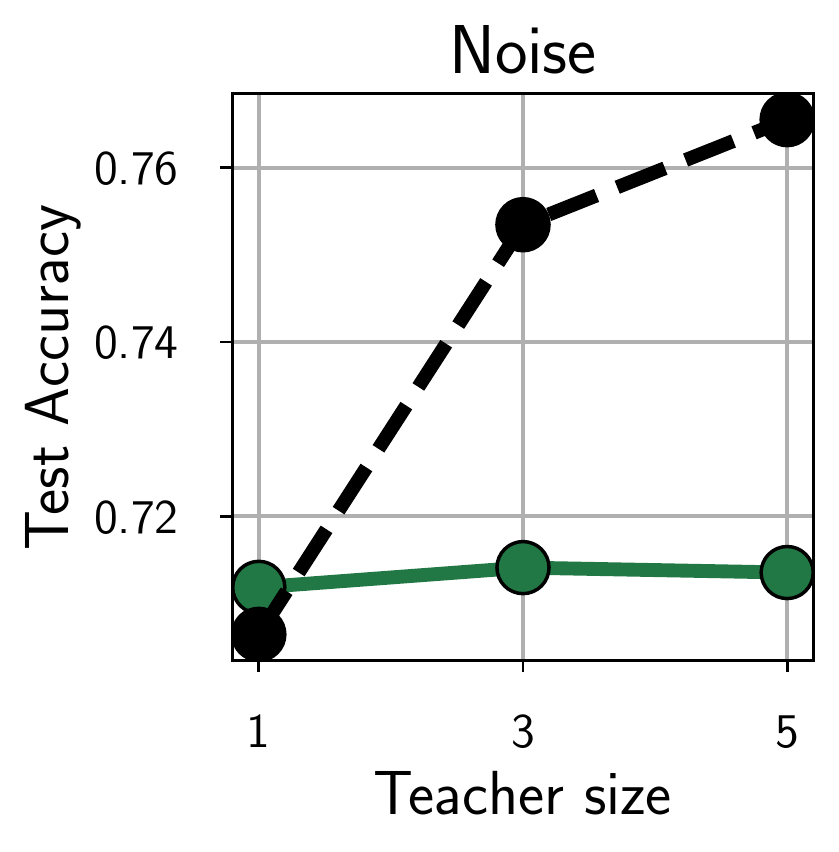} &
\includegraphics[height=0.12\textwidth]{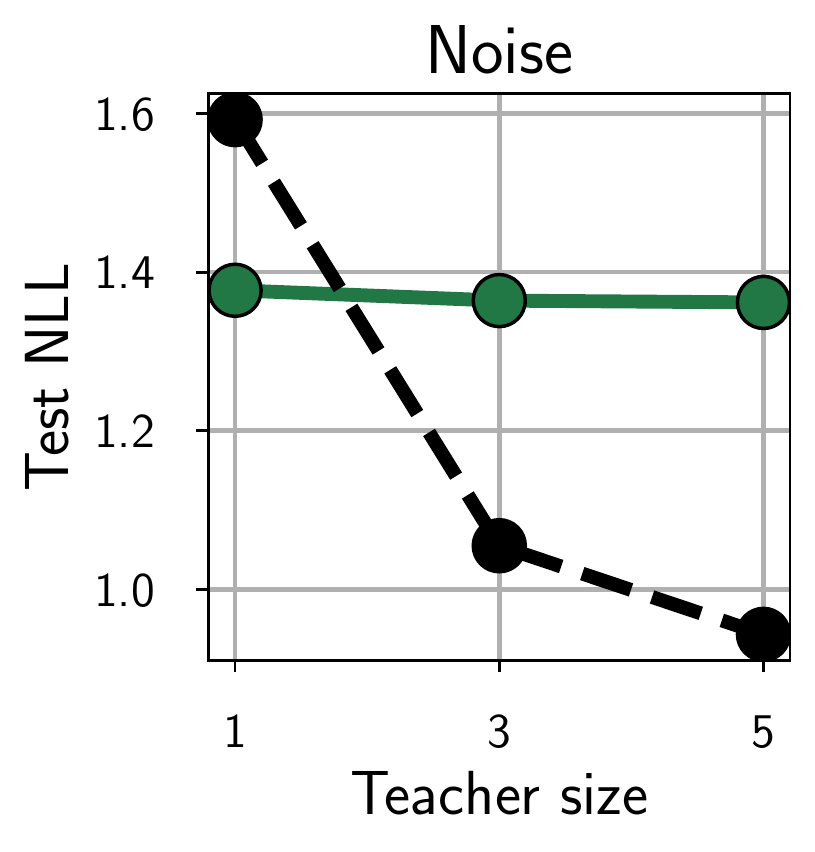} &
\includegraphics[height=0.12\textwidth]{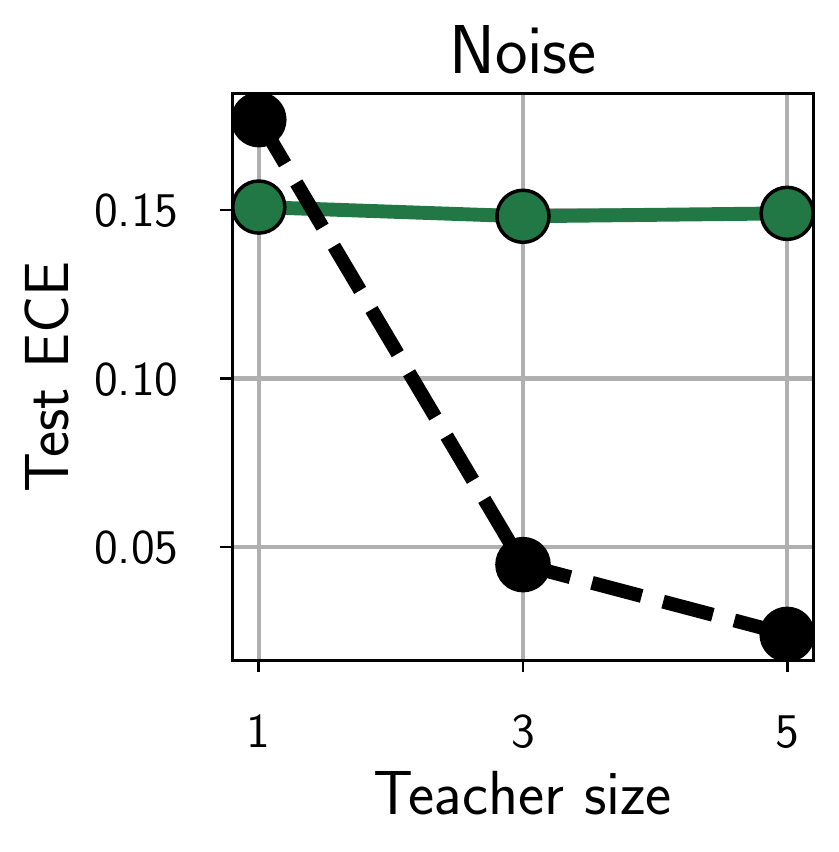} &
\includegraphics[height=0.12\textwidth]{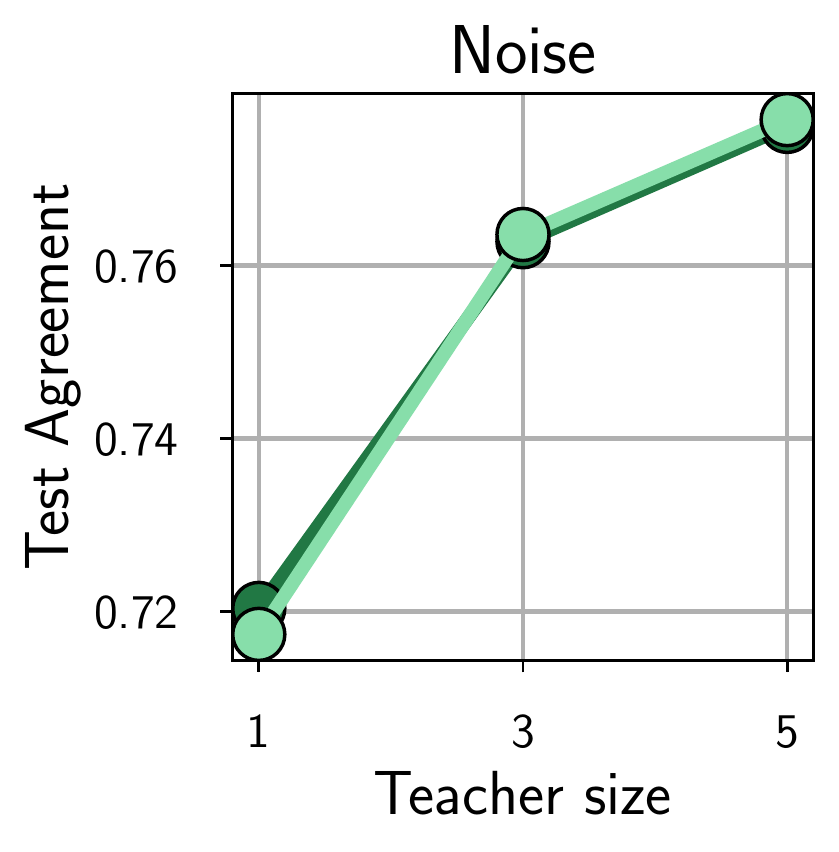} &
\includegraphics[height=0.12\textwidth]{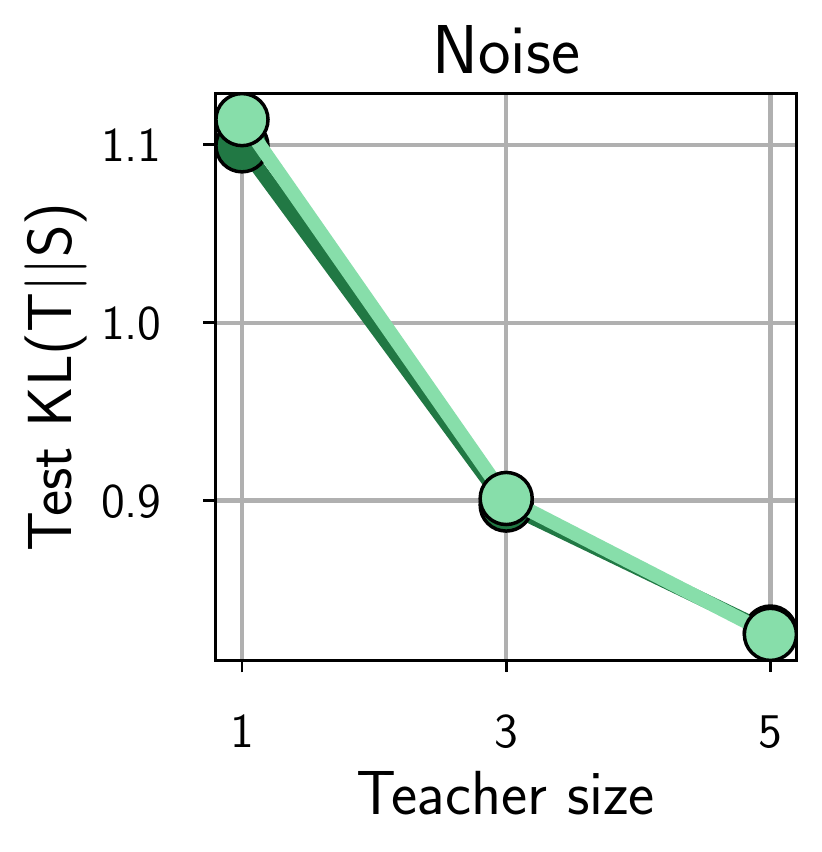}
\\[-0.2cm]
\multicolumn{5}{c}{\includegraphics[height=0.07\textwidth]{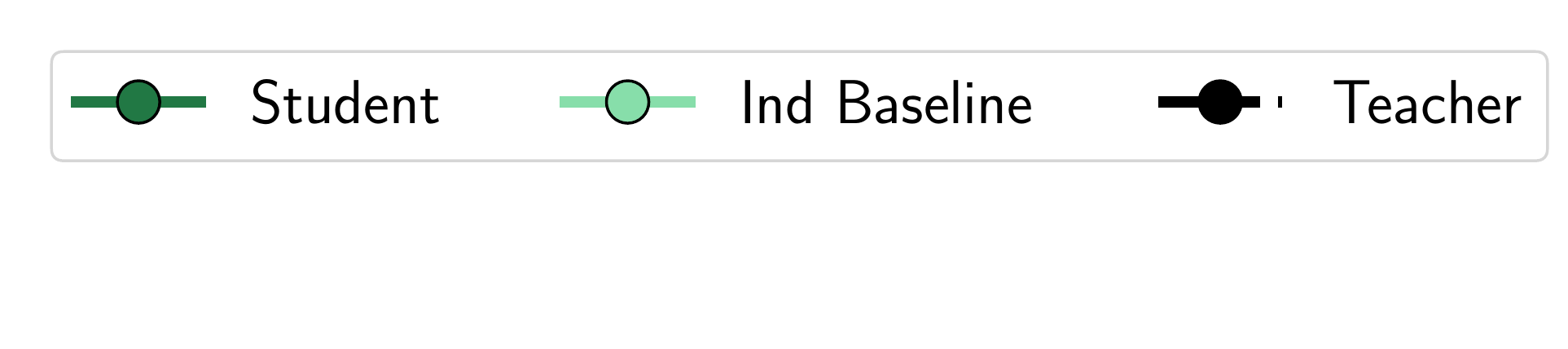}}
\\[-0.2cm]
\end{tabular}
\caption{Detailed results for the experiment in Section \ref{main/subsec:insufficient_data_hypothesis}. Each row corresponds to a different augmentation procedure, and each column is a different evaluation metric. Notably, we see that the student distilled with mixup and $\tau = 1$ is the best overall in terms of NLL (though not test accuracy) beating even the teacher-ensemble for all values of $m$. The independent baseline serves as a reference to aid in the interpretation of fidelity metrics.}
\label{supp/fig:detailed_results}
\end{figure}

In Figure \ref{supp/fig:detailed_results} we report more detailed results for the experiment in Figure \ref{main/fig:augmentation_comparison} (in the main text). In particular, for the sake of simplicity we only reported results for $m=5$ in the main text. Here we report results for $m=1$ and $m=3$ as well for comparison.

%\clearpage

\section{Addressing alternative causes of poor fidelity}
\label{supp/sec:additional_ablations}

In this section we provide evidence contrary to other possible explanations of low student fidelity posited in Section \ref{main/subsec:possible_causes}.
In section \ref{supp/subsec:student_capacity_ablation} we demonstrate that increasing student capacity does not substantially improve fidelity.
We also show that poor distillation fidelity is not specific to neural network architecture, dataset scale and data domain, and is observed for VGG networks (section \ref{supp/subsec:distilling_vgg}), larger-scale ImageNet dataset (section \ref{supp/subsec:imagenet}) and IMDB sentiment analysis classification with LSTM networks (section \ref{supp/subsec:imdb}). 
Further, in section \ref{supp/subsec:alpha_ablation} we demonstrate that the common practice of showing the student both the real labels (when available) and the teacher labels tends to decrease fidelity.

\subsection{Capacity: is the student capable of emulating the teacher?}
\label{supp/subsec:student_capacity_ablation}

\begin{figure}[h]
\centering
\begin{tabular}{ccccc}
\includegraphics[height=0.15\textwidth]{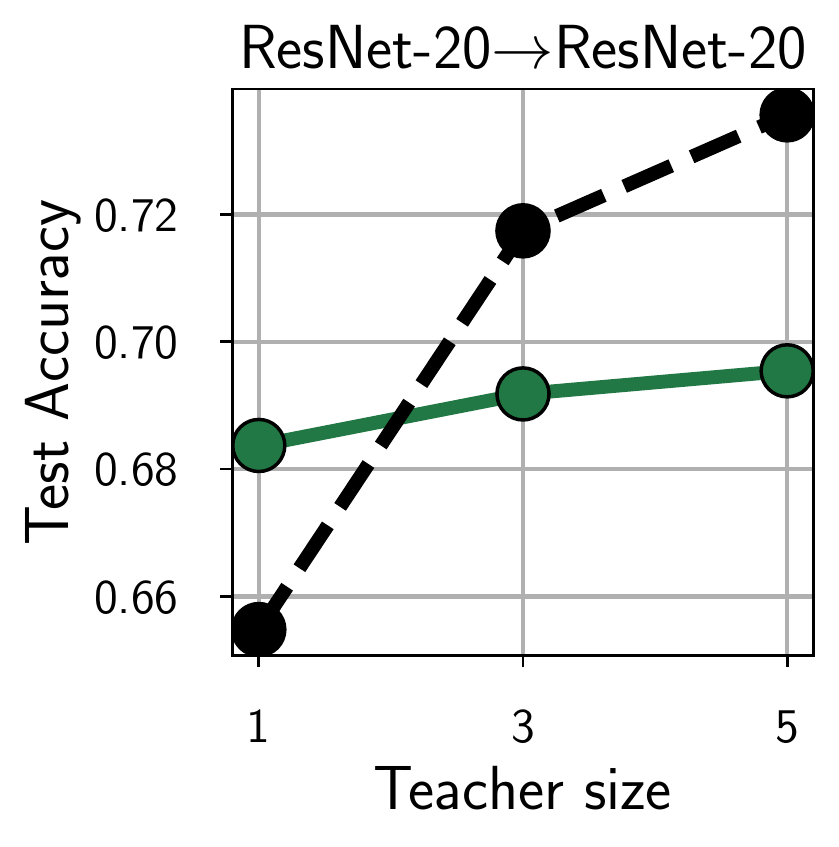} &
\includegraphics[height=0.15\textwidth]{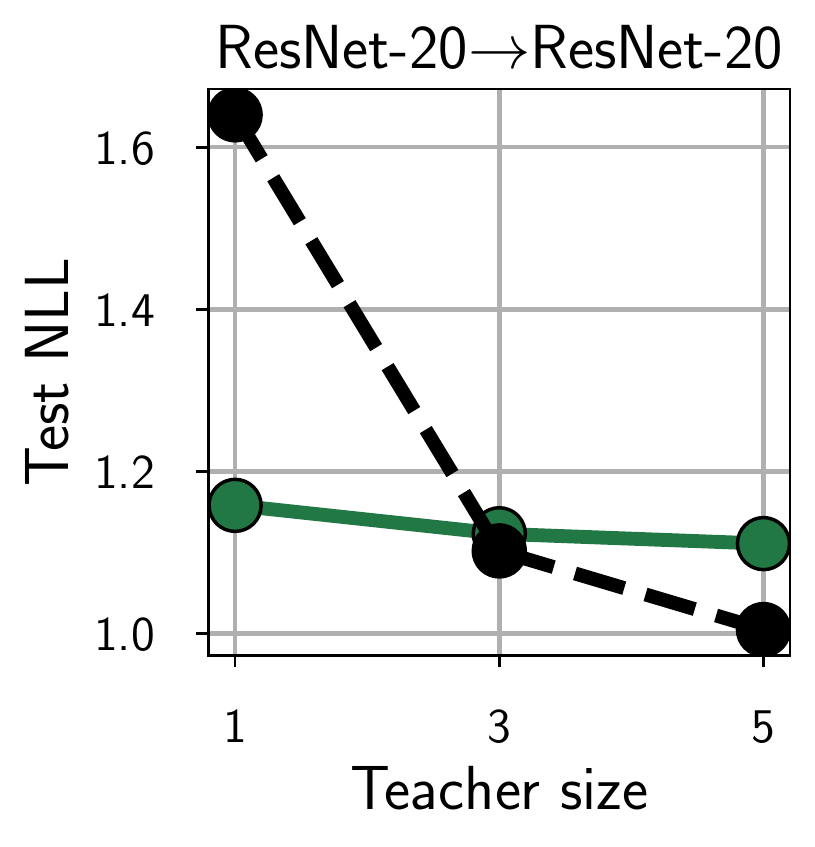} &
\includegraphics[height=0.15\textwidth]{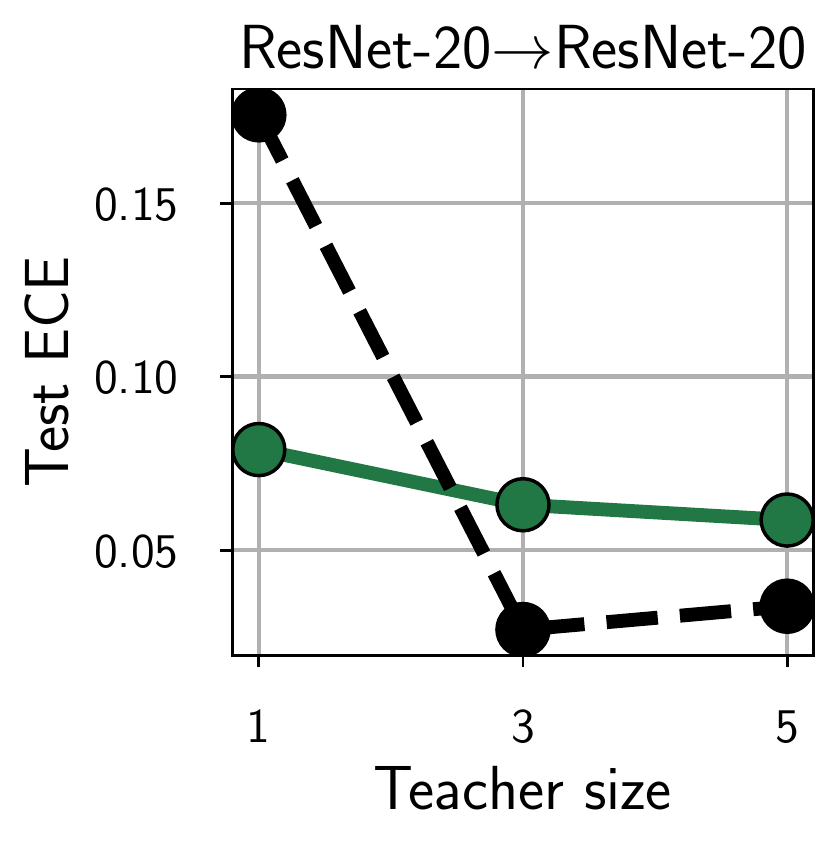} &
\includegraphics[height=0.15\textwidth]{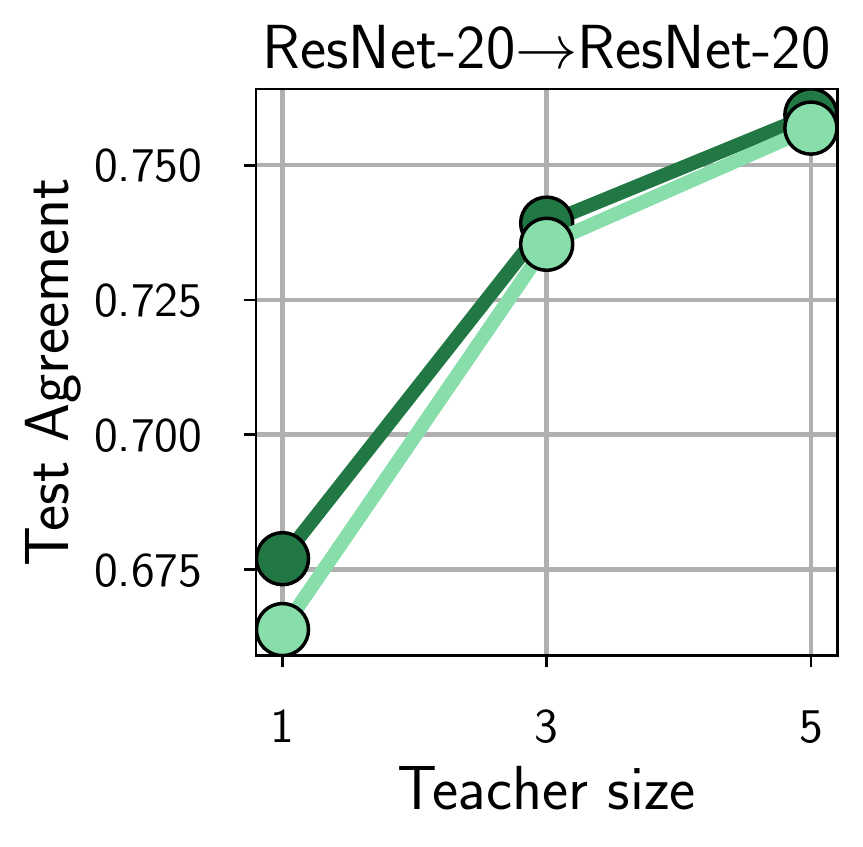} &
\includegraphics[height=0.15\textwidth]{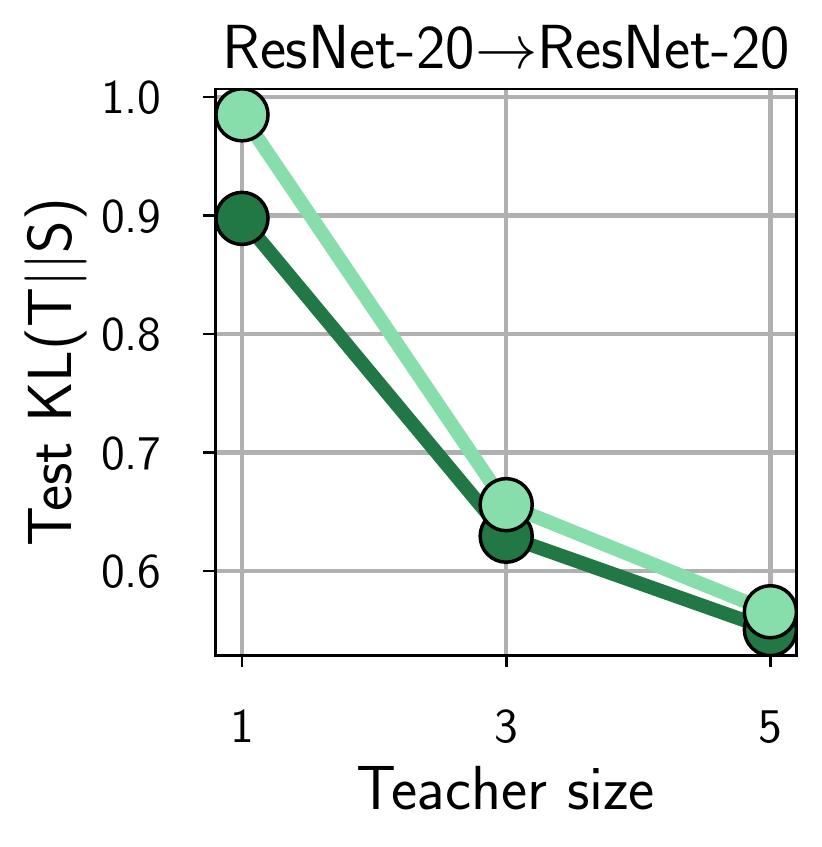}
\\[-0.cm]
\includegraphics[height=0.15\textwidth]{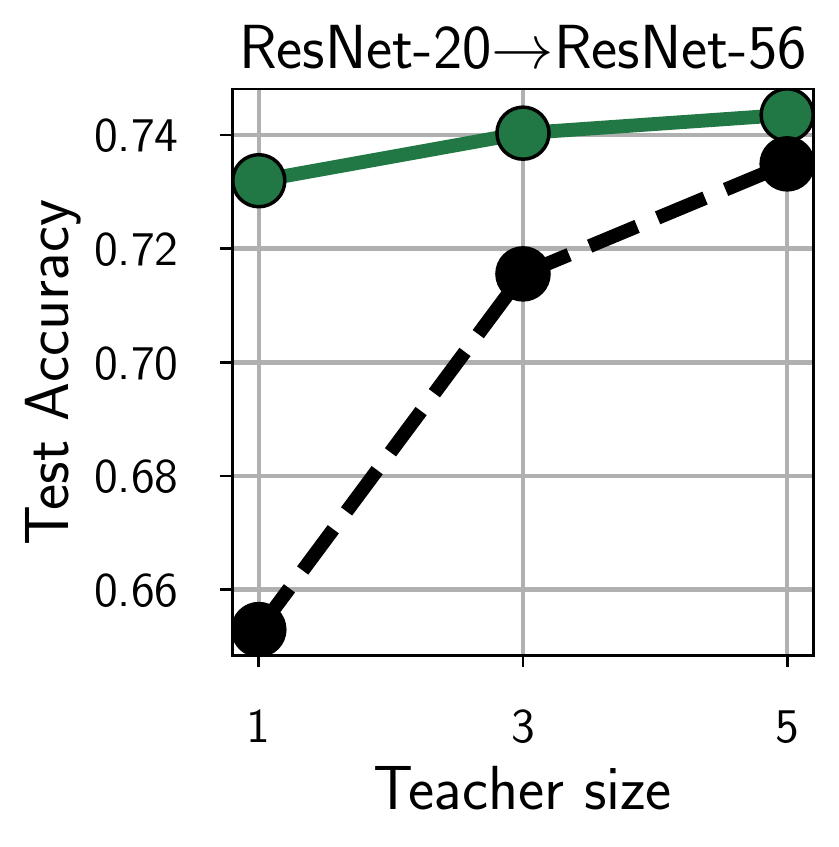} &
\includegraphics[height=0.15\textwidth]{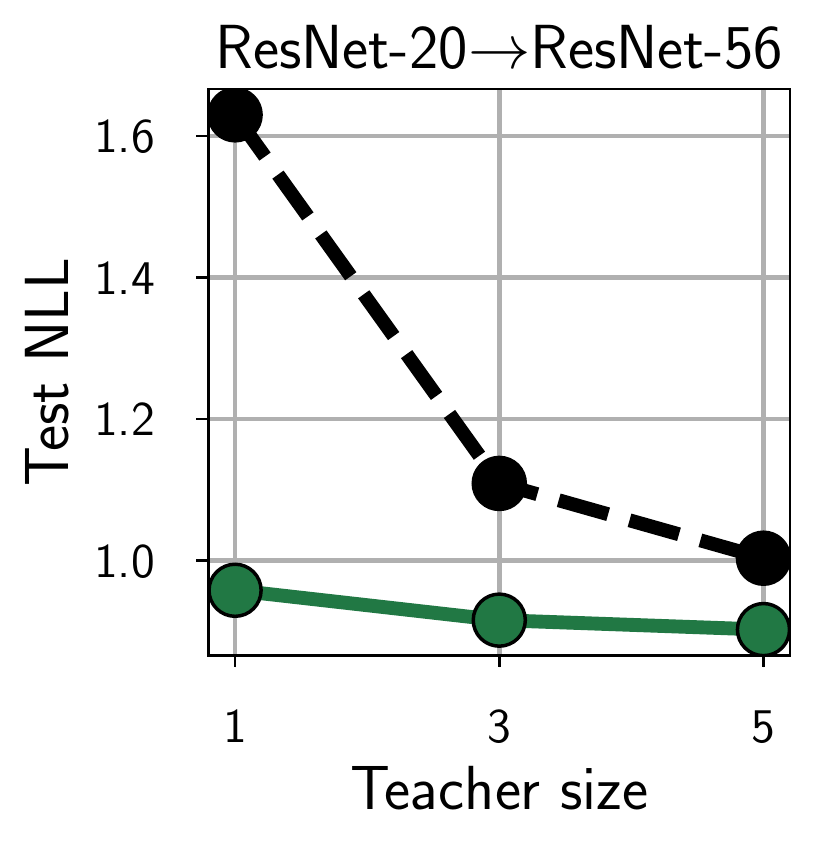} &
\includegraphics[height=0.15\textwidth]{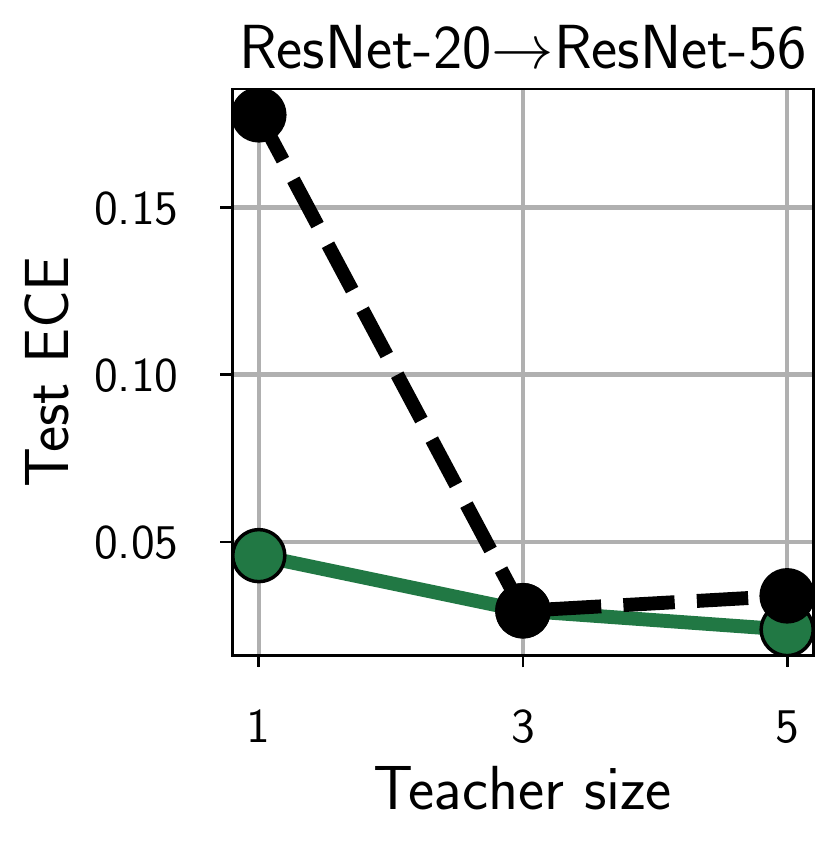} &
\includegraphics[height=0.15\textwidth]{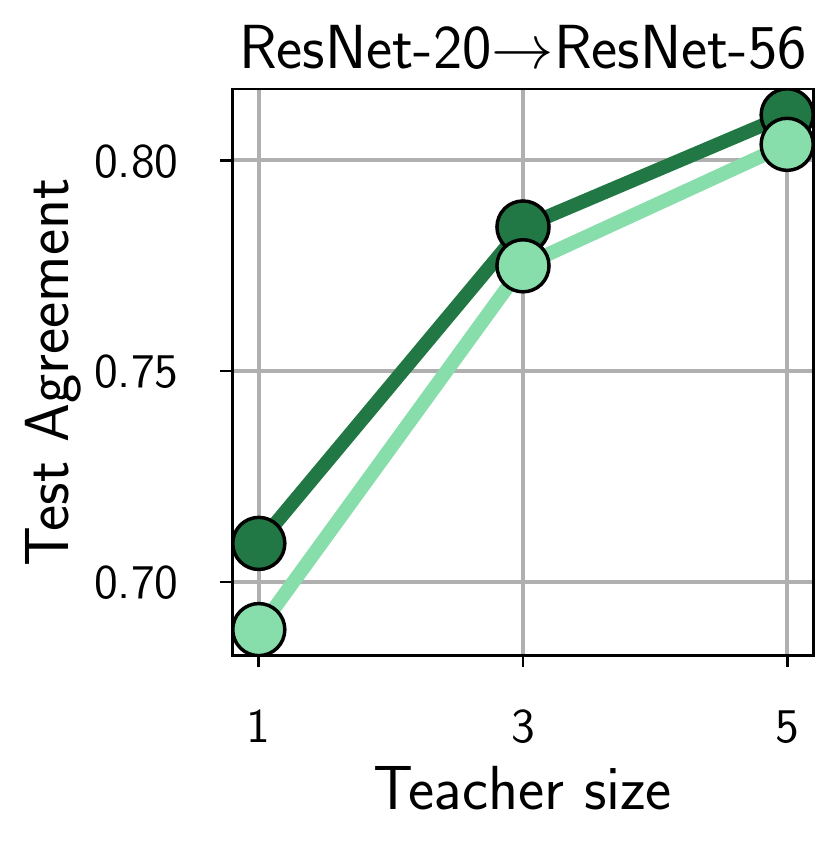} &
\includegraphics[height=0.15\textwidth]{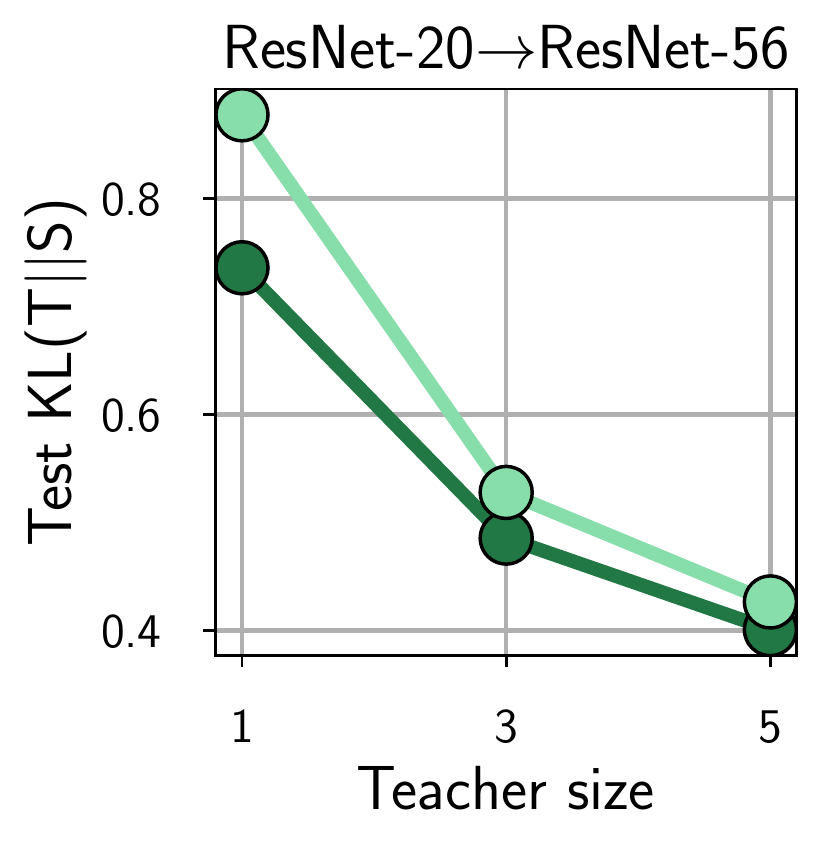}
\\[-0.cm]
\includegraphics[height=0.15\textwidth]{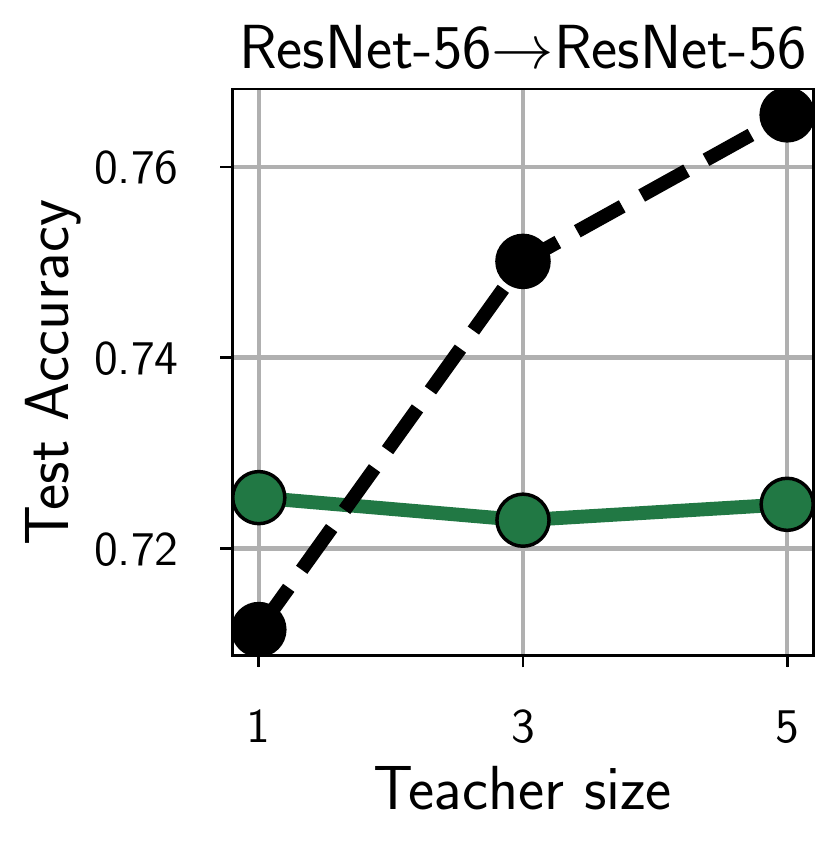} &
\includegraphics[height=0.15\textwidth]{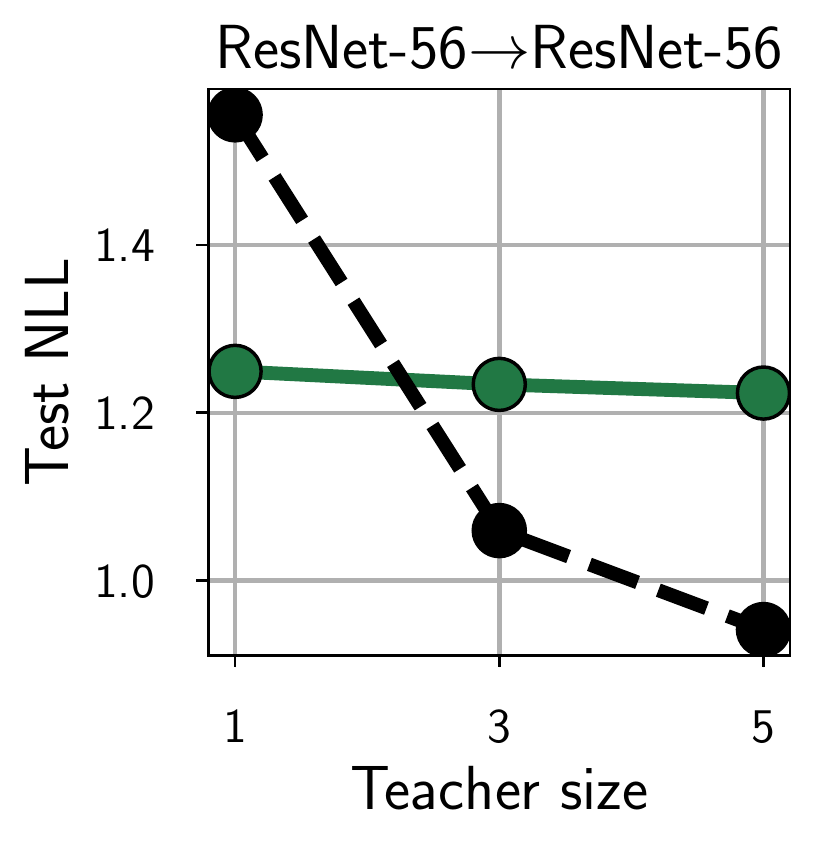} &
\includegraphics[height=0.15\textwidth]{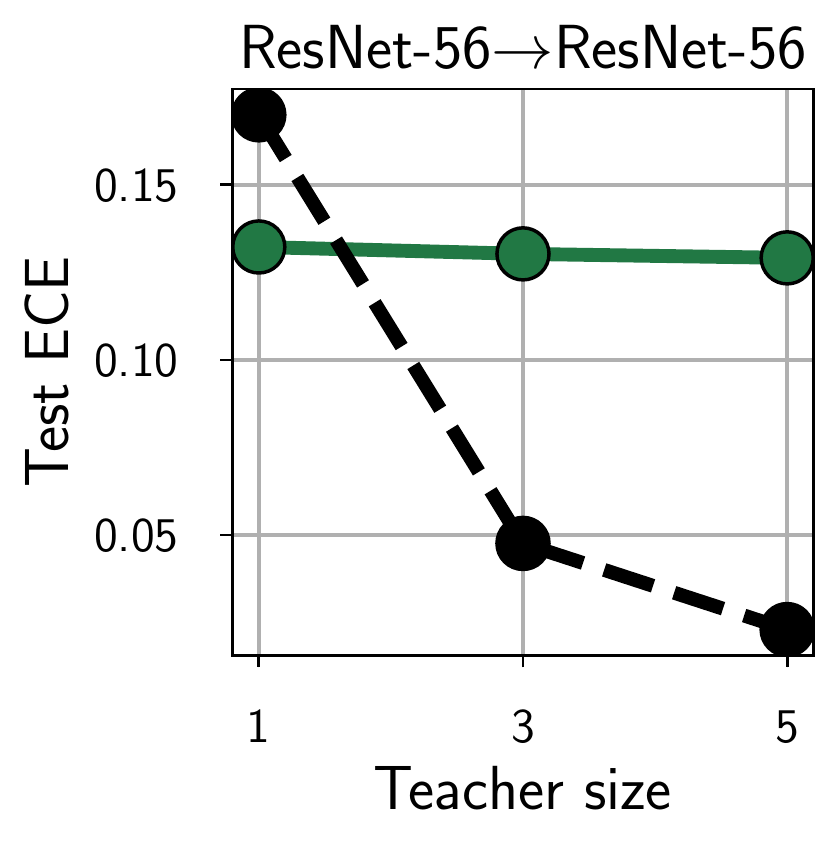} &
\includegraphics[height=0.15\textwidth]{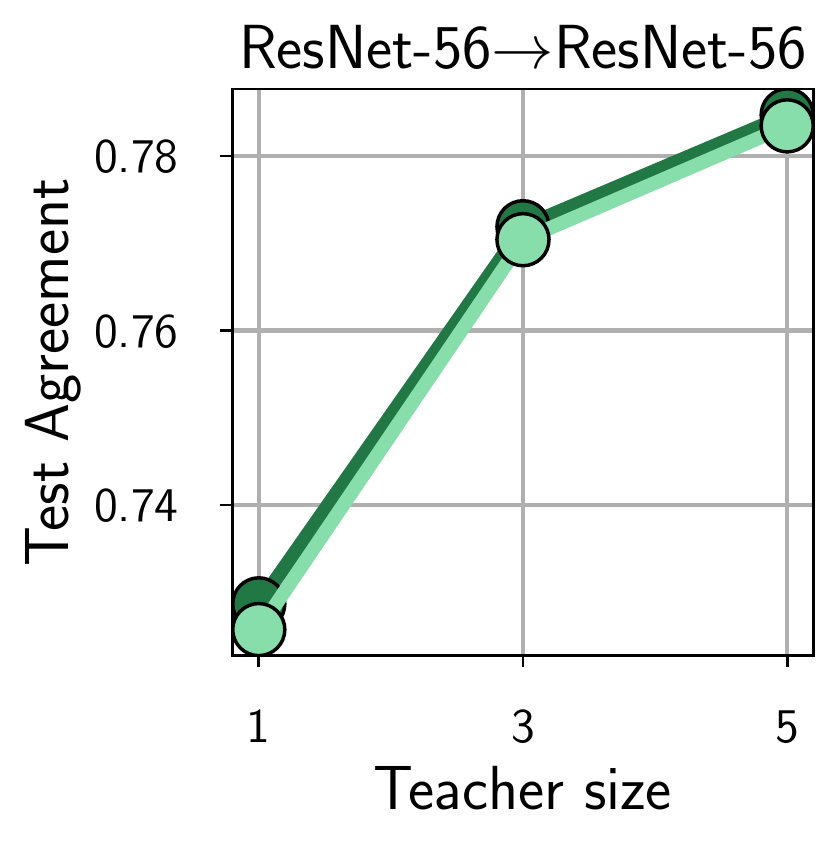} &
\includegraphics[height=0.15\textwidth]{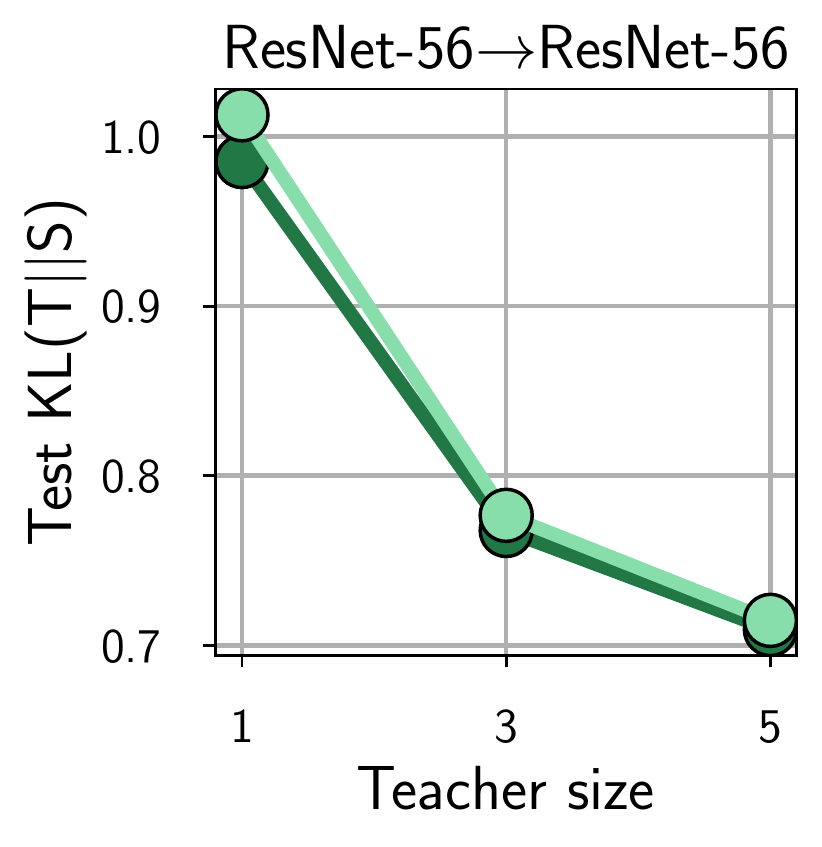}
\\[-0.cm]
\includegraphics[height=0.15\textwidth]{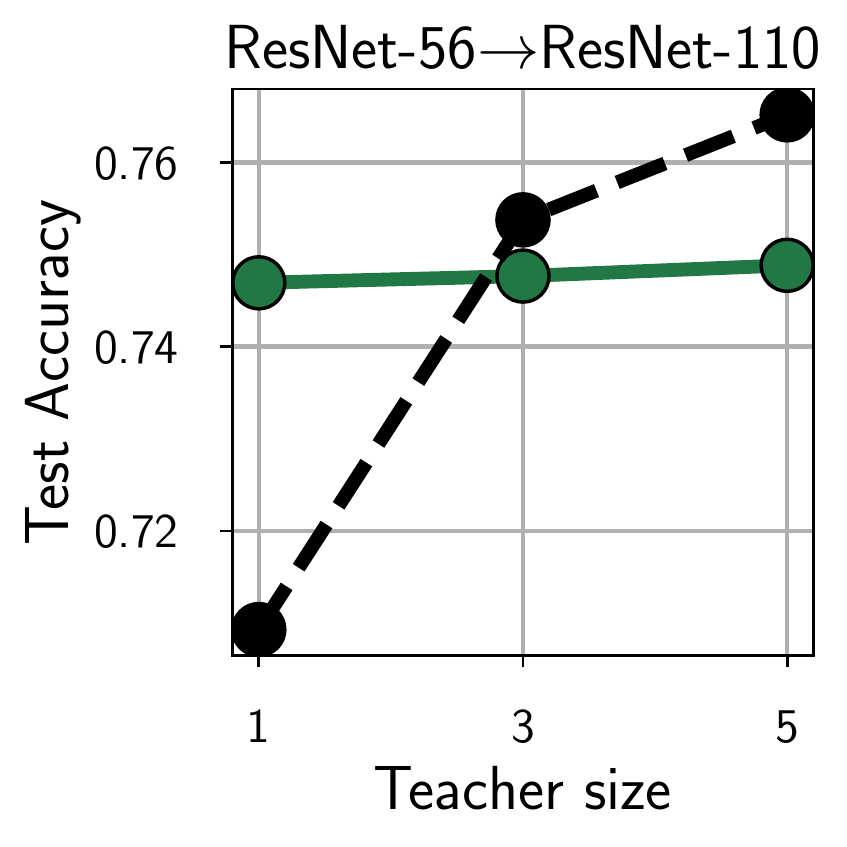} &
\includegraphics[height=0.15\textwidth]{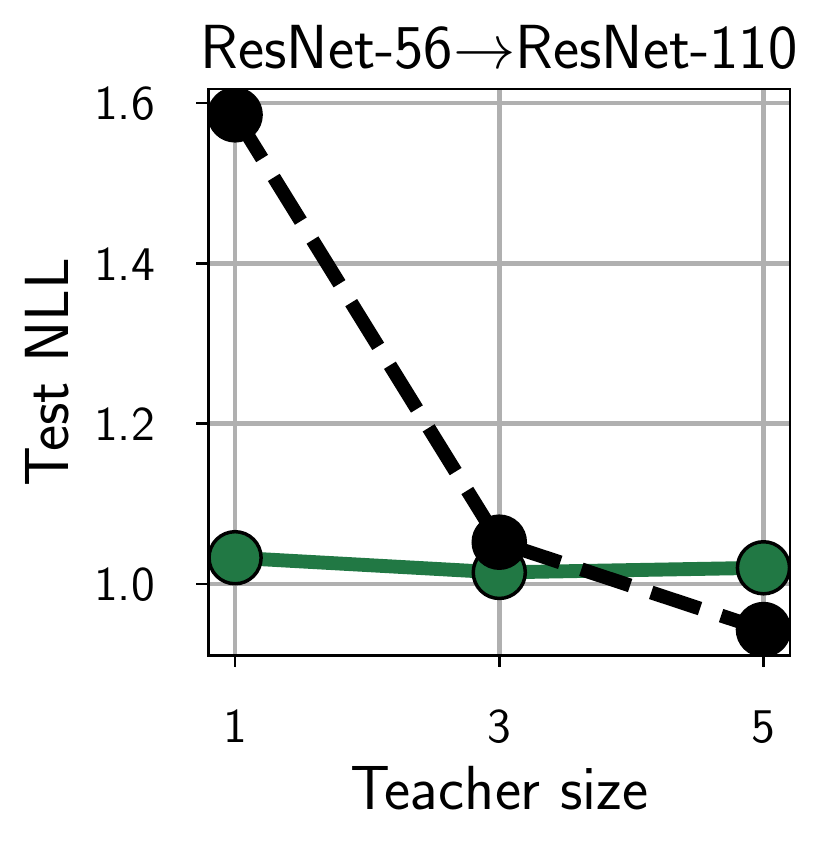} &
\includegraphics[height=0.15\textwidth]{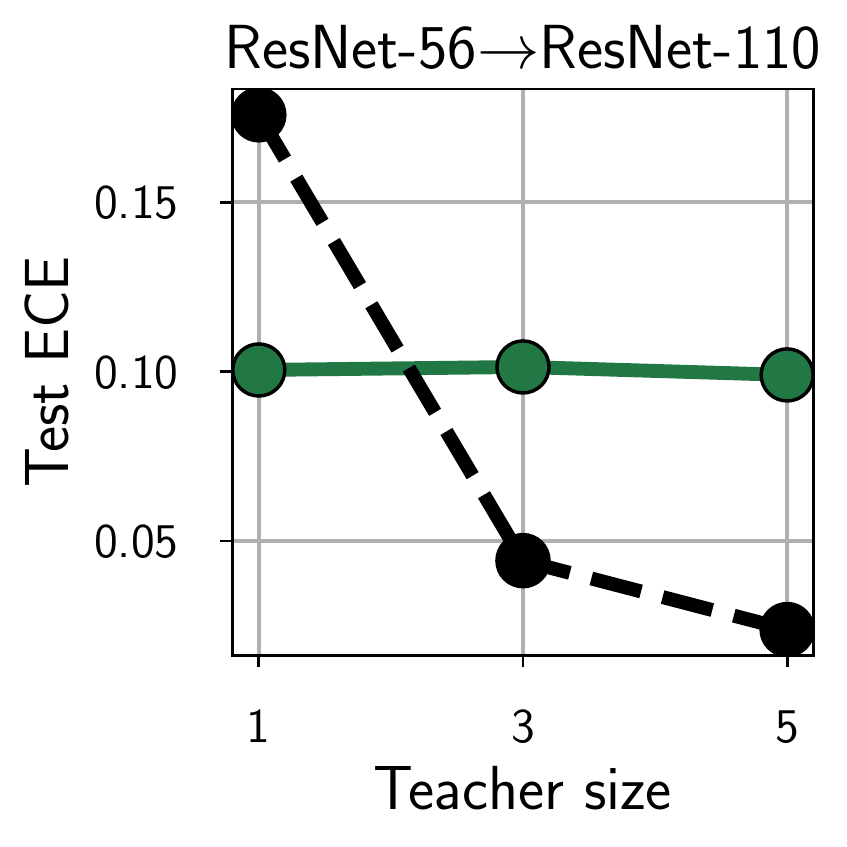} &
\includegraphics[height=0.15\textwidth]{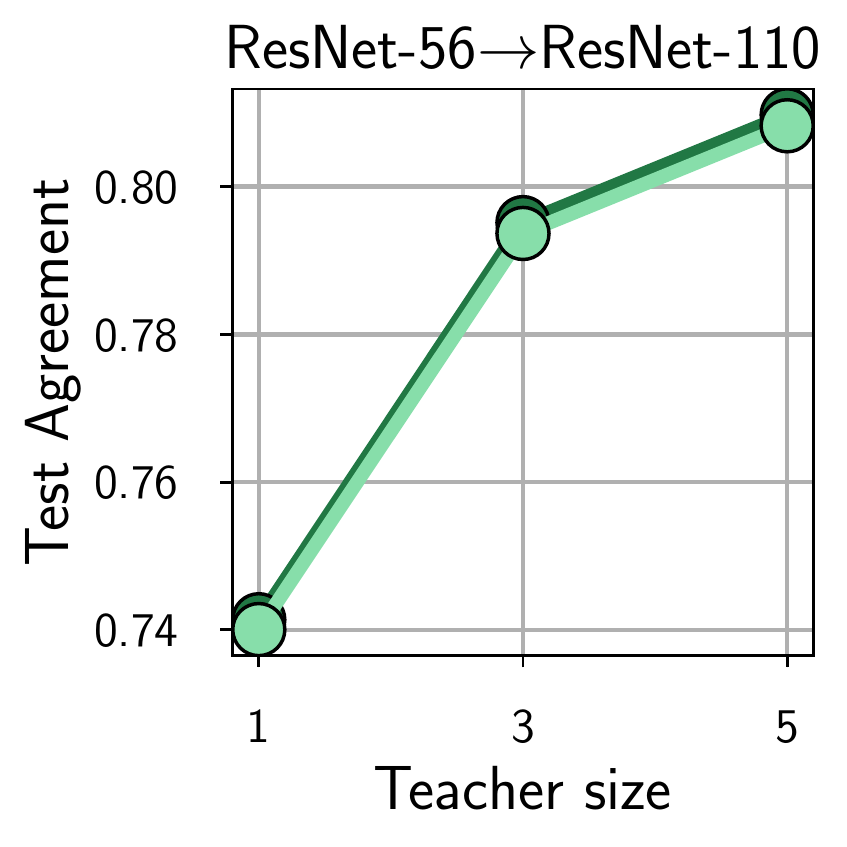} &
\includegraphics[height=0.15\textwidth]{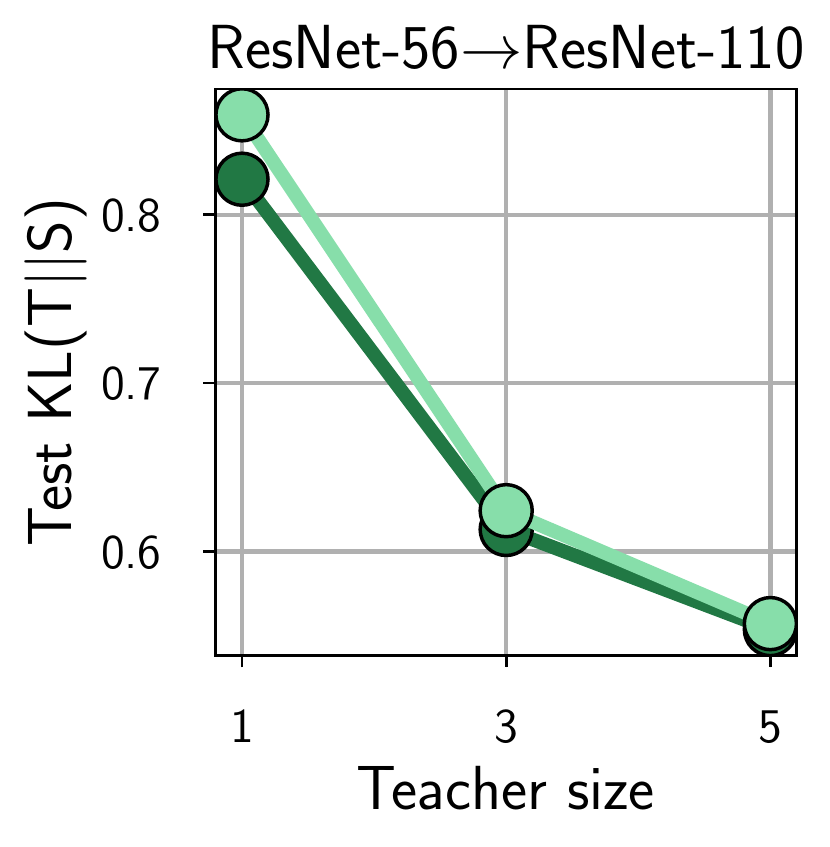}
\\[-0.cm]
\multicolumn{5}{c}{\includegraphics[height=0.07\textwidth]{figures/supp_augs_legend}}
\\[-0.2cm]
\end{tabular}
\caption{Here we show the effect of increasing the student capacity, holding the teacher capacity fixed. The top two rows correspond to ResNet20 teacher-ensemble components with ResNet20 and ResNet56 students, respectively. The bottom two rows are similarly ResNet56 teacher components with ResNet56 and ResNet110 students. The column corresponds to the evaluation metric. Increasing student capacity from 20 to 56 provides some benefit to both accuracy and fidelity, but increasing student capacity from 56 to 110 improves only accuracy.}
\label{supp/fig:capacity_ablation}
\end{figure}

One possible cause of low student fidelity when distilling ResNet ensembles on CIFAR-100 is that a single student network does not have \textit{capacity} to perfectly emulate an ensemble of multiple networks.
This explanation is already rendered unlikely by our similar observations in the context of self-distillation.
Nevertheless in Figure \ref{supp/fig:capacity_ablation} we demonstrate the effect of increasing student capacity beyond that of the individual teacher components as additional contrary evidence to the capacity explanation. 
Increasing the student capacity does slightly improve fidelity -- doubling the student network depth results in a 2\% to 3\% improvement in test agreement.
If capacity were a primary cause of low fidelity, we would expect a much larger effect on distillation fidelity when the student capacity is significantly increased.

\subsection{Architecture: is low fidelity an artifact of using ResNets?}\label{supp/subsec:distilling_vgg}

\begin{figure}
    \centering
    \includegraphics[width=0.99\textwidth]{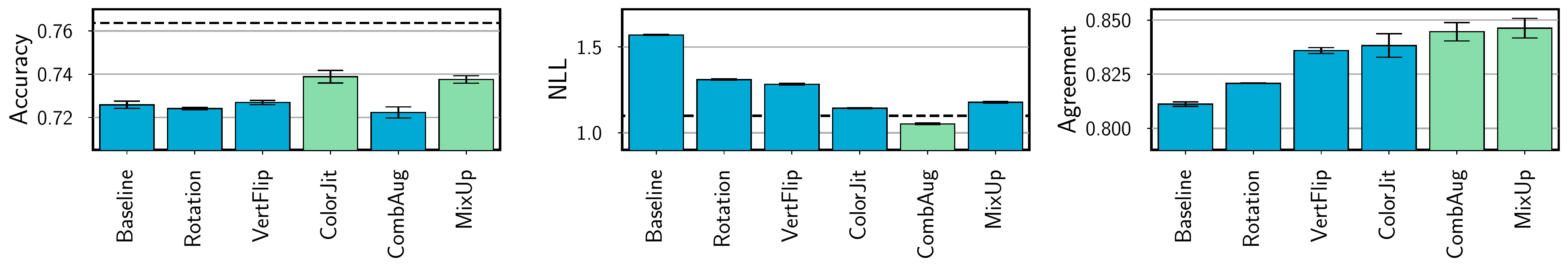}
    \caption{Test accuracy, negative log-likelihood and teacher-student agreement when
    distilling a 5-component VGG-16 teacher ensemble into a VGG-16 student on CIFAR-100 with
    varying augmentation policies.
    The best performing policy is shown in green, results averaged over 3
    runs and error bars indicate $\pm \sigma$.
    The results are generally analogous to the results for the ResNet-56 architecture reported in Section \ref{main/subsec:insufficient_data_hypothesis}.
    \textit{MixUp} and \textit{ColorJit} provide the best student accuracy, while \textit{CombAug} provides the best NLL.
    \textit{CombAug} and \textit{MixUp} provide the best teacher-student agreement.
    }
    \label{supp/fig:vgg_distillation_results}
\end{figure}

Another possible explanation of low student fidelity in our experiments is our choice of network architecture.
ResNet-style backbones are ubiquitous across most computer vision tasks, so even were the issue restricted to ResNets it would merit close investigation. 
Nevertheless, in the interest of empirical rigor we repeat the augmentation ablation in Section \ref{main/subsec:insufficient_data_hypothesis} with VGG networks and a subset of the augmentation policies.
In Figure \ref{supp/fig:vgg_distillation_results}, the teacher is a 5-component ensemble of VGG-16 networks trained with the \textit{Baseline} augmentation policy (horizontal flips and random crops).
We report the student accuracy, negative log-likelihood and teacher-student agreement for a VGG-16 student trained with different data augmentation policies.

The results are generally analogous to the ones for ResNet-56 presented in Section \ref{main/subsec:insufficient_data_hypothesis}.
The \textit{CombAug} augmentation strategy underperforms all other strategies, including \textit{Baseline},
on student accuracy, but provides the best results on NLL and only slightly loses to \textit{MixUp} on teacher-student agreement.
This result again highlights that the best augmentation policies for generalization do not necessarily provide the best distillation fidelity.
Finally, regardless of the augmentation strategy, the agreement on test does not exceed $85\%$.

\subsection{Dataset: does increasing the scale of the dataset increase fidelity?}\label{supp/subsec:imagenet}

We provide the results for distilling ensembles of $1$, $3$ and $5$ ResNet-50 teachers into a single ResNet-50 model in Table \ref{supp/tab:dataset_ablation}. For each setting, we report the results averaged over 3 independent runs. The results further validate our CIFAR-100 experiments. In particular, top-1 agreement is again in the $80-90$\% range, adding more ensemble components to the teacher improves student accuracy and fidelity, and both the accuracy and fidelity gap between teacher and student can be observed.

\begin{table}[h] 
\centering 
\begin{tabular}{lccccc} 
\toprule 
\textbf{Dataset} & \textbf{Teach. Size} & \textbf{Teach. Acc. $(\uparrow)$} & \textbf{Stud. Acc. $(\uparrow)$} & \textbf{Agree. $(\uparrow)$} & \textbf{KL} $(\downarrow)$\\ 
\midrule 
\multirow{3}{4.75em}{IMDB} 
& 1 & 79.361 (0.132) & 80.353 (0.198) & 86.488 (0.521) & 0.124 (0.012)\\ 
& 3 & 81.807 (0.129) & 81.129 (0.057) & 89.832 (0.349) & 0.064 (0.003)\\ 
& 5 & \textbf{82.216 (0.207)} & \textbf{81.167 (0.196)} & \textbf{90.793 (0.180)} & \textbf{0.052 (0.001)}\\ 
\midrule 
\multirow{3}{4.75em}{ImageNet} 
& 1 & 0.748 (0.001) & 0.753 (0.001) & 0.855 (0.001) & 0.217 (0.002)\\ 
& 3 & 0.764 (0.001) & 0.755 (0.001) & 0.878 (0.001) & 0.157 (0.001)\\ 
& 5 & \textbf{0.767 (0.001)} & \textbf{0.756 (0.001)} & \textbf{0.884 (0.001)} & \textbf{0.142 (0.001)}\\ 
\bottomrule 
\end{tabular}
\caption{Distillation results when the dataset is varied. All metrics are computed on the test set. We used bidirectional LSTM networks for IMDB and ResNet-56 networks for CIFAR-100 and ImageNet. 
Across all datasets we see the following consistent behavior: 1) larger teacher ensembles are more accurate and easier to distill, and 2) teacher-student disagree on at least 10\% of test points.
} 
\label{supp/tab:dataset_ablation} 
\end{table}

\begin{figure}[h]
  \begin{subfigure}{0.32\textwidth}
  \includegraphics[height=0.43\textwidth]{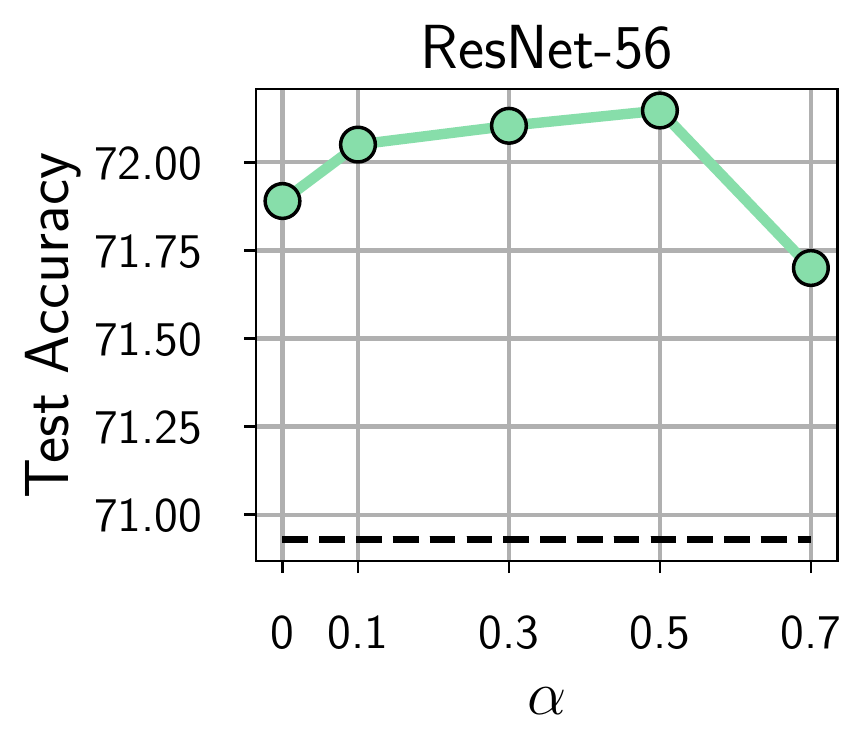}
  \includegraphics[height=0.43\textwidth]{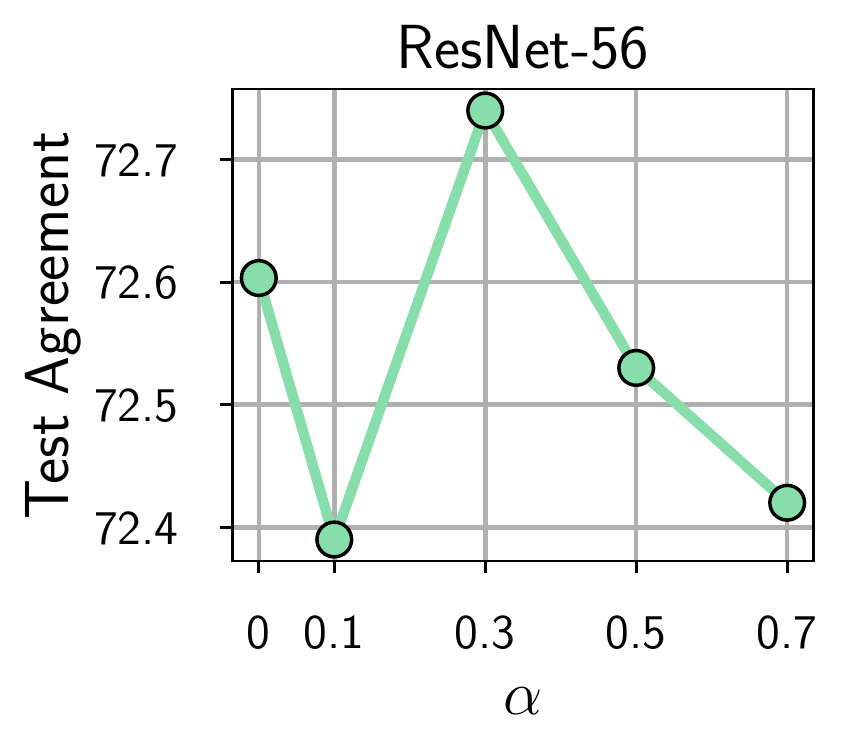}
  \caption{$1$ Teacher}
  \end{subfigure}
  \hfill
  \begin{subfigure}{0.32\textwidth}
  \includegraphics[height=0.43\textwidth]{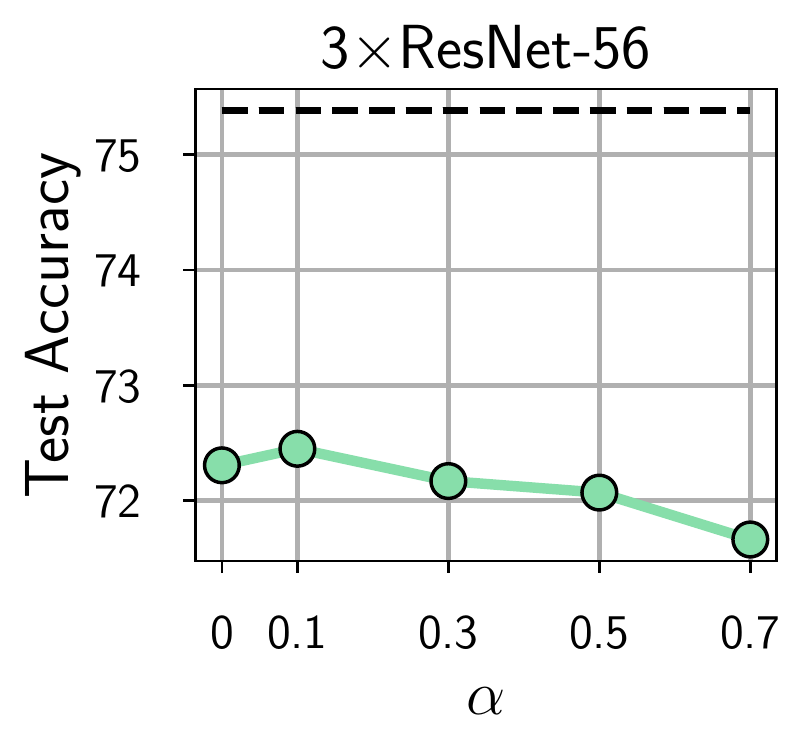}
  \includegraphics[height=0.43\textwidth]{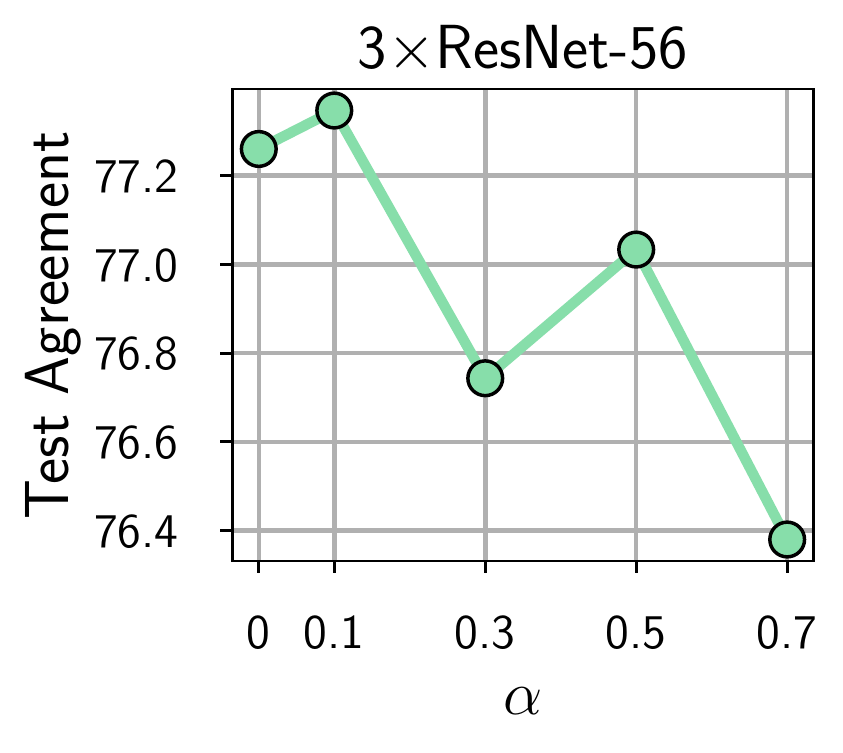}
  \caption{$3$ Teachers}
  \end{subfigure}
  \hfill
  \begin{subfigure}{0.32\textwidth}
  \includegraphics[height=0.43\textwidth]{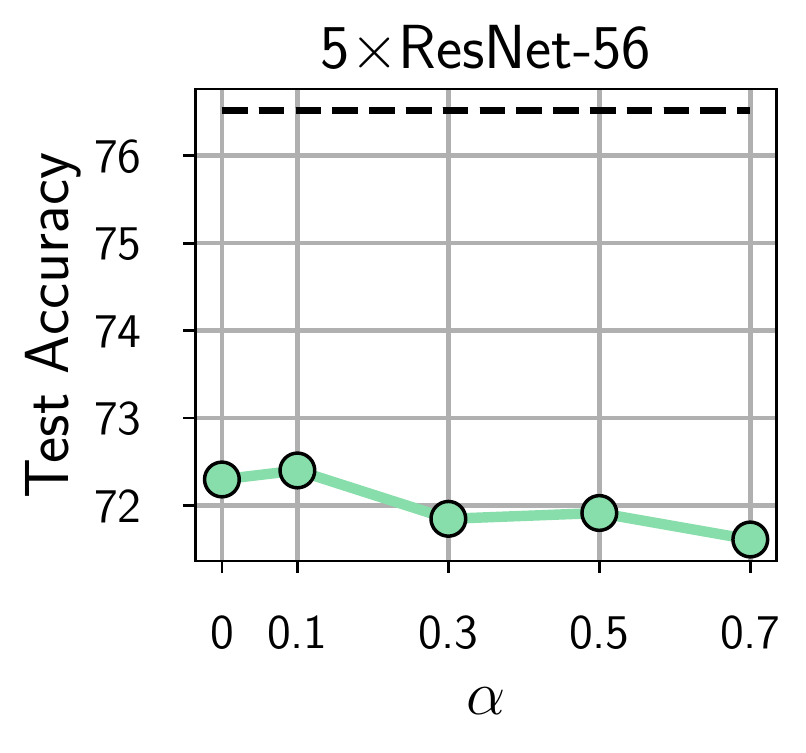}
  \includegraphics[height=0.43\textwidth]{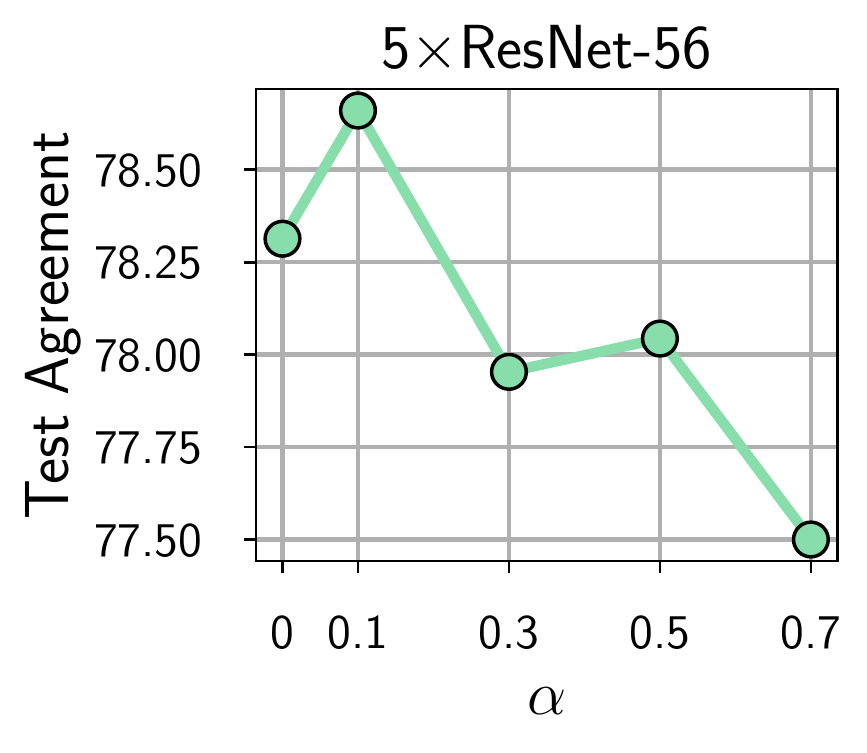}
  \caption{$5$ Teachers}
  \end{subfigure}
  \caption{Results ablating $\alpha$ $(\tau=1)$. Taking $\alpha > 0$ can improve student accuracy in the self-distillation regime, but does not consistently improve teacher-student agreement. When $k > 0$ there is a slight benefit at $\alpha=0.1$, after which the effect is negative for both accuracy and agreement.}
  \label{supp/fig:alpha_ablation}
\end{figure}

\subsection{Data domain: is low fidelity specific to image classification?}\label{supp/subsec:imdb}
To expand out results beyond the image domain, we also demonstrate the knowledge distillation results when distilling LSTM text classifiers on IMDB sentiment analysis data in Table \ref{supp/tab:dataset_ablation}. We distill ensembles of $2$-layer bidirectional LSTM teachers into a single bidirectional LSTM of the same architecture. Note that like in our other results, the teacher accuracy is strictly improving, whereas the student accuracy ceases to improve from 3 to 5 teacher ensemble components, which is associated with a similar lack of improvement in agreement.

\subsection{Does showing the student ground truth labels improve fidelity?}
\label{supp/subsec:alpha_ablation}

In Figure \ref{supp/fig:alpha_ablation} we investigate the effect of the relative weight of the distillation loss terms $\mathcal{L}_{\mathrm{NLL}}$ and $\mathcal{L}_{\mathrm{KD}}$ when distilling teacher-ensembles with ResNet56 components into a ResNet56 student on CIFAR-100 with $\tau=1$. We observe that in the self-distillation regime taking $\alpha > 0$ improves test accuracy, but not test agreement.  When $k > 0$, there is a slight benefit when $\alpha=0.1$, but for most values tried the effect was deleterious to both accuracy and fidelity.

\end{document}